\documentclass[aoas, prepint]{imsart}

\RequirePackage{amsthm,amsmath,amsfonts,amssymb}
\RequirePackage{algorithm}
\RequirePackage[authoryear]{natbib}
\RequirePackage[colorlinks,citecolor=blue,urlcolor=blue]{hyperref}
\RequirePackage{graphicx}
\RequirePackage{booktabs}
\RequirePackage{multirow}
\usepackage[font=footnotesize]{caption}

\startlocaldefs

\endlocaldefs

\begin{document}

\begin{frontmatter}
\title{Scalable Bayesian Additive Models for Stellar Flare Detection via Amortized Gaussian Process Inference and Hidden Markov Models}
\runtitle{Scalable Bayesian models for Stellar Flare detection}

\begin{aug}
\author[A,B]{\fnms{Rodrigo}~\snm{Herrera}\ead[label=e1]{r.herrera@mail.utoronto.ca}},
\author[A,B,C]{\fnms{Vianey}~\snm{Leos-Barajas}\ead[label=e2]{vianey.leosbarajas@utoronto.ca}},
\author[A,B,D]{\fnms{Gwendolyn}~\snm{Eadie}\ead[label=e3]{gwen.eadie@utoronto.ca}},
\author[E]{\fnms{Elizaveta}~\snm{Semenova}\ead[label=e4]{e.semenova@imperial.ac.uk}}
\and
\author[F]{\fnms{James}~\snm{Davenport}\ead[label=e5]{jrad@uw.edu}}

\address[A]{Department of Statistical Sciences, University of Toronto\printead[presep={,\ }]{e1}}

\address[B]{Data Sciences Institute, University of Toronto\printead[presep={,\ }]{e2}}

\address[C]{School of the Environment, University of Toronto\printead[presep={,\ }]{e3}}

\address[D]{David A. Dunlap Department of Astronomy and Astrophysics, University of Toronto\printead[presep={,\ }]{e4}}

\address[E]{School of Public Health, Imperial College London\printead[presep={,\ }]{e5}}

\address[F]{Department of Astronomy, University of Washington}

\end{aug}

\begin{abstract}
Gaussian Processes (GPs) are a powerful tool for Bayesian time-series modeling, yet their cubic computational cost remains a severe barrier for application to long, high-cadence datasets in astronomy. While specialized scalable solvers like \textit{Celerite} elegantly reduce this scaling to linear time, repeatedly evaluating the exact likelihood during iterative Bayesian sampling is a bottleneck for developing more complex models, like hierarchical or additive models in which \textit{Celerite} is only one component. To make this inference computationally tractable, we introduce a generative surrogate framework. By utilizing a Variational Autoencoder (VAE) to learn a compressed representation of the \textit{Celerite} prior, we map highly correlated stochastic dependencies into a low-dimensional, isotropic manifold. This transition completely bypasses exact covariance operations, shifting the computational burden to a rapid neural network forward pass. 
Through an extensive simulation study, we show that the generative surrogate accurately reproduces the structural fidelity of exact physical kernels like \textit{Celerite}. Finally, we demonstrate embedding our VAE approximation into an additive model that combines \textit{Celerite} and a hidden Markov model (HMM) for stellar flare detection in time series data of stars. We evaluate the joint VAE+HMM architecture against the exact \textit{Celerite}+HMM framework on empirical astrophysical time series and demonstrate that the proposed methodology achieves significant reductions in computational time, enabling the rigorous, large-scale characterization of stellar flares across massive data archives.
\end{abstract}


\end{frontmatter}



\section{Introduction}

Gaussian Processes (GPs) provide a highly flexible framework to define distributions over continuous functions. Because of their ability to adapt to complex covariance structures without requiring rigid parametric assumptions, GPs are frequently embedded as foundational components within larger hybrid or additive architectures. This structure appears across diverse scientific domains. In epidemiology, geostatistical models utilize GPs to model spatial correlations in disease risk maps, mixing the continuous spatial field with Poisson processes for discrete case counts \citep{diggle1998model}. In neuroscience, the Poisson-GP Latent Variable Model decomposes neural spike trains into a continuous latent trajectory and discrete firing events \citep{wu2017gaussian}. In finance, Stochastic Volatility Models employ a GP to capture the latent, time-varying volatility surface while a discrete state-space model handles regime shifts \citep{wu2017gaussian}. However, the widespread adoption of these additive GP frameworks is limited by its associated computational cost \citep{quinonero2005unifying, hensman2013gaussian}. Evaluating the marginal likelihood of an exact GP requires operations on an $N \times N$ covariance matrix, leading to a prohibitive $\mathcal{O}(N^3)$ computational scaling. When these models are evaluated within an iterative inference loop, this algebraic overhead renders the models intractable for massive datasets \citep{ambikasaran2015fast, foreman2017fast}.

Considerable research has focused on accelerating GP inference to bypass this bottleneck. Standard statistical approximations, such as the fully independent training conditional (FITC) method \citep{snelson2006} and sparse variational Gaussian Processes (SVGP) \citep{titsias2009}, attempt to reduce the dimensionality of the covariance matrix through the use of inducing points. In domain-specific applications like time series astronomy, specialized structured kernels like \textit{Celerite} have been developed to exploit semi-separable matrix properties, reducing the exact GP cost to $\mathcal{O}(N)$ \citep{foreman2017fast}. Alternatively, recent advancements in deep generative modeling have proposed replacing the GP entirely. Notably, the PriorVAE framework \citep{semenova2022priorvae} demonstrated that a variational autoencoder (VAE) can successfully encode spatial GP priors, shifting the heavy computational burden to an offline training phase and allowing rapid, deterministic function generation during inference. This neural-surrogate strategy provides a highly scalable alternative to foundational deterministic approximation frameworks for latent Gaussian models, most notably the Integrated Nested Laplace Approximation (INLA) \citep{rue2009approximate}.

While these advancements are significant, a critical gap remains in the literature regarding the application of generative surrogates to complex additive models---particularly those that compose multiple latent processes to capture real-world time-series dynamics \citep{duvenaud2011additive, roberts2013gaussian}. Sparse approximations like FITC and SVGP reduce complexity to $\mathcal{O}(NM^2)$ (where $M$ is the number of inducing points), but they still require costly matrix inversions at every inference step and frequently struggle to capture complex, high-frequency quasi-periodic structures restricting the number of inducing points inherently smooths out rapid variations \citep{bauer2016understanding, foreman2017fast}. Furthermore, even when utilizing $\mathcal{O}(N)$ optimized solvers like \textit{Celerite} within hierarchical architectures, e.g., \textit{Celerite} + hidden Markov model (HMM), the repeated evaluation of the exact GP likelihood inside the inference loop remains a significant bottleneck for large datasets \citep{esquivel2024detecting}. Exact GPs also induce highly correlated posterior geometries for their physical hyperparameters, severely hindering Markov chain Monte Carlo (MCMC) sampling efficiency \citep{filippone2013comparative, betancourt2015hamiltonian}. Consequently, there is a need for methodologies that can decouple the continuous GP approximation from the rest of the hierarchical model, thereby preserving the structural power of the GP while drastically reducing the computational complexity of the target likelihood.

This paper proposes a novel framework that bridges deep generative learning with exact probabilistic graphical modeling. Specifically, we demonstrate how a pre-trained VAE can be seamlessly integrated as a deterministic surrogate for the GP within a larger Bayesian additive model. By substituting the explicit GP term with the VAE decoder output, we collapse the intractable integral over the function space and bypass traditional covariance matrix operations. Instead of computing dense matrix inverses, the cost of generating the GP trend scales only with the forward pass of the neural network. This transforms the inference problem into finding a single latent vector with a standard normal prior, which creates an uncorrelated, isotropic latent space that is exceptionally efficient for advanced MCMC algorithms. We apply this methodology to the specific additive structure of a VAE+HMM, showing that deep learning tools can be seamlessly integrated with standard probabilistic models to facilitate highly complex inference.

To validate this proposed framework, we focus on the complex astronomical challenge of detecting stochastic stellar flares against quasi-periodic rotational backgrounds, an additive problem recently modeled via an exact GP+HMM architecture \citep{esquivel2024detecting}. Through an extensive simulation study, we systematically evaluate the architectural trade-offs of the VAE surrogate. We subsequently apply the joint VAE+HMM pipeline to real time series data of stars measured by the Transiting Exoplanet Survey Satellite (TESS) \citep{ricker2015transiting}. Our results demonstrate that the proposed surrogate not only approximates the exact GP well but can achieve massive computational speedups. 

The remainder of this paper is organized as follows. Section \ref{sec:data} describes the stellar datasets utilized in this study. Section \ref{sec:methodology} details the core methodology, transitioning from the theoretical GP formulation to the novel VAE+HMM inference pipeline. Section \ref{sec:simulation_study} presents the generative simulation study and architectural hyperparameter tuning. In Section \ref{sec:results} we present results of applying our framework to real-world astronomical time series. We conclude with a discussion in Section \ref{sec:discussion}.


\section{Data Description}
\label{sec:data}

To evaluate and demonstrate the proposed VAE+HMM framework, we utilized observations of brightness over time from three distinct M dwarf stars: TIC 031381302, TIC 089257479, and TIC 234526939. Specifically, we analyzed the two-minute short-cadence Pre-Search Data Conditioning Simple Aperture Photometry (PDCSAP) light curves (i.e., time series) \citep{jenkins2016tess} collected by the Transiting Exoplanet Survey Satellite (TESS) \citep{ricker2015transiting}. These datasets serve a dual purpose in this study, forming the observational foundation for both the generative simulation experiments detailed in Section \ref{sec:simulation_study} and the full implementation of the joint flare-detection framework presented in Section \ref{sec:results}.

\begin{figure}[t!]
    \centering
    
    \begin{tabular}{cc}
    
        \textbf{TIC 031381302} & \raisebox{-0.5\height}{\includegraphics[width=0.65\columnwidth]{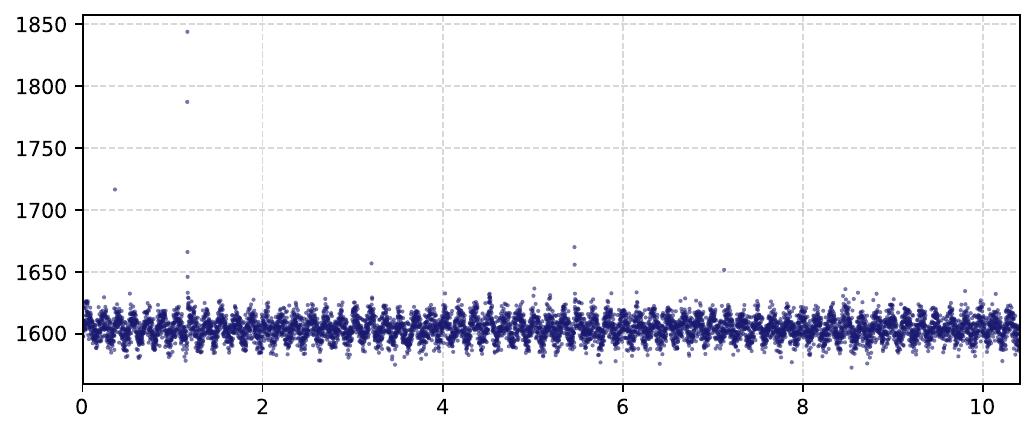}} \\[-0.1cm]
        
        \textbf{TIC 089257479} & \raisebox{-0.5\height}{\includegraphics[width=0.65\columnwidth]{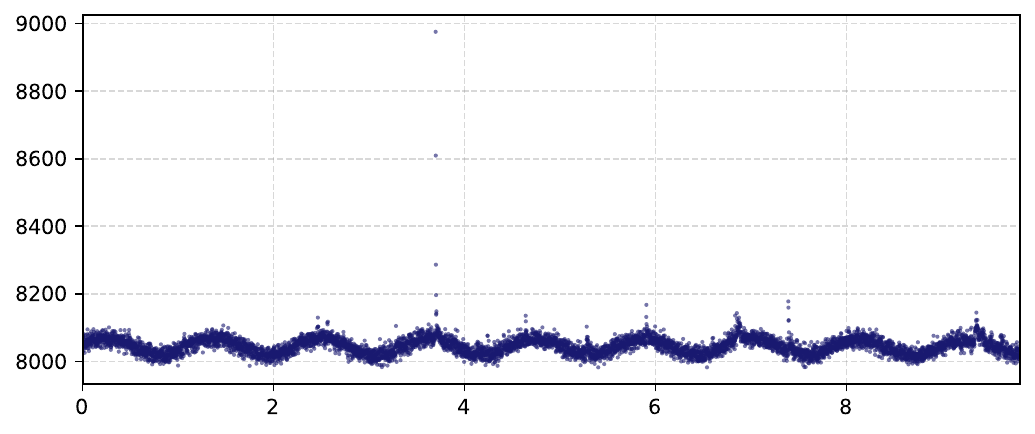}} \\[-0.1cm]
        
        \textbf{TIC 234526939} & \raisebox{-0.5\height}{\includegraphics[width=0.65\columnwidth]{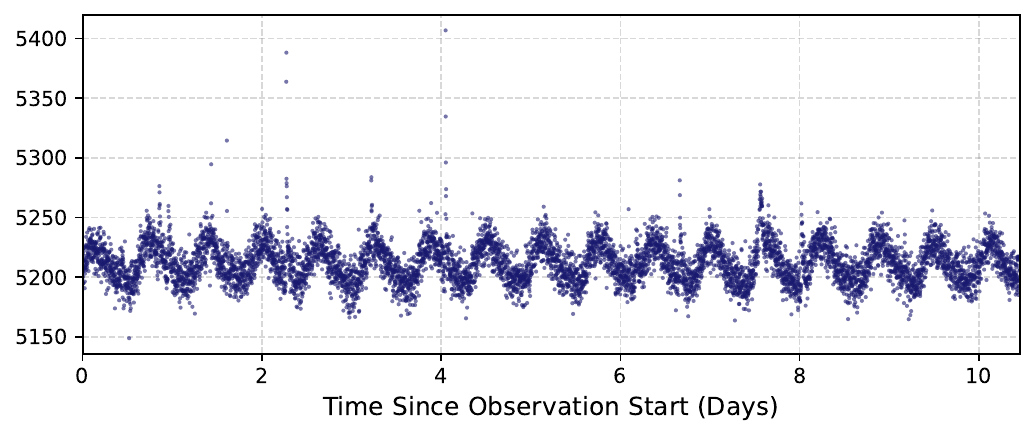}} \\
        
    \end{tabular}

    \includegraphics[width=0.5\columnwidth]{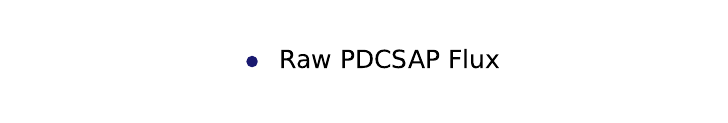}
    
    \vspace{-0.25cm} 
        
    \caption{Brightness over time from three distinct M dwarf stars. The left column identifies the object, and the plotted points represent the PDCSAP flux for each target.}
    \label{fig:raw_stars}
\end{figure}

These target stars were purposefully selected because their underlying continuous background emissions exhibit a highly diverse array of morphological structures (see Figure \ref{fig:raw_stars}). This structural diversity effectively spans the characteristic range of GP trends typically modeled by the exact \textit{Celerite} kernel, thereby providing a comprehensive and challenging testbed to evaluate the generalizability, accuracy, and computational efficiency of the VAE as a surrogate framework. Furthermore, the time series for these specific targets feature localized temporal windows containing a rich variety of distinct stellar flare events, ranging from isolated bursts to dense, complex clusters of flares. These specific flaring intervals are critical for validating the additive capability of the joint methodology. By successfully approximating and detrending the continuous stellar time series via the VAE surrogate, we isolate the remaining stochastic variation, thereby enabling the hidden Markov model (HMM) component to accurately identify, classify, and segment the flare events without conflating them with the background rotation.

\section{Methodology}
\label{sec:methodology}

\subsection{Gaussian Processes}
A Gaussian Process (GP) is a non-parametric probabilistic model that defines a distribution over functions. It generalizes the multivariate normal distribution to infinite dimensions, completely specified by a mean function, $m(t)$, and a covariance kernel, $k(t, t')$ over time. Specifically, the joint distribution for any finite collection of random variables $\{X_t\}$ for $t \subset T \in \mathbb{R}$ can be expressed as $p(X_t ) = \mathcal{N}(m(t), k(t, t'))$.

The choice of the kernel function $k(t, t')$, i.e.\ the covariance function, encodes our prior beliefs about the underlying structure and correlation of the data and can be adapted to varied and complex data sets. Common kernels chosen in practice are the squared exponential, Matérn, and periodic covariance functions \citep{rasmussen2006gaussian}. 
This flexibility makes GPs a popular choice for modeling complex, convoluted background trends in time-series analysis and is commonly used for modeling time series data from stars.

\subsection{The \textit{Celerite} Kernel for Stellar Rotation}

For many physical systems, such as stellar light curves, we require a kernel that captures quasi-periodicity: a signal with a dominant frequency that evolves or decays over time. Standard periodic kernels, even when multiplied by decay terms, scale cubically $\mathcal{O}(N^3)$ and become computationally prohibitive for large datasets. To model these physical variations efficiently at a lower computation cost, the \textit{Celerite} kernel was developed \citep{foreman2017fast}. 

The \textit{Celerite} kernel is defined via its power spectral density (PSD). The PSD describes how the variance (or power) of a time series is distributed across different frequencies. That is, if a PSD can be approximated as a specific rational function, its corresponding time-domain covariance kernel is exactly a sum of complex exponentials. We define this as the general underlying \textit{Celerite} kernel:
\begin{equation}
    k(\tau) = \sum_{j=1}^J a_j e^{-c_j \tau}
\end{equation}
where $\tau = |t_i - t_j|$ is the absolute time difference, and $a_j, c_j$ are complex coefficients. The specific structure of this kernel---a sum of exponentials---transforms the resulting covariance matrix into a \textit{semiseparable} matrix. Because of this semiseparable structure, the correlation between any two time points can be expressed as a product of terms. This factorization allows matrix operations, such as inversion and determinant evaluation, to be executed sequentially, which reduces the computational complexity from the standard $\mathcal{O}(N^3)$ to $\mathcal{O}(N)$ \citep{ambikasaran2015fast}. Thus, \textit{Celerite} is not a single kernel, but rather a flexible family of kernels that can take multiple functional derivative forms depending on the physical system being modeled.

A core building block in the \textit{Celerite} family is the stochastically driven damped, simple harmonic oscillator (SHO) \citep{foreman2017fast}. A damped SHO represents a system that naturally oscillates at a fundamental frequency but is subject to random perturbations and energy dissipation (damping) over time. This acts as a direct physical analogue for a rotating star with finite-lifetime surface features (e.g., stellar spots) \citep{angus2018inferring}. The PSD of a single, damped SHO is given by:
\begin{equation}
    S(\omega) = \sqrt{\frac{2}{\pi}} \frac{S_0 \omega_0^4}{(\omega^2 - \omega_0^2)^2 + \omega_0^2 \omega^2 / Q^2}
\end{equation}
where $S_0$ is proportional to the power, $\omega_0$ is the undamped resonant frequency, and $Q$ is the quality factor determining the damping. As derived by the \textit{Celerite} algorithm, the corresponding covariance kernel in the time domain physically translates to:
\begin{equation}
    k_{\text{SHO}}(\tau) = S_0 \omega_0 Q \exp\left(-\frac{\omega_0 \tau}{2Q}\right) \left[ \cos(\omega_d \tau) + \frac{\omega_0}{2Q\omega_d} \sin(\omega_d \tau) \right]
\end{equation}
where $\omega_d = \omega_0 \sqrt{1 - \frac{1}{4Q^2}}$ is the damped frequency.

While a single SHO captures a fundamental frequency, many complex systems---such as starspot distributions appearing on opposite stellar hemispheres---introduce strong periodic signals at the first harmonic (half the primary period). To robustly model this without losing physical consistency, we utilize a double harmonic oscillator. 
This is constructed by summing two SHO terms ($k(\tau) = k_{\text{SHO}_1}(\tau) + k_{\text{SHO}_2}(\tau)$). To maintain physical cohesion, the parameters of these two oscillators are algebraically coupled using a base set of parameters $\{\tau_1, \tau_2, \tau_3, \tau_4, \tau_5\}$, given by the primary oscillator (fundamental) and the secondary oscillator (first harmonic):
\begin{equation*}
    \begin{aligned}
        Q_1 &= 0.5 + \tau_3 + \tau_4 \\
        \omega_1 &= \frac{4\pi Q_1}{\tau_2 \sqrt{4Q_1^2 - 1}} \\
        S_1 &= \frac{\tau_1}{\omega_1 Q_1}
    \end{aligned}
    \qquad \text{and} \qquad
    \begin{aligned}
        Q_2 &= 0.5 + \tau_3 \\
        \omega_2 &= \frac{8\pi Q_2}{\tau_2 \sqrt{4Q_2^2 - 1}} \\
        S_2 &= \frac{\tau_5 \tau_1}{\omega_2 Q_2}.
    \end{aligned}
\end{equation*}
In this coupled structure, $\tau_2$ serves as the base period of the system. The mathematical formulation forces $\omega_2$ to operate at exactly twice the fundamental frequency of $\omega_1$.
The parameters $\tau_3$ and $\tau_4$ control the quality factors (damping timescales), while $\tau_1$ and $\tau_5$ dictate the base amplitude and the fractional amplitude of the harmonic, respectively.


\subsection{Approximation of the Gaussian Process via Variational Autoencoders}
\label{sec:vaeandgp}


While the computational cost of the \textit{Celerite} kernel is $\mathcal{O}(N)$, fitting \textit{Celerite} to larger time series ($>2000$ observations) may require hours, and sometimes days of computation time, especially in a Bayesian framework. In order to accelerate Bayesian inference, we propose a surrogate model to approximate \textit{Celerite} using a deep generative model, described next. 

\subsubsection{Neural Networks as Function Approximators}
To make the model computationally tractable during iterative Bayesian sampling, we employ a variational autoencoder (VAE), which uses a neural network architecture to approximate the background trend in the time series. 
A neural network is a parameterized function that maps inputs to outputs via a sequence of linear transformations and non-linear activations. These networks act as universal function approximators. 
To achieve this, we introduce a low-dimensional vector $z \in \mathbb{R}^L$, known as the latent variable. The space containing these vectors acts as a compressed, information-dense representation of the time series, useful here because it captures the fundamental physical properties of the data itself, such as rotation phase and spot amplitude in our stellar time series example.
The neural network's role is to learn the complex mapping from this simple latent space $z$ to the highly structured output of the target kernel.

\subsubsection{The VAE Architecture}
We adopt the VAE architecture from the PriorVAE framework \citep{semenova2022priorvae}. It consists of two coupled networks, the encoder $\mathcal{E}_\gamma(x)$ and the decoder $\mathcal{D}_{\psi}(z)$. The encoder $\mathcal{E}_\gamma(x)$ maps the input time series $x \in \mathbb{R}^N$ to a variational posterior distribution in the latent space,
\begin{equation}
\begin{aligned}
        (\mu_\gamma(x), \sigma^2_\gamma(x)) &= \mathcal{E}_\gamma(x) \\
        z|x &\sim \mathcal{N}(\mu_\gamma(x), \sigma^2_\gamma(x)I_L).
\end{aligned}
\end{equation}
 The decoder $\mathcal{D}_\psi(z)$ maps the latent vector $z$ back to the observation space to reconstruct the function,
    \begin{equation}
        \hat{x}|z \sim \mathcal{D}_\psi(z), \quad z \sim \mathcal{N}(0, I_L).
    \end{equation}
The objective of the VAE is to maximize the evidence lower bound (ELBO):
\begin{equation}
    \mathcal{L}(\psi, \gamma; x) = \mathbb{E}_{q_{\gamma}(z|x)}[\log p_{\psi}(x|z)] - D_{KL}(q_{\gamma}(z|x)||p(z))
\end{equation}

The reconstruction term $\mathbb{E}_{q_{\gamma}(z|x)}[\log p_{\psi}(x|z)]$ encourages the decoder to accurately reconstruct input samples drawn from the target \textit{GP prior} (e.g., the synthetic light curves used as the training dataset), while the regularization term $D_{KL}$ is the Kullback-Leibler divergence between the approximate posterior $q_{\gamma}(z|x)$ and the \textit{latent space prior $p(z)$}. We choose a standard normal prior $p(z) = \mathcal{N}(0, I_L)$ for the latent variables to ensure the space is well-regularized.

\begin{figure}[b!]
    \centering
    \includegraphics[width=0.8\textwidth]{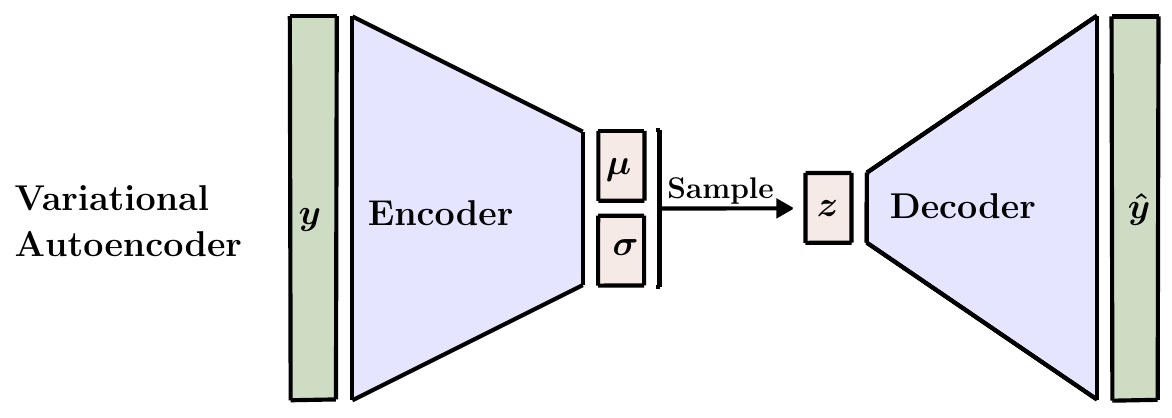} 
    \caption{Schematic representation of the overarching Variational Autoencoder (VAE) architecture. Diagram adapted from \citep{semenova2023priorcvae}}
    \label{fig:vae_architecture}
\end{figure}

By training the VAE on a large dataset generated from simulated light curves from the \textit{Celerite} kernel, the network learns to decouple the generative process into distinct training and inference phases:

\paragraph{The Training Phase}
During training, the encoder observes the simulated GP data $x_{sim}$ and maps it to the approximate posterior distribution. The KL-divergence term ensures this learned posterior structurally aligns with the standard normal prior:
\begin{equation*}
    q_\gamma(z | x_{sim}) \approx \mathcal{N}(0, I_L)
\end{equation*}
Concurrently, the decoder network's parameters $\psi$---which consist of the trainable weights and biases of its hidden layers---are optimized to map these structured latent vectors back into high-fidelity time series reconstructions. The optimal network parameters are found by maximizing the ELBO:
 \begin{equation*}
    \psi^*, \gamma^* = \underset{\psi, \gamma}{\mathrm{argmax}} \,\, \mathcal{L}(\psi, \gamma; x_{sim})
\end{equation*}
Through this optimization, the decoder parameters learn the exact covariance structure of the GP.

\paragraph{The Inference Phase (Surrogate Model)}
Once the VAE is fully trained, the encoder's job is complete, and it is no longer needed. 
For our downstream Bayesian inference, the pre-trained decoder acts as our stand-alone surrogate model. We can now generate new, valid realizations of the continuous trend (e.g., stellar rotation) simply by sampling $z$ from the standard normal prior and passing it through the optimized weights of the decoder:
\begin{equation*}
\begin{aligned}
    z &\sim \mathcal{N}(0, I_L) \\
    f_{trend} &= \mathcal{D}_{\psi^*}(z)
\end{aligned}
\end{equation*}
This sequential strategy allows us to completely bypass the expensive GP covariance matrix operations, providing a highly efficient, deterministic method for evaluating the background trend in any subsequent Bayesian model.

\subsubsection{Architectural Hyperparameters: Latent and Hidden Dimensions}
\label{sec:methodology_hyperparameters}
The computational viability and predictive performance of the surrogate model depends on two hyperparameters: the latent dimension ($L$) and the hidden dimension ($H$). Properly setting these architectural parameters prior to maximizing the ELBO is essential to balance data size, computational efficiency, and model generalization. The latent dimension $L$ defines the size of the latent vector $z \in \mathbb{R}^L$, and acts as the primary information bottleneck. It dictates the number of independent variables utilized to compress and describe the underlying physical state of the system (e.g., phase, amplitude, and spot evolution). The hidden dimension $H$ dictates the width (number of neurons) of the dense hidden layers comprising the encoder and decoder neural networks themselves. It determines the mathematical capacity of the network to approximate the highly non-linear, quasi-periodic dependencies of the target kernel.

The algebraic scaling of the network illustrates precisely why selecting these dimensions is a critical prerequisite. Consider an input time series of size $N$. During the forward pass to evaluate the ELBO, the encoder linearly transforms the $N$-dimensional data into the hidden space, applies a non-linear activation function $\sigma$, and maps the result to the $L$-dimensional latent space. The core mathematical operations at each layer are matrix multiplications:
\begin{align*}
    h_{layer} &= \sigma(W_{in} x + b_{in}) \\
    z_{\mu}, z_{\sigma} &= W_{latent} h_{layer} + b_{latent}
\end{align*}
where $h_{layer} \in \mathbb{R}^H$ is the intermediate hidden representation, and $z_{\mu}, z_{\sigma} \in \mathbb{R}^L$ define the mean and standard deviation vectors of the approximate posterior distribution. Here, $b_{in}$ and $b_{latent}$ represent the bias vectors, which shift the output of the linear transformation to increase the model's flexibility in fitting the data. The function $\sigma$ represents the non-linear activation function (e.g., ReLU or ELU). 

The initial weight matrix $W_{in}$ has dimensions $H \times N$, requiring $\mathcal{O}(H \cdot N)$ floating-point operations. The subsequent projection matrix $W_{latent}$ has dimensions $L \times H$, requiring $\mathcal{O}(L \cdot H)$ operations. After sampling $z$ from the latent distribution, the decoder performs the exact inverse spatial expansion, mapping $L \rightarrow H \rightarrow N$ algebraically:
\begin{align*}
    h_{dec} &= \sigma(W_{dec\_in} z + b_{dec\_in}) \\
    \hat{x} &= W_{out} h_{dec} + b_{out}
\end{align*}
where $W_{dec\_in}$ has dimensions $H \times L$ and $W_{out}$ has dimensions $N \times H$, reconstructing the high-dimensional continuous signal $\hat{x} \in \mathbb{R}^N$. Note that the activation function $\sigma$ applied in the decoder is still a \textit{forward} non-linear activation (typically the same function used in the encoder), not an inverse function. This enables the network to progressively construct the complex non-linearities of the target output. Consequently, the total number of trainable parameters and the memory footprint per optimization step scale proportionally with $(N \cdot H) + (H \cdot L)$. This imposes practical trade-offs:
\begin{enumerate}
    \item \textbf{Computational Limits:} If $H$ and $L$ are excessively large relative to a massive input size $N$, the resulting weight matrices will require large amounts of memory, potentially exhausting available system memory or rendering the ELBO maximization impractically slow.
    \item \textbf{Overfitting:} If the dimensions are too large, rather than learning the generalized underlying physics of the GP kernel, the model will simply memorize the training set. While the training error will approach zero, the surrogate model will overfit, meaning its predictive metrics on new, unseen data trajectories will rapidly degrade, ultimately undermining the downstream Bayesian inference.
    \item \textbf{Underfitting:} Conversely, if $H$ and $L$ are too compact, the information bottleneck is too severe. The network will lack the capacity to capture the true covariance structure, leading to poor reconstructions and loss of vital signal information.
\end{enumerate}
Therefore, identifying the optimal balance for $H$ and $L$ relative to the input data size $N$ is an important step to ensure the VAE successfully compresses the physical signal without memorizing noise, resulting in a fast, robust, and generalizable surrogate model.

\subsubsection{Bayesian Suitability: Geometry of the Posterior}
The VAE framework offers a distinct, fundamental advantage for Bayesian inference due to the geometry of the posterior. In standard GP inference, Bayesian analysis requires directly sampling the physical hyperparameters (e.g., Period, Amplitude, Damping). These parameters are often highly correlated (e.g., amplitude and damping often trade off against one another), creating a complex, "banana-shaped" posterior geometry that is notoriously difficult for MCMC samplers to explore efficiently. 

In the VAE surrogate framework, instead of sampling these correlated physical hyperparameters, the MCMC algorithm simply samples the components of the latent vector $z = [z_1, \dots, z_L]^T$. By design---due to the KL-divergence regularization term maximized during the VAE training---the prior on $z$ is a standard normal distribution $\mathcal{N}(0, I_L)$ with uncorrelated components. This mathematical restructuring of the parameter space ensures the posterior geometry is structurally simpler and closer to a symmetric, isotropic sphere. This simplified geometry is exceptionally well-suited for advanced samplers like Hamiltonian Monte Carlo (HMC) or the No-U-Turn Sampler (NUTS).


\newpage

\subsection{Additive Models and Compositional Inference}

\subsubsection{The Additive Model Setup}
In many scientific domains, the GP is only one component of a larger model. 
In this project, we propose an additive model that is the combination of a GP, namely the \textit{Celerite} kernel to describe stellar rotation, and an HMM to describe the flaring process (GP+HMM). We assume the observed data at time $t$, denoted $Y_t$, is conditionally dependent on a discrete latent state sequence $S_t$ which we interpret as the star's flaring states. The observation is modeled as the sum of two components:
\begin{equation}
    Y_t | S_t = f_t + R_t | S_t
\end{equation}
Where the trend $f_t$ is modeled via \textit{Celerite} and is independent of the flaring process and the residual $R_t|S_t$ is the state-dependent flaring process which describes whether the star is in a quiet, flaring, or decaying state.  
The challenge is that the HMM models the residuals $R_t = Y_t - f_t$ and since the true continuous trend $f_t$ is unknown, we must jointly infer it alongside the parameters of the HMM.

\subsubsection{The Standard GP+HMM}
In the standard framework \citep{esquivel2024detecting}, the trend $f$ is a random variable drawn from the GP prior. To compute the full marginal likelihood of the observations $Y$, we must integrate over the continuous GP function space and sum over all possible discrete state sequences $S$:
\begin{equation*}
    \mathcal{L}_{GP+HMM}(Y) = \int \left( \sum_{S} P(Y | f, S) P(S) \right) P(f | K) \, df
\end{equation*}
Here, $K$ represents the dense $N \times N$ covariance matrix generated by the GP kernel (e.g., the \textit{Celerite} kernel), which captures the temporal correlation of the trend. The evaluation couples the residuals with this covariance structure through the GP prior:
\begin{equation*}
    P(f | K) \propto \frac{1}{\sqrt{|K|}} \exp\left(-\frac{1}{2} (Y - R)^T K^{-1} (Y - R)\right)
\end{equation*}
This coupling means that evaluating the joint likelihood requires evaluating the covariance matrix $K^{-1}$ repeatedly during inference, making the MCMC loop scaling highly dependent on the cost of matrix inversion.

\subsubsection{The Proposed VAE+HMM}

In our proposed framework, we substitute the stochastic GP prior with the deterministic VAE decoder. The continuous trend is no longer a random variable drawn from $K$, but a direct, deterministic transformation of the low-dimensional latent vector $z$: $f = \mathcal{D}_\psi(z)$. Therefore, the intractable integral over the function space is entirely replaced by inference over the low-dimensional latent space. Crucially, we treat $z$ as an active, inferable parameter. We perform joint Bayesian inference to simultaneously estimate the posterior distribution of the latent vector $z$ alongside the parameters of the HMM. During this joint estimation, the conditional likelihood of the data given $z$ simplifies dramatically to evaluating the HMM strictly on the deterministic residuals:
\begin{equation*}
    \mathcal{L}_{VAE+HMM}(Y | z) = \sum_{S} P(Y | \mathcal{D}_\psi(z), S) P(S)
\end{equation*}

Because the trend is deterministic for any proposed $z$ during the sampling loop, there is no covariance matrix to invert. The target log-likelihood for our joint inference collapses completely to standard HMM operations on the residual vector $R_t = Y_t - \mathcal{D}_\psi(z)_t$:
\begin{equation*}
    \mathcal{L}_{VAE+HMM} = \sum_{t=1}^N \log P_{\text{HMM}}(R_t | S_t)
\end{equation*}
As demonstrated mathematically, the substitution of the dense covariance matrix---and thus the explicit GP itself---within any larger hierarchical system can be performed in a similar manner. This approach yields a generalized likelihood reduction, resulting in significant computational savings across a wide variety of hybrid mixture architectures. Crucially, because this substitution completely preserves the fundamental probabilistic structure of the HMM \citep{zucchini2016hidden}, exact inference on the deterministic residuals remains straightforward. We utilize the forward algorithm to efficiently compute the marginal likelihood required for MCMC sampling, and subsequently apply the Viterbi algorithm to decode the optimal sequence of hidden states upon convergence. Full mathematical details of these procedures are provided in the HMM Inference: Forward and Viterbi Algorithms supplementary material section.

\subsubsection{Full Bayesian Inference Algorithm}
The specific mathematical structure of the HMM implemented in this project---including the state definitions, allowable transition paths, and non-standard emission distributions---is directly adapted from the GP+HMM framework developed by \citet{esquivel2024detecting}.

We perform joint Bayesian inference on the latent variable $z$ (which generates the VAE trend) and the HMM parameters using the No-U-Turn Sampler (NUTS) \citep{hoffman2014nuts}, implemented via the NumPyro probabilistic programming library in Python \citep{phan2019composable}. To illustrate how the VAE architecture completely replaces the GP within the additive model while leaving the HMM probabilistic logic fully intact, we detail the complete end-to-end methodology in Algorithm \ref{alg:vae_hmm}. This pipeline combines descriptive logic with the exact algebraic operations required to execute it, accommodating real-world data issues by incorporating a binary mask $m_t \in \{0, 1\}$ for missing observations.

\begin{algorithm}[htbp]
\caption{End-to-End VAE-HMM Inference Pipeline}
\label{alg:vae_hmm}
\textbf{Phase I: Prior Surrogate Training}
\begin{enumerate}
    \item \textbf{Generate Simulated Data:} Sample a training dataset of $M$ continuous trajectories, each of length $N$, from the exact GP prior defined by the target kernel $K$: 
    \begin{equation*}
        X_{sim}^{(i)} \sim \mathcal{N}(0, K)
    \end{equation*}
    \item \textbf{Maximize ELBO:} Train the VAE architecture to map $X_{sim}$ through the latent space $z$. 
    \item \textbf{Extract Surrogate:} Discard the encoder and retain the optimized decoder weights $\psi^*$ as the deterministic generative function $\mathcal{D}_{\psi^*}(\cdot)$.
\end{enumerate}

\textbf{Phase II: Joint MCMC Inference (Bypassing Covariance Matrix Operations)}\\
\textit{For each MCMC iteration $k$:}
\begin{enumerate}
    \setcounter{enumi}{3}
    \item \textbf{Sample Priors:} Draw continuous variables for the latent space and HMM parameters from their respective prior distributions. $\mu, \sigma,$ and $\alpha$ represent fixed prior hyperparameters. To enforce physically logical sequences, the transition matrix $\Theta^{(k)}$ forbids specific state changes (e.g., forcing a Quiet state before a Decay state is impossible):
    \begin{align*}
        z^{(k)} &\sim \mathcal{N}(0, I_L), \quad \sigma_{noise}^{(k)} \sim \text{HalfNormal}(1.0) \\
        \log(\lambda_f^{(k)}) &\sim \mathcal{N}(\mu_{f}, \sigma_{f}), \quad \text{logit}(\lambda_d^{(k)}) \sim \mathcal{N}(\mu_{d}, \sigma_{d}) \\
        \Theta_Q, \Theta_F, \Theta_D &\sim \text{Dirichlet}(\alpha) \implies \Theta^{(k)} = \begin{bmatrix} \theta_{QQ} & \theta_{QF} & 0 \\ 0 & \theta_{FF} & \theta_{FD} \\ \theta_{DQ} & \theta_{DF} & \theta_{DD} \end{bmatrix}
    \end{align*}
    \item \textbf{Deterministic Transform (The VAE Pass):} Map the sampled $z$ through the fixed surrogate to generate the continuous trend. \\
    \begin{equation*}
        f_{trend}^{(k)} = \mathcal{D}_{\psi^*}(z^{(k)})
    \end{equation*}
    \item \textbf{Residual Computation:} Isolate the state-dependent signals: 
    \begin{equation*}
        R_t^{(k)} = Y_t - f_{trend, t}^{(k)}
    \end{equation*}
    \item \textbf{Likelihood Evaluation (Forward Algorithm):} We define $E_t(j) \equiv \log P(R_t | S_t = j)$ as the log-emission probability of the residual under state $j$. This is evaluated subject to the observed data mask $m_t$. If an observation is missing ($m_t = 0$), its log-emission $E_t$ is safely forced to $0.0$.
    \begin{itemize}
        \item \textit{Calculate Log-Emissions $E_t$ for each state:}
        \begin{align*}
            E_t(Q) &= m_t \cdot \log \mathcal{N}(R_t^{(k)} | 0, \sigma_{noise}^{(k)}) \\
            E_t(F) &= m_t \cdot \log \text{ExpModNormal}(R_t^{(k)} | R_{t-1}^{(k)}, \sigma_{noise}^{(k)}, \lambda_f^{(k)}) \\
            E_t(D) &= m_t \cdot \log \mathcal{N}(R_t^{(k)} | \lambda_d^{(k)} R_{t-1}^{(k)}, \sigma_{noise}^{(k)})
        \end{align*}
        \item \textit{Recursive LogSumExp Update for Forward Probabilities $\alpha_t(j)$:}
        \begin{equation*}
            \alpha_t(j) = \log \left( \sum_{i \in \{Q,F,D\}} \exp \left( \alpha_{t-1}(i) + \log \Theta_{ij}^{(k)} \right) \right) + E_t(j)
        \end{equation*}
        \item \textit{Total Log-Likelihood:} $\log \mathcal{L}^{(k)} = \text{LogSumExp}(\alpha_N)$
    \end{itemize}
    \item \textbf{HMC Step:} Compute gradients of the log-posterior, $\log P(\theta | Y) \propto \log \mathcal{L}^{(k)} + \log P(\theta)$, with respect to all sampled variables to propose the next step via Hamiltonian dynamics.
\end{enumerate}

\textbf{Phase III: Optimal State Decoding}
\begin{enumerate}
    \setcounter{enumi}{8}
    \item \textbf{Sample-Wise Viterbi:} For each retained MCMC sample $s \in \{1, \dots, S\}$, apply the Viterbi Algorithm using its specific parameters $\Phi^{(s)}$ and residuals $R^{(s)}$:
    \begin{align*}
        V_t^{(s)}(j) &= \max_i \left( V_{t-1}^{(s)}(i) + \log \Theta_{ij}^{(s)} \right) + E_t^{(s)}(j) \\
        \text{BackPtr}_{t, j}^{(s)} &= \text{argmax}_i \left( V_{t-1}^{(s)}(i) + \log \Theta_{ij}^{(s)} \right)
    \end{align*}
\end{enumerate}
\end{algorithm}


\section{Simulation Study}
\label{sec:simulation_study}

\subsection{Practical Application of the VAE and Data Normalization}
While the theoretical formulation of the VAE provides a robust mathematical surrogate for GPs (as detailed in Section \ref{sec:vaeandgp}), its practical implementation requires amplitude normalization. In standard deep learning, neural networks struggle to optimize when input features exhibit extreme, unconstrained variability. In our context, raw stellar brightness data can vary wildly in amplitude. If passed directly into the VAE, the optimization of the ELBO function becomes numerically unstable, often resulting in exploding gradients.

To ensure stability, we force the simulated GP trajectories to match the standardized scale of the real astronomical test data (approximately $-3$ to $+3$) prior to network optimization. However, standard normalization (using the mean and standard deviation) is highly sensitive to outliers, such as the flares we aim to detect. Therefore, we employ a robust normalization pipeline based on the median and the median absolute deviation (MAD).

For each simulated light curve $x$, we compute the median flux across the entire time series and the absolute deviations from that median. The MAD is then scaled by a constant factor of $1.4826$ to make it asymptotically consistent with the standard deviation of a normal distribution. The normalized flux is thus computed as:
\[
    x_{norm} = \frac{x - \text{median}(x)}{1.4826 \cdot \text{median}(|x - \text{median}(x)|)}
\]
Concurrently, the temporal x-axis is constrained via a standard min-max scaling to project the observation times onto a strictly $[0, N]$ interval. This dual-normalization ensures the VAE weights remain stable during ELBO maximization, focusing entirely on the structural morphology of the trend rather than arbitrary scaling factors.

\noindent \textbf{Scenario 1 (Smooth \& Simple)} \\
This scenario represents a smooth trend with minimal high-frequency oscillations. It effectively models a star with slow rotation and stable, long-lived starspots.

\begin{figure}[H]
    \centering
    \begin{minipage}{0.45\textwidth}
        \centering
        \includegraphics[width=\linewidth]{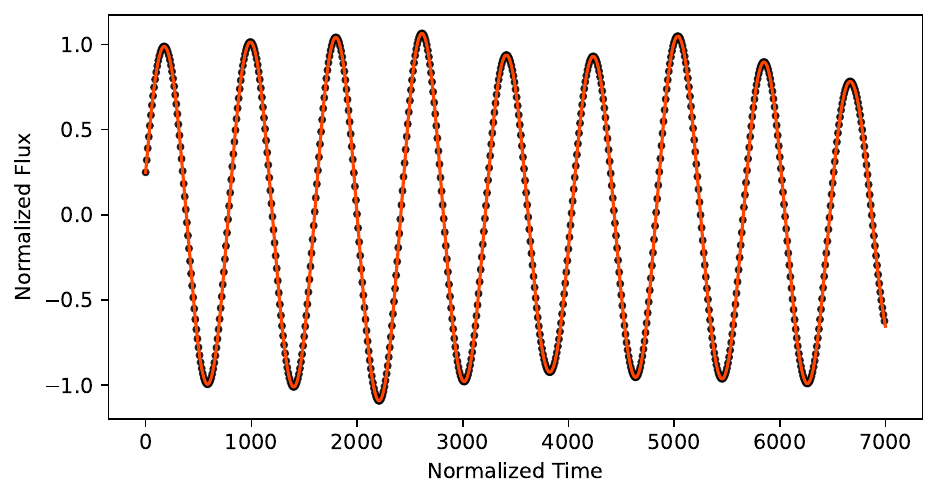} 

        \vspace{0.1cm} 

        \includegraphics[width=.55\textwidth]{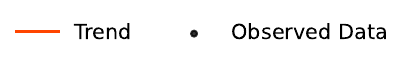} 
    
        \vspace{0.1cm} 
        
    \end{minipage}\hfill
    \begin{minipage}{0.25\textwidth}
        \centering
        \begin{tabular}{cc}
            \toprule
            \textbf{Parameter} & \textbf{Value} \\
            \midrule
            $Q_1$ & 5.337618 \\
            $S_1$ & 7.216781 \\
            $\omega_1$ & 1.02038 \\
            $Q_2$ & 3.186107 \\
            $S_2$ & 5.836664 \\
            $\omega_2$ & 1.718205 \\
            \bottomrule
        \end{tabular}
    \end{minipage}
    \caption{Scenario 1: Smooth and simple trend (left) alongside its exact generative \textit{Celerite} parameters (right).}
    \label{fig:scenario_1}
\end{figure}

\vspace{-0.35cm}

\noindent \textbf{Scenario 2 (Fast \& Noisy)} \\
This setup features a much faster periodicity with a higher density of oscillations. The rapid fluctuations create a visually ``noisy'' background, testing the network's ability to compress high-frequency variations without blurring the actual continuous signal.

\begin{figure}[H]
    \centering
    \begin{minipage}{0.45\textwidth}
        \centering
        \includegraphics[width=\linewidth]{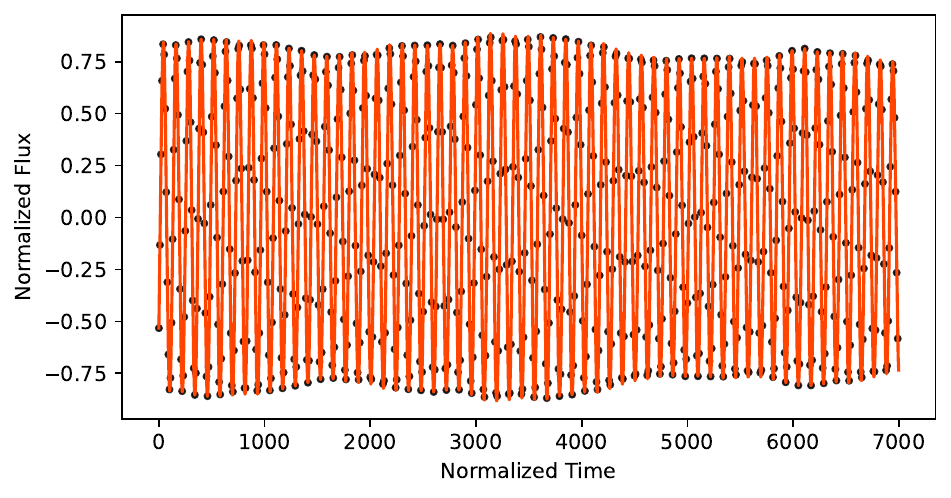} 

        \vspace{0.1cm} 

        \includegraphics[width=.55\textwidth]{celerite_simulations_legend_simulations.pdf} 
    
        \vspace{0.1cm} 

    \end{minipage}\hfill
    \begin{minipage}{0.25\textwidth}
        \centering
        \begin{tabular}{cc}
            \toprule
            \textbf{Parameter} & \textbf{Value} \\
            \midrule
            $Q_1$ & 6.543659 \\
            $S_1$ & -1.503824 \\
            $\omega_1$ & 2.932907 \\
            $Q_2$ & 6.227217 \\
            $S_2$ & -4.181246 \\
            $\omega_2$ & 3.626107 \\
            \bottomrule
        \end{tabular}
    \end{minipage}
    \caption{Scenario 2: Fast and noisy highly oscillatory trend (left) alongside its generative parameters (right).}
    \label{fig:scenario_2}
\end{figure}

\vspace{-0.35cm}

\noindent \textbf{Scenario 3 (Complex \& Amplitude Modulated)} \\
The final scenario introduces a highly complex morphological structure, driven by a dominant first harmonic in the \textit{Celerite} parameters. This generates a dynamic amplitude envelope and irregular, beating oscillatory patterns. 

\begin{figure}[H]
    \centering
    \begin{minipage}{0.45\textwidth}
        \centering
        \includegraphics[width=\linewidth]{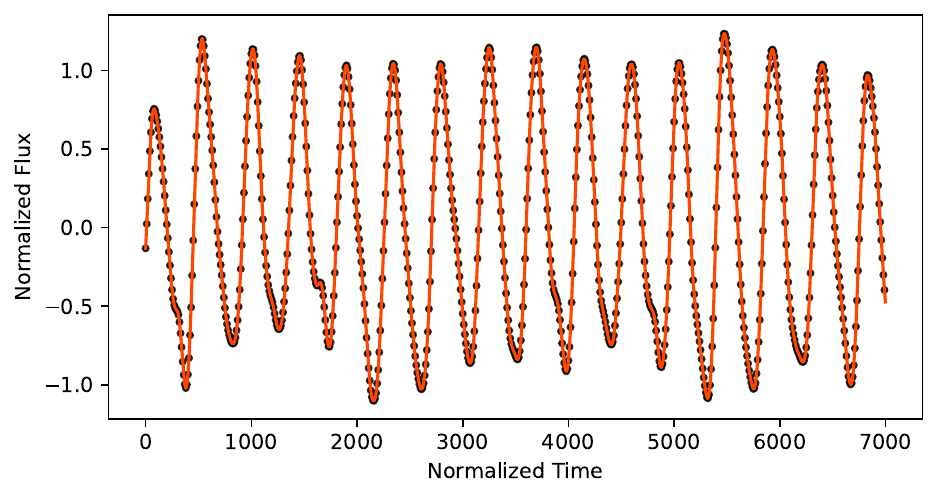} 

        \vspace{0.1cm} 

        \includegraphics[width=.55\textwidth]{celerite_simulations_legend_simulations.pdf} 
    
        \vspace{0.1cm} 

    \end{minipage}\hfill
    \begin{minipage}{0.25\textwidth}
        \centering
        \begin{tabular}{cc}
            \toprule
            \textbf{Parameter} & \textbf{Value} \\
            \midrule
            $Q_1$ & 5.277176 \\
            $S_1$ & 4.481018 \\
            $\omega_1$ & 1.937558 \\
            $Q_2$ & 2.35105 \\
            $S_2$ & 2.785322 \\
            $\omega_2$ & 2.640813 \\
            \bottomrule
        \end{tabular}
    \end{minipage}
    \caption{Scenario 3: Complex multi-peaked trend (left) alongside its generative parameters (right).}
    \label{fig:scenario_3}
\end{figure}

\subsection{Simulated \textit{Celerite} Scenarios and Data Scales}
To evaluate the surrogate model under realistic astronomical conditions, we based our three distinct simulation scenarios on real stellar brightness data (see Figures \ref{fig:scenario_1}, \ref{fig:scenario_2} and \ref{fig:scenario_3} for visual representations and accompanying descriptions of these scenarios). For each case, the example continuous trend shown in the subsequent figures was obtained directly by fitting the full GP+HMM framework \citep{esquivel2024detecting} to actual TESS observations. From these empirical fits, we extracted the mean generative \textit{Celerite} parameters ($Q_1, \omega_1, S_1, Q_2, S_2, \omega_2$). These exact parameters were subsequently used to generate the large synthetic GP datasets required to train our VAE. This empirical grounding ensures that our training simulations capture a diverse and physically accurate spectrum of true background stellar behavior. Additionally, for visual clarity in the figures, we plot only every 10th data point of these highly dense time-series.

Furthermore, another important aspect for the computational efficiency of the VAE surrogate is the size of the observation vectors. To mirror the reality of our target astronomical analyses, we evaluated each scenario across three distinct data lengths: $N = 2000$, $N = 5000$, and $N = 7000$. These specific sizes were chosen because they encompass the typical observational windows provided by TESS. For strict comparability across these scales, the underlying generative sequence remains identical within each scenario; the smaller datasets are exact temporal truncations of the largest sequence (i.e., the $N=5000$ dataset consists exactly of the first 5000 observations of the $N=7000$ trend).

\subsection{Hyperparameter Tuning}
As outlined in Section \ref{sec:methodology_hyperparameters}, the dimensions of the VAE dictate its capacity to approximate the GP kernel. While theoretical intuition suggests that expanding the network dimensions should strictly improve the approximation, practical Bayesian applications present a strict trade-off between geometric accuracy and computational cost.

A fundamental rule in VAE design is that the latent dimension ($L$) must be strictly less than or equal to the hidden dimension ($H$) \citep{kingma2013auto}. The latent space is designed to act as an information bottleneck, forcing the network to distill the complex temporal sequence into dense, fundamental generative features \citep{goodfellow2016deep}. If $L > H$, the bottleneck is inverted; the network mathematically loses the capacity to compress the data, resulting in sparse, unstructured mappings that severely degrade the model's predictive power. 

To map the trade-off between computational cost and approximation accuracy, we propose and test five distinct architectural tuning configurations, summarized in Table \ref{tab:architectures}.

\begin{table}[b!]
    \centering
    \caption{The five architectural tuning configurations evaluated in the simulation study, detailing their respective Hidden ($H$) and Latent ($L$) dimensions and estimated VRAM footprint.}
    \label{tab:architectures}
    \begin{tabular}{lccc}
        \toprule
        \textbf{Model Configuration} & \textbf{Hidden Dim ($H$)} & \textbf{Latent Dim ($L$)} & \textbf{Est. VRAM Usage} \\
        \midrule
        \textbf{1. Tiny}   & 128  & 32  & Very Low \\
        \textbf{2. Small}  & 256  & 64  & Low \\
        \textbf{3. Medium} & 1024 & 128 & Moderate \\
        \textbf{4. Large}  & 2048 & 256 & High \\
        \textbf{5. Ultra}  & 3072 & 512 & Very High \\
        \bottomrule
    \end{tabular}
\end{table}

\subsection{Performance and Efficiency Metrics}
To benchmark the five VAE configurations across the different scenarios and data sizes, we divide our analysis into two distinct categories: performance metrics and efficiency metrics (assessing computational speed and sampling geometry). For every experimental combination (scenario $\times$ data size $\times$ configuration), the MCMC algorithm was run for 2000 total iterations, comprising 1000 burn-in tuning steps and 1000 drawn posterior samples. To ensure a standardized evaluation of computational time, all inference tasks were executed on a 14-inch MacBook Pro equipped with an Apple M3 Pro chip and 18 GB of memory.

\subsubsection{Performance Metrics}
Model accuracy is quantified using the mean squared error (MSE) and mean absolute error (MAE) to measure the absolute deviation between the true GP continuous trend and the VAE surrogate reconstruction. Additionally, we calculate the coefficient of determination ($R^2$) to assess the proportion of the trend's variance successfully captured by the surrogate. While the following analysis focuses on these aggregated quantitative metrics, the complete visual validation---including individual VAE fits and residual diagnostic plots for all 45 simulated conditions---is provided in the Extended Simulation Diagnostics supplementary material section.

\begin{figure}[b!]
    \centering
    \begin{tabular}{lccc}
        & \textbf{Scenario 1} & \textbf{Scenario 2} & \textbf{Scenario 3} \\

        & Smooth \& Simple & Fast \& Noisy & Complex \& Amplitude Modulated \\        
        
        \raisebox{1.5cm}{\textbf{MSE}} & 
        \includegraphics[width=0.28\textwidth]{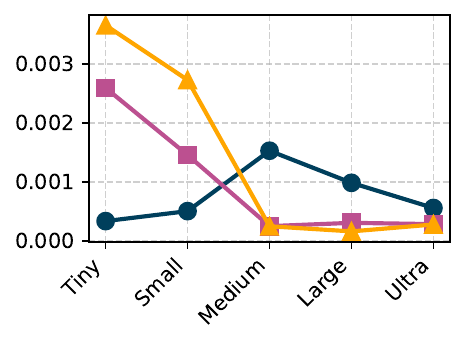} & 
        \includegraphics[width=0.28\textwidth]{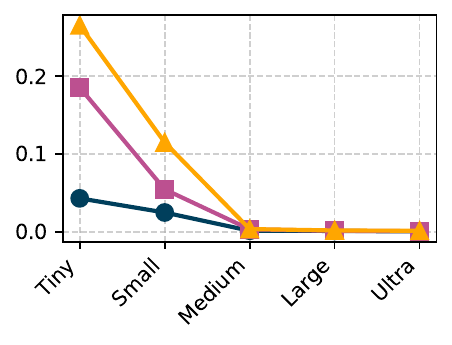} & 
        \includegraphics[width=0.28\textwidth]{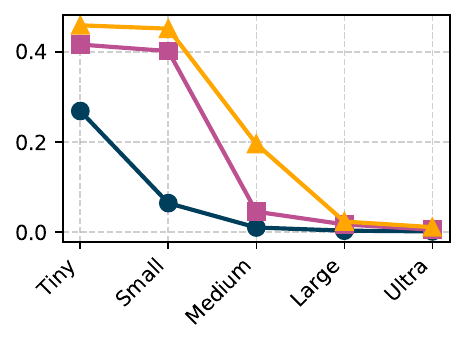} \\ 
        
        \raisebox{1.5cm}{\textbf{MAE}} & 
        \includegraphics[width=0.28\textwidth]{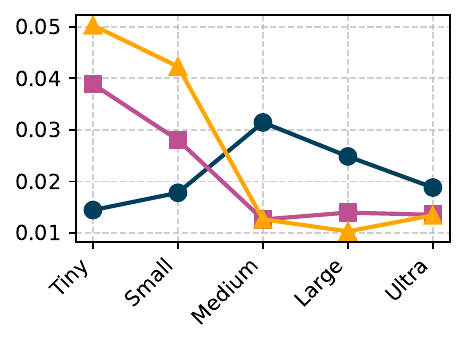} & 
        \includegraphics[width=0.28\textwidth]{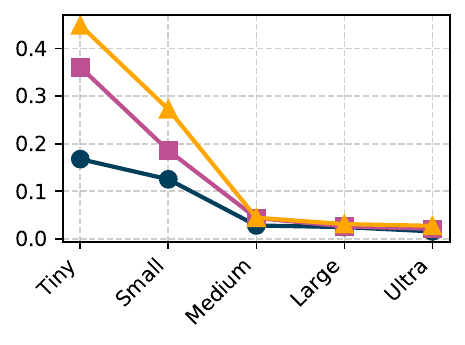} & 
        \includegraphics[width=0.28\textwidth]{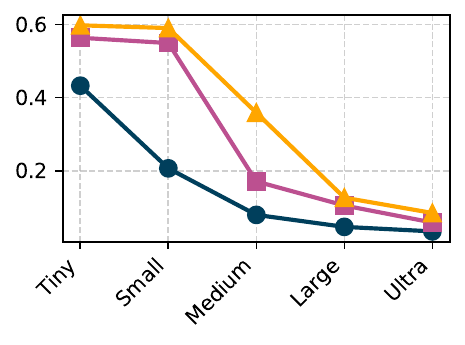} \\ 
        
        \raisebox{1.5cm}{\textbf{R\textsuperscript{2}}} & 
        \includegraphics[width=0.28\textwidth]{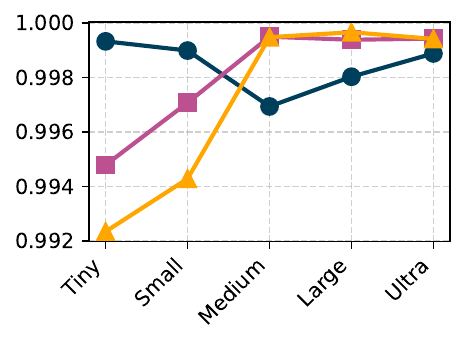} & 
        \includegraphics[width=0.28\textwidth]{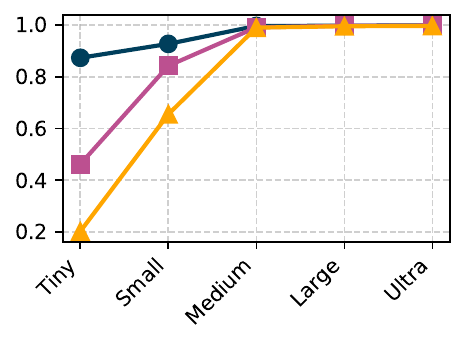} & 
        \includegraphics[width=0.28\textwidth]{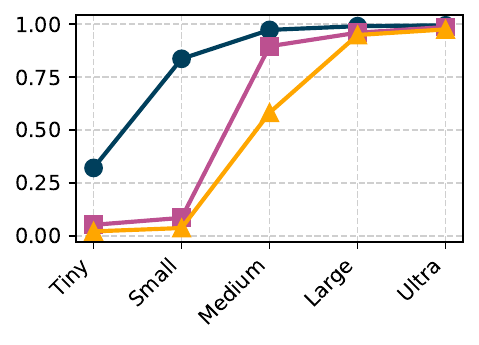} \\ 
    \end{tabular}

    \vspace{-0.3cm} 

    \includegraphics[width=0.4\textwidth]{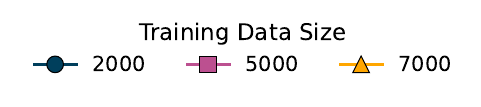} 

    \vspace{-0.5cm} 

    \caption{Performance metrics (MSE, MAE, and $R^2$) comparing the model's predictive accuracy across three distinct simulated scenarios. The shared legend for all subplots is provided below.}
    \label{fig:metrics_grid}
\end{figure}


The results in Figure \ref{fig:metrics_grid} highlight a clear spectrum of structural capability across the architectures. The \textit{Tiny} and \textit{Small} configurations generally underfit the data when faced with highly complex structures (Scenario 3) or massive data sizes ($N=7000$), yielding higher absolute errors and lower $R^2$ values. Conversely, there is a visible optimization point around the \textit{Medium} and \textit{Large} models, where the network capacity is sufficient to drive the error near zero and the $R^2$ toward 1.0. 

However, Figure \ref{fig:metrics_grid} also reveals the subtle dangers of capacity scaling combined with simple, limited data. In Scenario 1, the performance metrics for the smallest data size ($N=2000$, dark blue line) show a distinct ``hump'' or degradation in performance when moving from the \textit{Small} to the \textit{Medium} architecture. This behavior is symptomatic of the well-documented ``double descent'' phenomenon in modern machine learning \cite{belkin2019reconciling}. At this specific intermediate capacity---often referred to as the interpolation threshold---expanding the network dimensions on a smaller dataset initially leads to severe overfitting. The neural network uses its available capacity to precisely memorize localized noise rather than generalizing the highly predictable underlying trend, causing a dip in predictive accuracy. As network dimensions increase further into the \textit{Large} and \textit{Ultra} configurations, the model enters a massively over-parameterized regime, effectively smoothing out these memorized artifacts and allowing the test metrics to recover.

Ultimately, the choice of architecture depends heavily on the user's specific goals and computational constraints; we do not prescribe a single universally ``best'' architecture. If higher precision is required for a highly convoluted physical trend, a configuration like the \textit{Medium} or \textit{Large} model could be an appropriate option. However, if the underlying GP structure is relatively simple (like Scenario 1), the \textit{Tiny} and \textit{Small} architectures work well. As explicitly indicated by the extremely small scale of the y-axes in the Scenario 1 plots, the absolute difference in error between the smallest and largest networks is practically negligible. In such cases, or when operating under strict hardware constraints, smaller architectures remain viable and efficient options.

\subsubsection{Efficiency Metrics}
Because the primary motivation of substituting a GP with a VAE is computational acceleration, we monitor the inference time (in seconds). Finally, to validate the quality of the posterior geometry, we record the effective sample size (ESS) across all sampled parameters.

\begin{table}[b!]
    \centering
    \caption{Computational time (in seconds) and the mean effective sample size (ESS) calculated across all inferred model parameters for the five surrogate model architectures. Results are evaluated across three data sizes ($N$) and divided by the three simulation scenarios.}
    \label{tab:efficiency_metrics}
    \begin{tabular}{ll cccccc}
        \toprule
        & & \multicolumn{2}{c}{\textbf{$N = 2,000$}} & \multicolumn{2}{c}{\textbf{$N = 5,000$}} & \multicolumn{2}{c}{\textbf{$N = 7,000$}} \\
        \cmidrule(lr){3-4} \cmidrule(lr){5-6} \cmidrule(lr){7-8}
          & \textbf{Model} & \textbf{Time} & \textbf{ESS} & \textbf{Time} & \textbf{ESS} & \textbf{Time} & \textbf{ESS} \\
        \midrule
        
        \multirow{5}{*}{\textbf{Scenario 1}} 
        & 1. Tiny   & 28  & 196  & 30  & 1839 & 18  & 1977 \\
        & 2. Small  & 11  & 2107 & 36  & 2155 & 33  & 2067 \\
        & 3. Medium & 31  & 2110 & 87  & 3018 & 94  & 2721 \\
        & 4. Large  & 106 & 2203 & 211 & 1993 & 184 & 2836 \\
        & 5. Ultra  & 364 & 2029 & 489 & 2383 & 546 & 2660 \\
        \midrule
        
        \multirow{5}{*}{\textbf{Scenario 2}} 
        & 1. Tiny   & 5 & 2252 & 6 & 2381 & 30 & 2064 \\
        & 2. Small  & 7 & 2818 & 26 & 2145 & 23 & 2226 \\
        & 3. Medium & 37 & 1913 & 69 & 2585 & 228 & 2454 \\
        & 4. Large  & 155 & 2385 & 248 & 2441 & 269 & 2536 \\
        & 5. Ultra  & 392 & 1792 & 477 & 2534 & 418 & 2385 \\
        \midrule

        \multirow{5}{*}{\textbf{Scenario 3}} 
        & 1. Tiny   & 6 & 1869 & 7 & 2122 & 11 & 1300 \\
        & 2. Small  & 5 & 2537 & 15 & 1974 & 48 & 490 \\
        & 3. Medium & 26 & 2141 & 57 & 2941 & 87 & 2911 \\
        & 4. Large  & 105 & 2476 & 148 & 2827 & 209 & 3079 \\
        & 5. Ultra  & 274 & 1787 & 344 & 2960 & 428 & 2589 \\
        \bottomrule
    \end{tabular}
\end{table}

The efficiency metrics reiterate the need to carefully balance network dimensions. Table \ref{tab:efficiency_metrics} provides explicit empirical evidence of the computational penalty incurred by over-parameterization. While the \textit{Ultra} model possesses the highest theoretical capacity, moving from the \textit{Medium} to the \textit{Ultra} architecture results in a massive inflation of inference time (e.g., jumping from 87 to 489 seconds in Scenario 1, $N=5000$) with rapidly diminishing structural returns. Furthermore, the latent space $L=512$ of the \textit{Ultra} model creates a high-dimensional geometry that is significantly harder for the NUTS sampler to explore efficiently, explicitly reflected in a persistent drop in the average ESS. Therefore, selecting a configuration that matches---but does not exceed---the complexity of the data is critical for preserving the accelerated Bayesian sampling the VAE is designed to provide.


\section{Results and Applications}
\label{sec:results}

As demonstrated in Section \ref{sec:simulation_study}, the variational autoencoder (VAE) possesses the mathematical capacity to approximate complex \textit{Celerite} trends independently of the specific underlying structure or data size. Building upon this validation, the complete additive VAE+HMM framework detailed in the methodology was applied to real empirical stellar brightness data (as introduced in Section \ref{sec:data}).

\subsection{Comparison: \textit{Celerite}+HMM vs. VAE+HMM}

To validate the practical utility of the surrogate model, the inference pipeline described in Algorithm \ref{alg:vae_hmm} was implemented on real observations from TESS. Here, we directly compare the results obtained from the exact \textit{Celerite}+HMM framework against our proposed VAE+HMM architecture for two specific target stars: TIC 089257479 and TIC 234526939. The results for the third target, TIC 031381302, are provided in the supplementary material, as the exact \textit{Celerite}+HMM baseline failed to converge. 

The specific configurations (hidden and latent dimensions) chosen to process these real-world datasets, detailed in Table \ref{tab:real_data_architectures}, were directly informed by the empirical intuition developed during our simulation study, coupled with an \textit{a priori} exploratory data analysis of each respective time series to assess structural complexity and the likely prevalence of flare events. To ensure a rigorously fair baseline for comparison, both the exact and surrogate Bayesian approaches were executed under identical sampling conditions: 2 parallel MCMC chains, each run for 2000 total iterations (comprising 1000 burn-in samples and 1000 posterior samples for each chain).

\begin{table}[b!]
    \centering
    \caption{Configurations of the proposed VAE+HMM framework implemented for the empirical analysis of the target stars. The table details the chosen hidden and latent dimensions alongside the total observation sequence length ($N$) evaluated for each respective TESS light curve.}
    \label{tab:real_data_architectures}
    \begin{tabular}{lccc}
        \toprule
        \textbf{Target Observation} & \textbf{Hidden Dim ($H$)} & \textbf{Latent Dim ($L$)} & \textbf{Data Size ($N$)} \\
        \midrule
        \textbf{TIC 089257479} & 257 & 64 & 7067 \\
        \textbf{TIC 234526939} & 5200 & 125 & 7529 \\
        \bottomrule
    \end{tabular}
\end{table}

\begin{figure}[t!]
    \centering
    
    \begin{minipage}{0.48\textwidth}
        \centering
        \includegraphics[width=\linewidth]{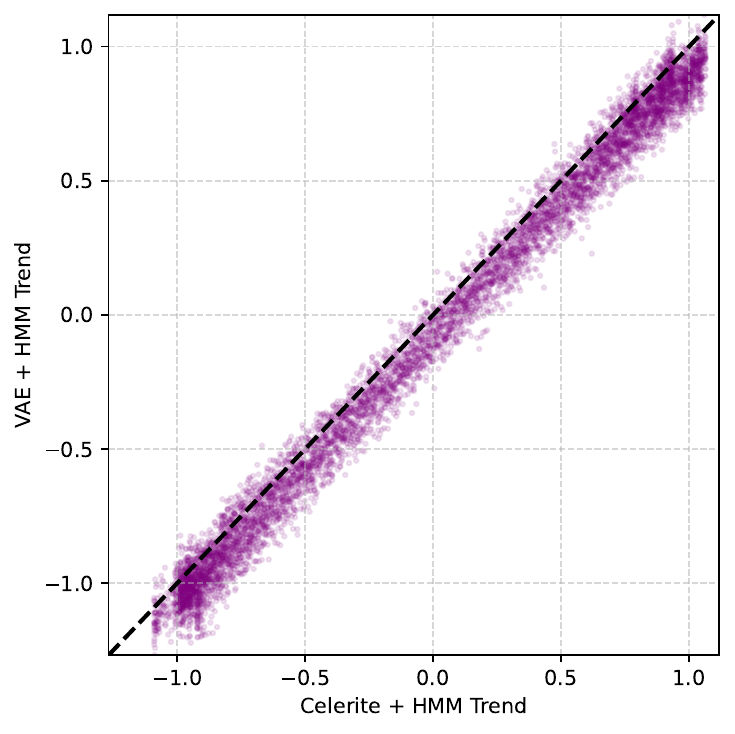} 
        
        \vspace{0.15cm} 
        (a) TIC 089257479
    \end{minipage}\hfill
    \begin{minipage}{0.48\textwidth}
        \centering
        \includegraphics[width=\linewidth]{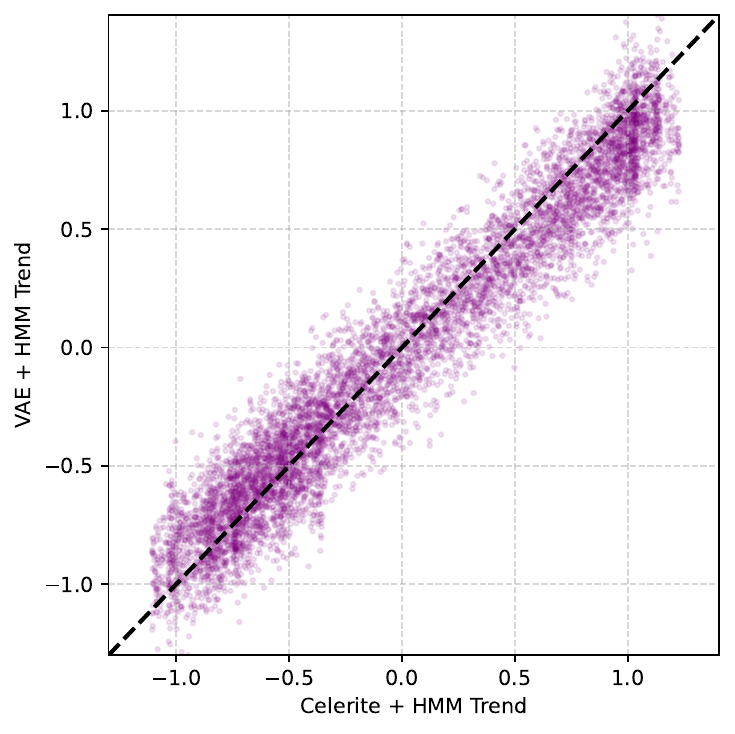} 
        
        \vspace{0.15cm} 
        (b) TIC 234526939
    \end{minipage}

    \vspace{0.25cm} 

    \includegraphics[width=0.6\textwidth]{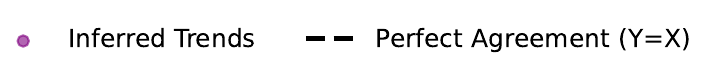} 

    \vspace{-0.5cm}

    \caption{Scatter plots comparing the continuous background trend estimated by the standard \textit{Celerite}+HMM framework (x-axis) against the proposed VAE+HMM surrogate model (y-axis). The dashed black line represents the $Y=X$ identity. The tight adherence of the estimations to this diagonal demonstrates the high accuracy and structural fidelity of the VAE approximation compared to the exact Gaussian Process.}
    \label{fig:yx_scatter_comparison}
\end{figure}

While the simulation study established that the VAE can independently approximate a \textit{Celerite} trend, it is critical to assess its stability when fit jointly with the hidden Markov model component on noisy, real-world data. Figure \ref{fig:yx_scatter_comparison} illustrates this joint performance by plotting the continuous background trend estimated by the exact \textit{Celerite}+HMM model against the trend inferred by the VAE+HMM surrogate for both target stars.

The tight, consistent clustering directly along the dashed $Y=X$ identity line provides strong empirical evidence that the VAE+HMM framework successfully captures the underlying Gaussian Process. In the left panel (TIC 089257479), the variance of the scatter is extremely low. A close inspection of the extreme positive values in this panel reveals a marginal underestimation, where the core density of the surrogate predictions dips slightly below the identity line. This indicates that the VAE inherently applies a minor smoothing effect to the absolute highest peaks of the trend. In the context of scalable Bayesian inference, this minor peak-smoothing is a highly acceptable theoretical trade-off for computational acceleration. 

In contrast, the right panel (TIC 234526939) presents a demonstrably more challenging dataset characterized by a highly complex background structure and a denser population of localized flare events. While this structural complexity naturally introduces a slightly wider predictive variance---visible as a broader dispersion of the purple scatter---the core density of the VAE surrogate remains robustly centered along the identity line. Furthermore, the use of alpha-blending (transparency) in both panels visually confirms that the vast majority of the posterior density remains aligned with the baseline.

Finally, while convergence was successfully achieved in all cases---yielding potential scale reduction factors ($\hat{R}$) strictly less than 1.05 and an effective sample size (ESS) greater than 2000---the VAE+HMM framework demonstrated a large improvement in computational cost.

\begin{table}[b!]
    \centering
    \caption{Comparison of computational inference time (in seconds) between the exact \textit{Celerite}+HMM framework and the proposed VAE+HMM surrogate model when applied to the empirical TESS light curves. The speedup factor highlights the substantial acceleration achieved by the surrogate while maintaining the structural fidelity shown in Figure \ref{fig:yx_scatter_comparison}.}
    \label{tab:real_data_time_comparison}
    \begin{tabular}{lccc}
        \toprule
        \textbf{Target Observation} & \textbf{\textit{Celerite} + HMM} & \textbf{VAE + HMM} & \textbf{Speedup Factor} \\
        \midrule
        \textbf{TIC 089257479} & 39568 & 152 & \textbf{$\sim$ 260x} \\
        \textbf{TIC 234526939} & 55607 & 18584 & \textbf{$\sim$ 3x} \\
        \bottomrule
    \end{tabular}
\end{table}

As detailed in Table \ref{tab:real_data_time_comparison}, replacing the iterative $\mathcal{O}(N)$ matrix inversions of the exact \textit{Celerite} kernel with the pre-trained neural forward pass of the VAE drastically reduced inference times. For TIC 089257479, the surrogate achieved a massive \textbf{$\sim$260x} speedup, compressing what would be hours of computation into minutes. Furthermore, as demonstrated by the results for TIC 234526939, even exceptionally difficult cases featuring complex trend structures and numerous flares remain highly manageable under the proposed framework. While the absolute computation time naturally increases for such demanding time series, the VAE+HMM approach still achieves a \textbf{$\sim$3x} speedup, reaching full convergence significantly faster than the exact GP alternative.

\subsection{Stellar Time Series Analysis}

For our primary target star, TIC 089257479, we present the comprehensive results of the VAE+HMM framework applied to the entire dataset. The model successfully approximates the trend (estimated by the VAE) and distributions of observations in our 3 states (Quiet, Firing, Decay) dictated by the HMM. Because the sheer volume of data in the full light curves visually compresses individual events, we subsequently isolate a specific time window for each star. These zoomed-in panels clearly illustrate the precise detection of localized flare morphologies and are accompanied by their respective Viterbi decoding posterior distributions, demonstrating that we are able to both estimate the trend as well as detect stellar flares with the HMM. To examine the framework's performance on a significantly more challenging dataset, the corresponding analysis for our second target, TIC 234526939, is provided as supplementary material.

\begin{figure}[t!]
    \centering
    \includegraphics[width=0.9\textwidth]{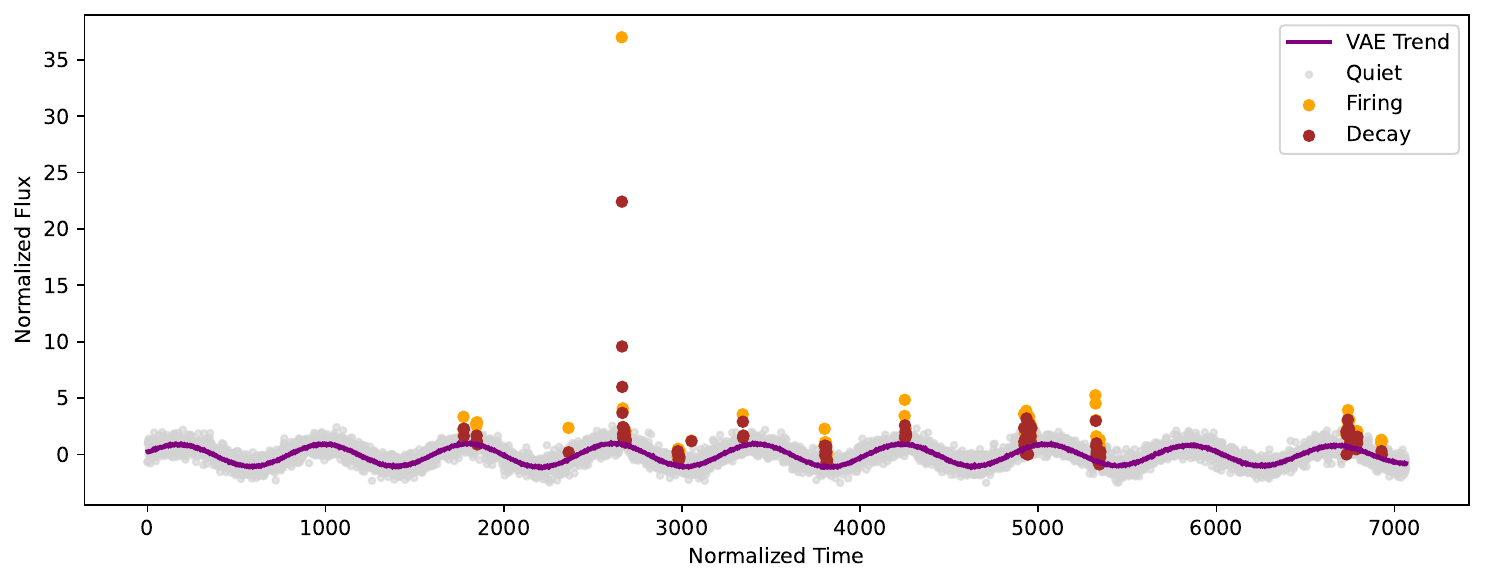} 
    
    \caption{TIC 089257479 mean-centered light curve along with the fit of the proposed VAE+HMM framework, which simultaneously models the continuous background trend using the VAE surrogate, models the distribution of the observations across 3 states (Quiet, Firing, Decay) and assigns each point to its most likely state in the time series.}
    \label{fig:full_model_stareasy}
\end{figure}

Figure \ref{fig:full_model_stareasy} displays the full mean-centered light curve for TIC 089257479. The VAE successfully tracks the star's smooth, highly regular periodic background trend. Against this baseline, the HMM isolates several distinct, high-amplitude flare events. Because the underlying physics of this star are relatively stable, the vast majority of the time series is classified as the Quiet state, punctuated by sparse but energetic outbursts reaching normalized fluxes in excess of 35.

\begin{figure}[t!]
    \centering
    
    \begin{minipage}{\textwidth}
        \centering
        \includegraphics[width=0.85\textwidth]{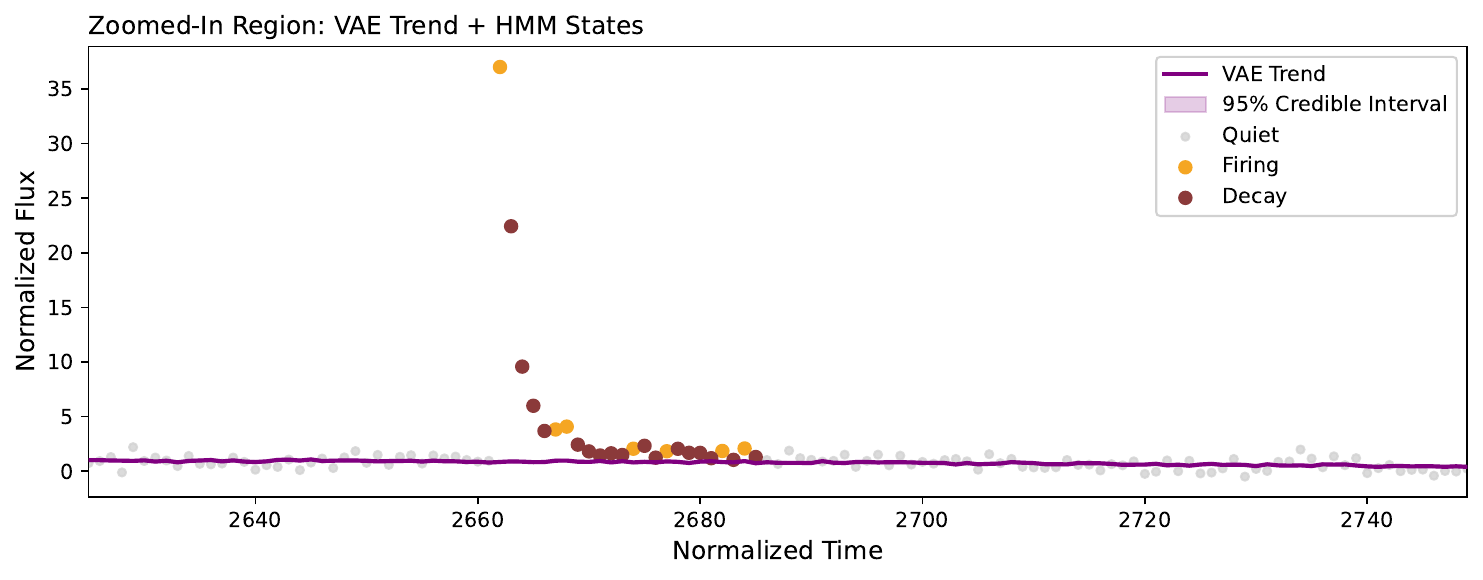} 
    \end{minipage}
    
    \vspace{0.3cm} 
    
    \begin{minipage}{\textwidth}
        \centering
        \includegraphics[width=0.85\textwidth]{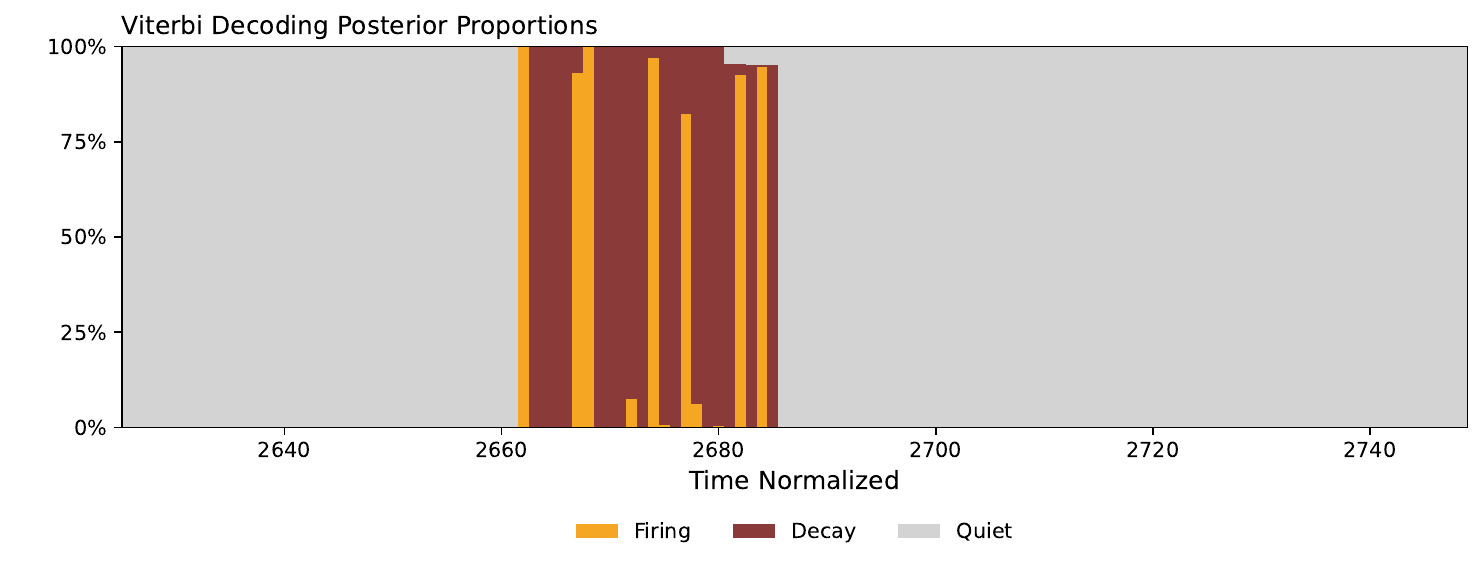} 
    \end{minipage}
    
    \caption{Detected flare example from implementing the VAE+HMM framework on the real-time series of TIC 089257479, focusing on the specific normalized time interval from $t = 2625$ to $t = 2750$. \textbf{Top panel:} Shows the fit of the VAE+HMM model to this localized flare event, including the estimated continuous background trend and the assigned state for each point. \textbf{Bottom panel:} Posterior distribution of Viterbi sequences per observation, representing the proportion of the assigned states over this exact same time interval.}
    \label{fig:zoom_flare_stareasy}
\end{figure}

To examine the model's localized behavior, Figure \ref{fig:zoom_flare_stareasy} zooms in on the most prominent flare event occurring between normalized times $t = 2625$ and $t = 2750$. The top panel illustrates a classic flare morphology: a rapid, singular burst accurately tagged as the Firing state, followed by a gradual return to the baseline appropriately classified as the Decay state. The bottom panel confirms the robustness of the inference; the Viterbi decoding posterior proportions demonstrate near-absolute certainty ($100\%$) in the state assignments. This suggests that the VAE baseline provides a stable foundation for the HMM to identify discrete states.

\section{Discussion}
\label{sec:discussion}

In this paper, we introduced a scalable generative surrogate framework designed to bypass the computational bottlenecks of exact GP inference within additive Bayesian models for astronomical time series. By training a VAE to compress the structured covariance of the \textit{Celerite} kernel into an isotropic latent space, we eliminated the need for repetitive, dense matrix inversions during iterative MCMC sampling. We demonstrated the efficacy of this VAE+HMM architecture on empirical TESS light curves, successfully decomposing the continuous, quasi-periodic stellar rotation from stochastic flaring events. Crucially, our methodology achieved massive computational speedups while preserving the structural fidelity and probabilistic rigor of the exact mathematical formulation, thereby making the fitting of high-cadence datasets highly efficient.

With the computational barriers of the continuous baseline alleviated, this framework is primed for large-scale application across diverse stellar catalogs. Because our joint model efficiently isolates stochastic flares and captures the uncertainty of their durations directly with the HMM, applying this pipeline to massive databases of M-dwarf stars will allow for highly precise estimations of population-level flare energy and flare frequency distributions (FFD). To further accelerate this pipeline at the catalog level, future work will focus on optimizing the downstream discrete inference. While the VAE surrogate successfully resolves the continuous GP bottleneck, implementing the HMM components---specifically the Forward marginalization and Viterbi state-decoding algorithms---within a high-performance, vectorized computing framework like JAX would fully utilize GPU hardware, rendering the end-to-end additive inference almost instantaneous.

Furthermore, expanding the generative surrogate itself offers a highly promising avenue for unified time-series modeling. Currently, the VAE is trained on simulations tailored to individual observational targets. A natural evolution is to construct a single, amortized surrogate model capable of generalizing across a vast array of stars. By conditioning the learning process of the neural network directly on the generative physical parameters (e.g., the exact \textit{Celerite} frequencies and damping factors), the network could dynamically adapt to diverse stellar morphologies without requiring retraining. This conditioned architecture, drawing on advancements such as the PriorCVAE framework \citep{semenova2023priorcvae}, could subsequently be extended to accommodate the multivariate, multi-band observations anticipated from future satellite missions. Ultimately, coupling a conditioned generative surrogate with vectorized discrete inference creates a uniquely powerful paradigm for hierarchical, population-level astrophysical characterization.


\begin{acks}[Acknowledgments]
 RH, VLB and GE received funding from the Data Sciences Institute at the University of Toronto and the Canadian Statistical Sciences Institute, Ontario. VLB and GE were supported by the Natural Sciences and Engineering Research Council of Canada. 
\end{acks}
%


\begin{supplement}
    \stitle{Data and code}
    \sdescription{Data and code developed for reproducibility of the results in the paper are available as Supplementary Material in the form of a compressed folder. The folder includes the empirical TESS light curve files, the complete VAE+HMM codebase featuring an example to run the full model for the results presented in Section \ref{sec:results}, and a detailed README file for its use. This can also be downloaded from a GitHub repository: \url{https://github.com/rodri98bh/VAE_HMM_Stellar_Flares}.}
\end{supplement}


\bibliographystyle{imsart-nameyear} 
\bibliography{bibliography}       

@article{diggle1998model,
  title={Model-based geostatistics},
  author={Diggle, Peter J and Tawn, Jonathan A and Moyeed, Rana A},
  journal={Journal of the Royal Statistical Society: Series C (Applied Statistics)},
  volume={47},
  number={3},
  pages={299--350},
  year={1998},
  publisher={Wiley Online Library}
}

@article{wu2017gaussian,
  title={Gaussian process based nonlinear latent structure discovery in multivariate spike train data},
  author={Wu, Anqi and Koyejo, Oluwasanmi and Pillow, Jonathan},
  journal={Advances in Neural Information Processing Systems},
  volume={30},
  year={2017}
}

@article{esquivel2024detecting,
  title={Detecting stellar flares in photometric data using hidden Markov models},
  author={Esquivel, J Arturo and Shen, Yunyi and Leos-Barajas, Vianey and Eadie, Gwendolyn and Speagle, Joshua and Craiu, Radu V and Medina, Amber and Davenport, James},
  journal={arXiv preprint arXiv:2404.13145},
  year={2024}
}

@article{semenova2022priorvae,
  title={PriorVAE: encoding spatial priors with variational autoencoders for small-area estimation},
  author={Semenova, Elizaveta and Xu, Yidan and Howes, Adam and Rashid, Theo and Bhatt, Samir and Mishra, Swapnil and Flaxman, Seth},
  journal={Journal of The Royal Society Interface},
  volume={19},
  number={191},
  pages={20220094},
  year={2022},
  publisher={The Royal Society}
}

@article{foreman2017fast,
  title={Fast and scalable Gaussian process modeling with applications to astronomical time series},
  author={Foreman-Mackey, Daniel and Agol, Eric and Ambikasaran, Sivaram and Angus, Ruth},
  journal={The Astronomical Journal},
  volume={154},
  number={6},
  pages={220},
  year={2017},
  publisher={IOP Publishing}
}

@book{goodfellow2016deep,
  title={Deep Learning},
  author={Goodfellow, Ian and Bengio, Yoshua and Courville, Aaron},
  publisher={MIT Press},
  year={2016}
}

@inproceedings{snelson2006,
  title={Sparse Gaussian processes using pseudo-inputs},
  author={Snelson, Edward and Ghahramani, Zoubin},
  booktitle={Advances in Neural Information Processing Systems},
  volume={18},
  year={2005}
}

@inproceedings{titsias2009,
  title={Variational learning of inducing variables in sparse Gaussian processes},
  author={Titsias, Michalis},
  booktitle={Artificial Intelligence and Statistics},
  pages={567--574},
  year={2009},
  organization={PMLR}
}

@article{semenova2023priorcvae,
  title={PriorCVAE: scalable MCMC parameter inference with Bayesian deep generative modelling},
  author={Semenova, Elizaveta and Verma, Prakhar and Cairney-Leeming, Max and Solin, Arno and Bhatt, Samir and Flaxman, Seth},
  journal={arXiv preprint arXiv:2304.04307},
  year={2023}
}

@article{belkin2019reconciling,
  title={Reconciling modern machine-learning practice and the classical bias--variance trade-off},
  author={Belkin, Mikhail and Hsu, Daniel and Ma, Siyuan and Mandal, Soumik},
  journal={Proceedings of the National Academy of Sciences},
  volume={116},
  number={32},
  pages={15849--15854},
  year={2019},
  publisher={National Acad Sciences}
}

@book{zucchini2016hidden,
  title     = {Hidden {Markov} Models for Time Series: An Introduction Using {R}},
  author    = {Zucchini, Walter and MacDonald, Iain L. and Langrock, Roland},
  year      = {2016},
  edition   = {2nd},
  publisher = {CRC Press},
  address   = {Boca Raton, FL},
  series    = {Monographs on Statistics and Applied Probability},
  isbn      = {9781482253832}
}

@article{hoffman2014nuts,
  title   = {The {No-U-Turn} Sampler: Adaptively Setting Path Lengths
             in {Hamiltonian} {Monte} {Carlo}},
  author  = {Hoffman, Matthew D. and Gelman, Andrew},
  journal = {Journal of Machine Learning Research},
  year    = {2014},
  volume  = {15},
  number  = {47},
  pages   = {1593--1623},
  url     = {http://jmlr.org/papers/v15/hoffman14a.html}
}

@article{ambikasaran2015fast,
  title={Fast direct methods for Gaussian processes},
  author={Ambikasaran, Sivaram and Foreman-Mackey, Daniel and Greengard, Leslie and Hogg, David W and O'Neil, Michael},
  journal={IEEE Transactions on Pattern Analysis and Machine Intelligence},
  volume={38},
  number={2},
  pages={252--265},
  year={2015},
  publisher={IEEE}
}

@article{ricker2015transiting,
  title={Transiting Exoplanet Survey Satellite (TESS)},
  author={Ricker, George R and Winn, Joshua N and Vanderspek, Roland and Latham, David W and Bakos, G{\'a}sp{\'a}r {\'A} and Bean, Jacob L and Berta-Thompson, Zachory K and Brown, Timothy M and Buchhave, Lars and Butler, Nathaniel R and others},
  journal={Journal of Astronomical Telescopes, Instruments, and Systems},
  volume={1},
  number={1},
  pages={014003},
  year={2015},
  publisher={SPIE}
}

@inproceedings{jenkins2016tess,
  title={The TESS science processing operations center},
  author={Jenkins, Jon M and Twicken, Joseph D and McCauliff, Sean and Campbell, Jennifer and Sanderfer, Dwight and Lung, David and Mansouri-Samani, Masoud and Girouard, Wendy and Tenenbaum, Peter and Klaus, Todd and others},
  booktitle={Software and Cyberinfrastructure for Astronomy IV},
  volume={9913},
  pages={99133E},
  year={2016},
  organization={SPIE}
}

@book{rasmussen2006gaussian,
  title={Gaussian Processes for Machine Learning},
  author={Rasmussen, Carl Edward and Williams, Christopher K. I.},
  year={2006},
  publisher={MIT Press},
  address={Cambridge, MA}
}

@article{phan2019composable,
  title={Composable effects for flexible and accelerated probabilistic programming in NumPyro},
  author={Phan, Du and Pradhan, Neeraj and Jankowiak, Martin},
  journal={arXiv preprint arXiv:1912.11554},
  year={2019}
}

@article{quinonero2005unifying,
  title={A unifying view of sparse approximate Gaussian process regression},
  author={Qui{\~n}onero-Candela, Joaquin and Rasmussen, Carl Edward},
  journal={Journal of Machine Learning Research},
  volume={6},
  number={Dec},
  pages={1939--1959},
  year={2005}
}

@inproceedings{hensman2013gaussian,
  title={Gaussian processes for big data},
  author={Hensman, James and Fusi, Nicol{\`o} and Lawrence, Neil D},
  booktitle={Proceedings of the 29th Conference on Uncertainty in Artificial Intelligence (UAI)},
  pages={282--290},
  year={2013},
  organization={AUAI Press}
}

@article{rue2009approximate,
  title={Approximate Bayesian inference for latent Gaussian models by using integrated nested Laplace approximations},
  author={Rue, H{\aa}vard and Martino, Sara and Chopin, Nicolas},
  journal={Journal of the Royal Statistical Society: Series B (Statistical Methodology)},
  volume={71},
  number={2},
  pages={319--392},
  year={2009},
  publisher={Wiley Online Library}
}

@inproceedings{duvenaud2011additive,
  title={Additive Gaussian processes},
  author={Duvenaud, David K and Nickisch, Hannes and Rasmussen, Carl Edward},
  booktitle={Advances in Neural Information Processing Systems},
  volume={24},
  pages={226--234},
  year={2011}
}

@article{roberts2013gaussian,
  title={Gaussian processes for time-series modelling},
  author={Roberts, Stephen and Osborne, Michael and Ebden, Mark and Reece, Steven and Gibson, Neil and Aigrain, Suzanne},
  journal={Philosophical Transactions of the Royal Society A: Mathematical, Physical and Engineering Sciences},
  volume={371},
  number={1984},
  pages={20110550},
  year={2013},
  publisher={The Royal Society Publishing}
}

@inproceedings{bauer2016understanding,
  title={Understanding probabilistic sparse Gaussian process approximations},
  author={Bauer, Matthias and van der Wilk, Mark and Rasmussen, Carl Edward},
  booktitle={Advances in Neural Information Processing Systems},
  volume={29},
  pages={1533--1541},
  year={2016}
}

@article{filippone2013comparative,
  title={A comparative evaluation of stochastic-based inference methods for Gaussian process models},
  author={Filippone, Maurizio and Zhong, Mingjun and Girolami, Mark},
  journal={Machine Learning},
  volume={93},
  number={1},
  pages={93--114},
  year={2013},
  publisher={Springer}
}

@incollection{betancourt2015hamiltonian,
  title={Hamiltonian Monte Carlo for hierarchical models},
  author={Betancourt, Michael and Girolami, Mark},
  booktitle={Current trends in Bayesian methodology with applications},
  pages={79--101},
  year={2015},
  publisher={CRC Press}
}

@article{angus2018inferring,
  title={Inferring probabilistic stellar rotation periods using Gaussian processes},
  author={Angus, Ruth and Morton, Timothy and Aigrain, Suzanne and Foreman-Mackey, Daniel and Rajpaul, Vinesh},
  journal={Monthly Notices of the Royal Astronomical Society},
  volume={474},
  number={2},
  pages={2094--2108},
  year={2018},
  publisher={Oxford University Press}
}

@article{kingma2013auto,
  title={Auto-encoding variational bayes},
  author={Kingma, Diederik P and Welling, Max},
  journal={arXiv preprint arXiv:1312.6114},
  year={2013}
}


\newpage

\appendix

\setcounter{figure}{0}
\renewcommand{\thefigure}{\Alph{section}.\arabic{figure}}

\setcounter{equation}{0}
\renewcommand{\theequation}{\Alph{section}.\arabic{equation}}

\section{HMM Inference: Forward and Viterbi Algorithms}
Substituting the Gaussian Process (GP) with the variational autoencoder (VAE) decoder only changes how the continuous trend is generated; it does not alter the fundamental probabilistic structure of the hidden Markov model (HMM). This means we can still fit an HMM to the data and estimate the parameters associated with the three states of interest. Once fitted, we can perform state decoding and assign the most likely states that generated our data using standard dynamic programming methods such as the Viterbi algorithm. 

We depend on two algorithms to perform inference and do state decoding in HMMs for this paper, the forward algorithm for likelihood evaluation and the Viterbi algorithm for state decoding. 

\subsection{The Forward Algorithm (Likelihood Evaluation)}
The forward algorithm computes the (marginal) likelihood of an HMM. Because there are many possible sequences of hidden states that could have produced the data, enumerating every single path is computationally intractable. Instead, the forward algorithm uses dynamic programming to efficiently sum over the probabilities of all possible state sequences step-by-step.

Let $\alpha_t(j) = P(R_1, \dots, R_t, S_t = j)$ be the forward probability of observing the residuals up to time $t$ and ending in state $j$. The recursive update relies on this summation:
\begin{equation}
    \alpha_t(j) = \left( \sum_{i \in \{Q,F,D\}} \alpha_{t-1}(i) \Theta_{ij} \right) \cdot P(R_t | S_t = j)
\end{equation}
where $\Theta_{ij}$ is the state transition matrix and $P(R_t | S_t = j)$ is the emission probability of the residual under the distribution of state $j$. The likelihood $\mathcal{L}_{VAE+HMM}$ can be computed by $\sum_{i \in \{Q, F, D\}}\alpha_T(i)$.

\subsection{The Viterbi Algorithm (State Decoding)}
The Viterbi algorithm estimates the most likely sequence of hidden states that produced the observations given the parameter values. Once the MCMC sampler converges and we have our final model parameters, we want to know exactly when the system was in each specific state. Let $V_t(j)$ be the maximum probability of any path leading to state $j$ at time $t$. The recursive update is:
\begin{equation}
    V_t(j) = \max_{i \in \{Q,F,D\}} \left( V_{t-1}(i) \Theta_{ij} \right) \cdot P(R_t | S_t = j)
\end{equation}
During this forward pass, the algorithm actively stores "backpointers" to remember the best previous state choice at each step. Finally, it traces these backpointers in reverse to reconstruct the optimal hidden state sequence (the Viterbi path).

\section{Extended Simulation Diagnostics}

This supplementary document provides the complete set of diagnostic plots for the simulation study presented in Section 4 of the main manuscript. 

For each of the three simulated Celerite scenarios (Smooth \& Simple, Fast \& Noisy, Complex \& Amplitude Modulated), we provide visual diagnostics across the three evaluated data sizes ($N = 2000, 5000, 7000$). Within each condition, we plot the variational autoencoder (VAE) approximation against the true continuous Gaussian Process (GP) trend (left column), and the corresponding residual distributions (right column) for all five evaluated architectural configurations (Tiny through Ultra).

\clearpage

\subsection{Scenario 1: Smooth \& Simple}

\begin{center}
    \begin{tabular}{ccc}
        & \textbf{VAE Approximation Fit} & \textbf{Residuals} \\
        
        \raisebox{1.5cm}{\textbf{Tiny}} & 
        \includegraphics[width=0.4\textwidth]{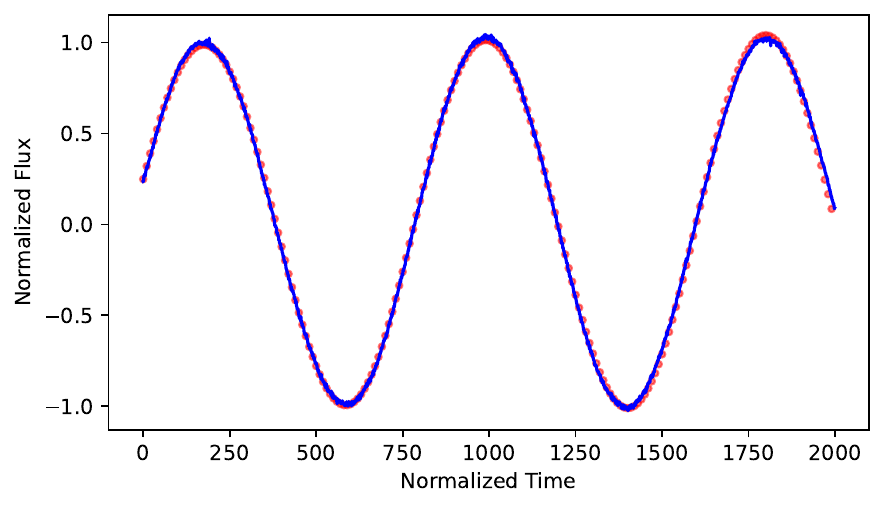} & 
        \includegraphics[width=0.4\textwidth]{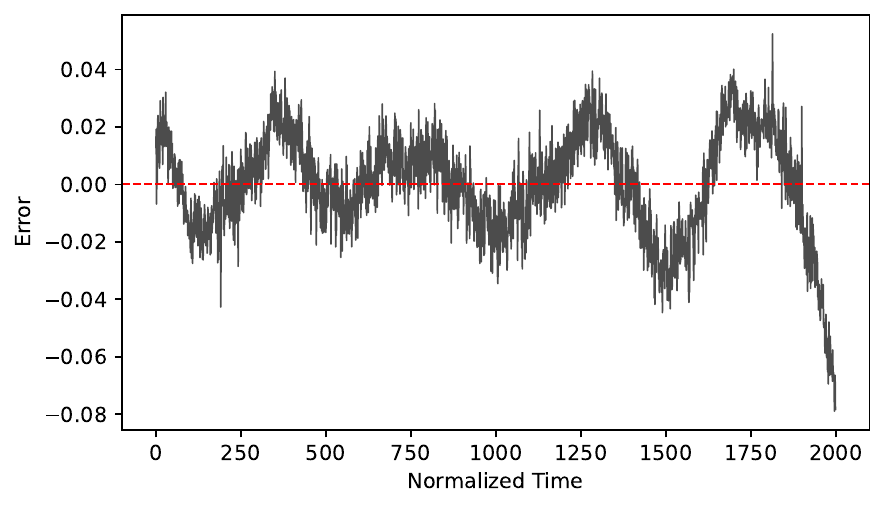} \\
        
        \raisebox{1.5cm}{\textbf{Small}} & 
        \includegraphics[width=0.4\textwidth]{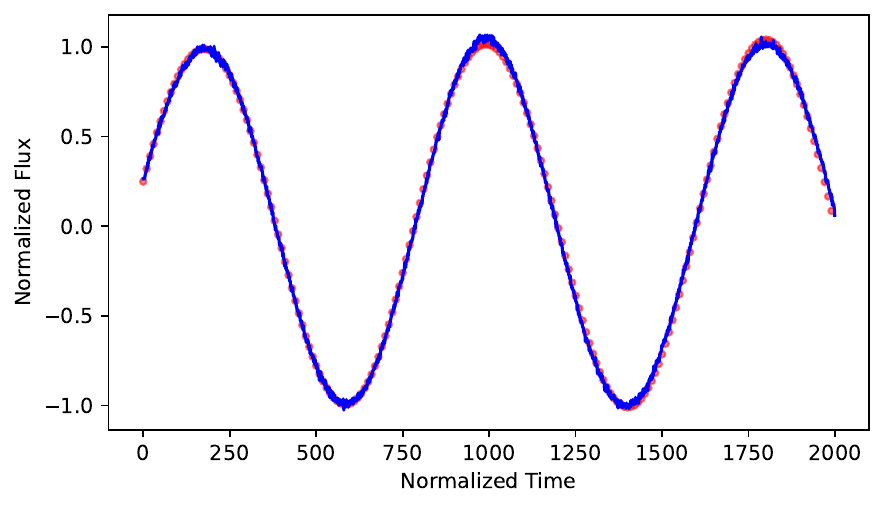} & 
        \includegraphics[width=0.4\textwidth]{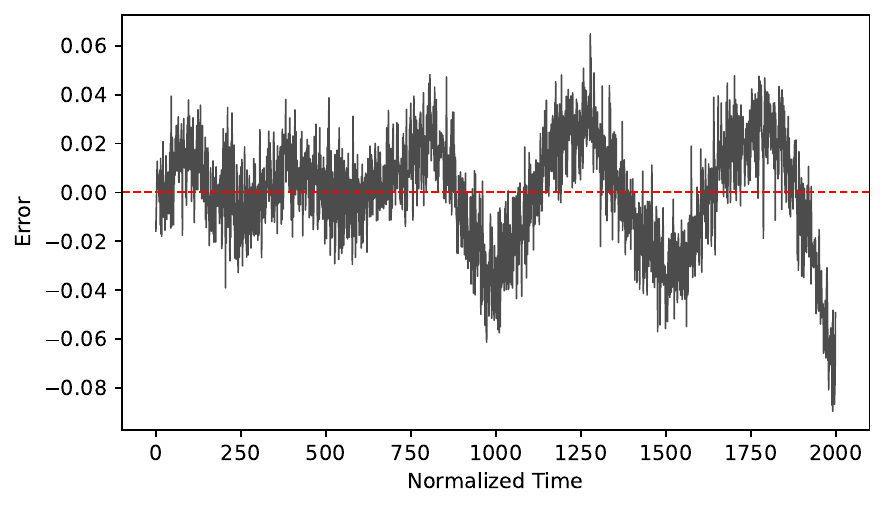} \\
        
        \raisebox{1.5cm}{\textbf{Medium}} & 
        \includegraphics[width=0.4\textwidth]{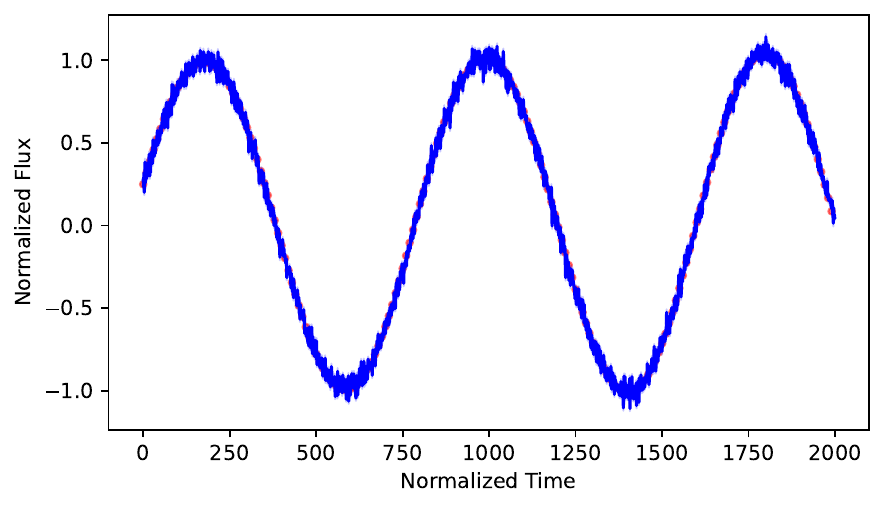} & 
        \includegraphics[width=0.4\textwidth]{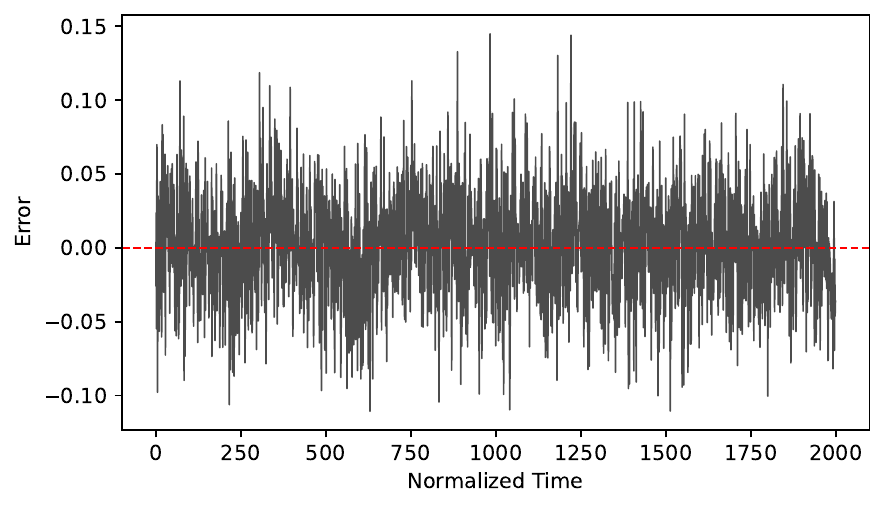} \\
        
        \raisebox{1.5cm}{\textbf{Large}} & 
        \includegraphics[width=0.4\textwidth]{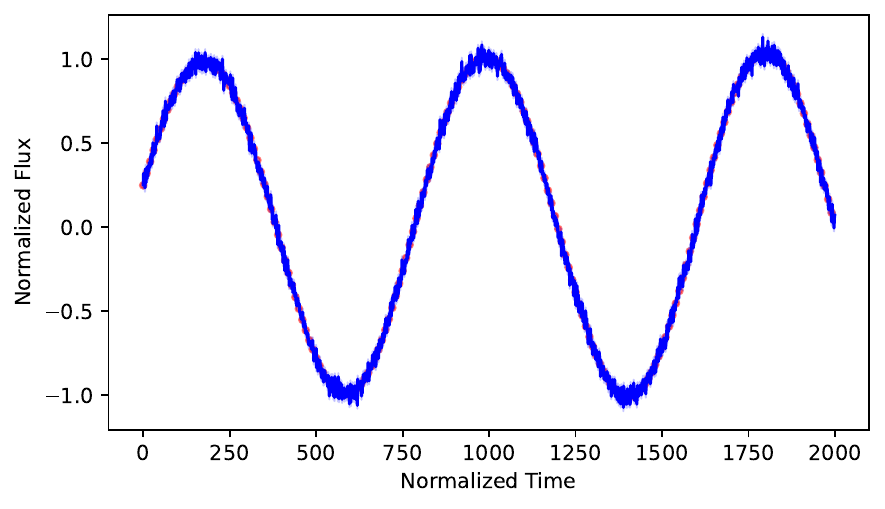} & 
        \includegraphics[width=0.4\textwidth]{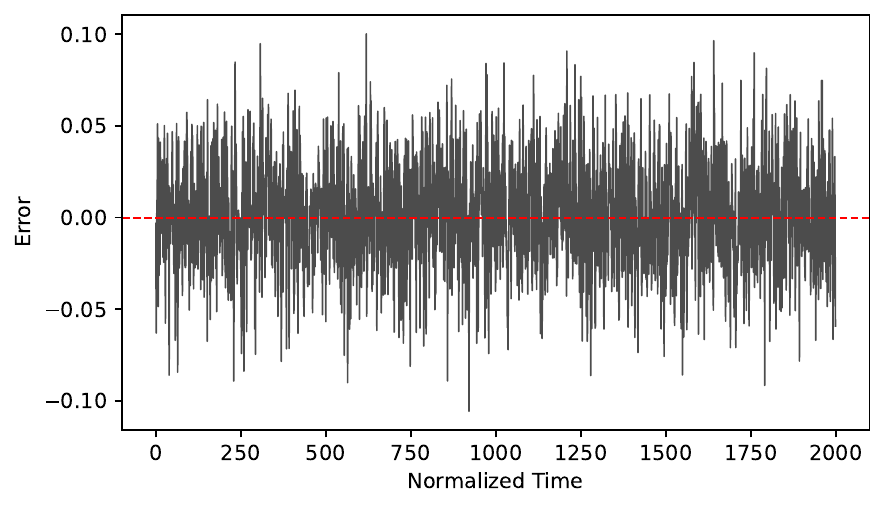} \\
        
        \raisebox{1.5cm}{\textbf{Ultra}} & 
        \includegraphics[width=0.4\textwidth]{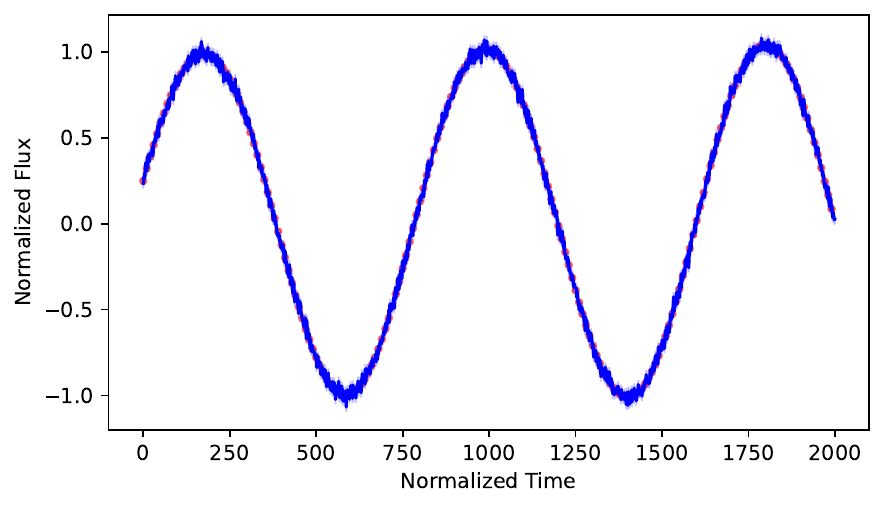} & 
        \includegraphics[width=0.4\textwidth]{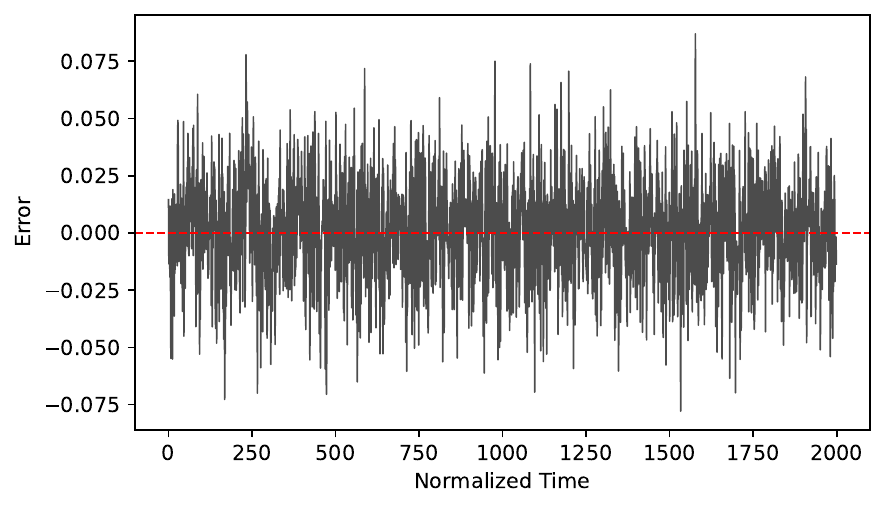} \\
    \end{tabular}

    \vspace{0.5cm} 

    \includegraphics[width=0.6\textwidth]{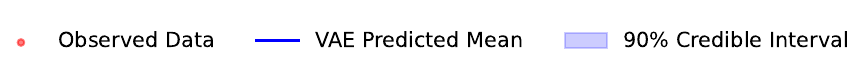} 

    \vspace{0.5cm} 
    
    \captionof{figure}{Diagnostic plots for \textbf{Scenario 1 ($N=2000$)}. The left column displays the true GP trend versus the surrogate VAE approximation. The right column displays the corresponding residual errors.}
    \label{fig:S1_N2000}
\end{center}

\begin{center}
    \begin{tabular}{ccc}
        & \textbf{VAE Approximation Fit} & \textbf{Residuals} \\
        
        \raisebox{1.5cm}{\textbf{Tiny}} & 
        \includegraphics[width=0.4\textwidth]{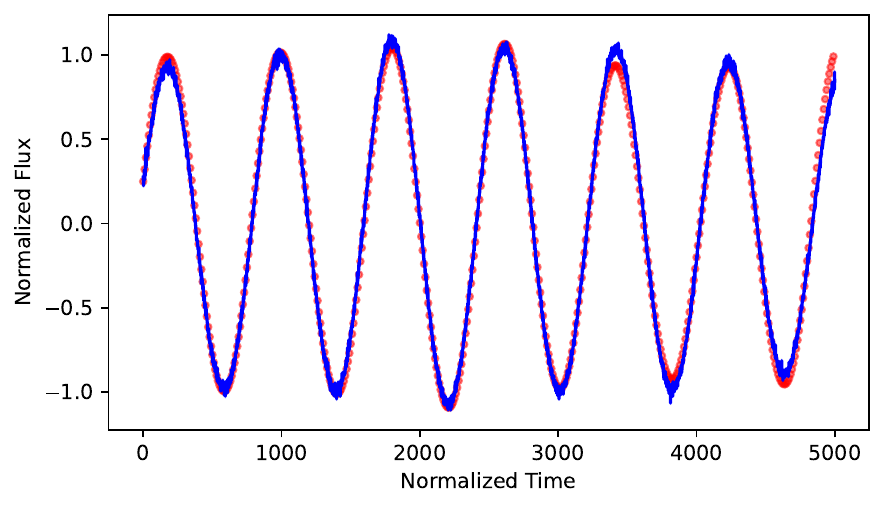} & 
        \includegraphics[width=0.4\textwidth]{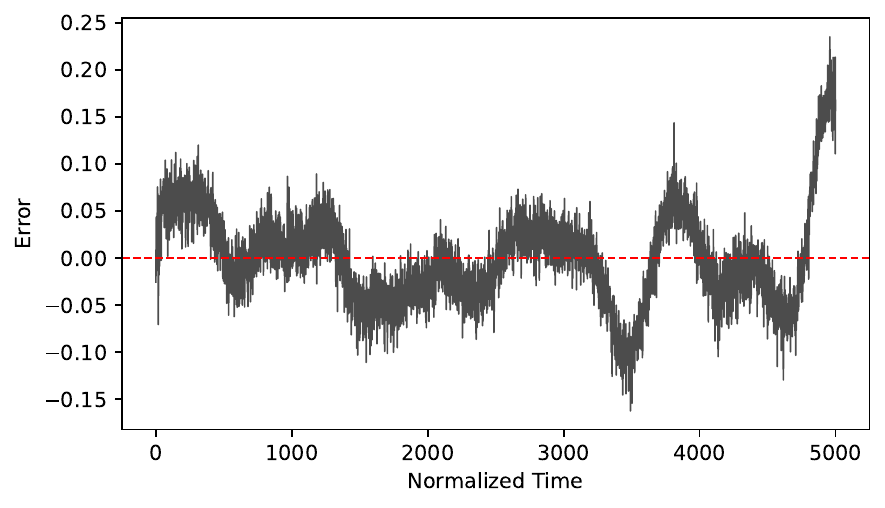} \\
        
        \raisebox{1.5cm}{\textbf{Small}} & 
        \includegraphics[width=0.4\textwidth]{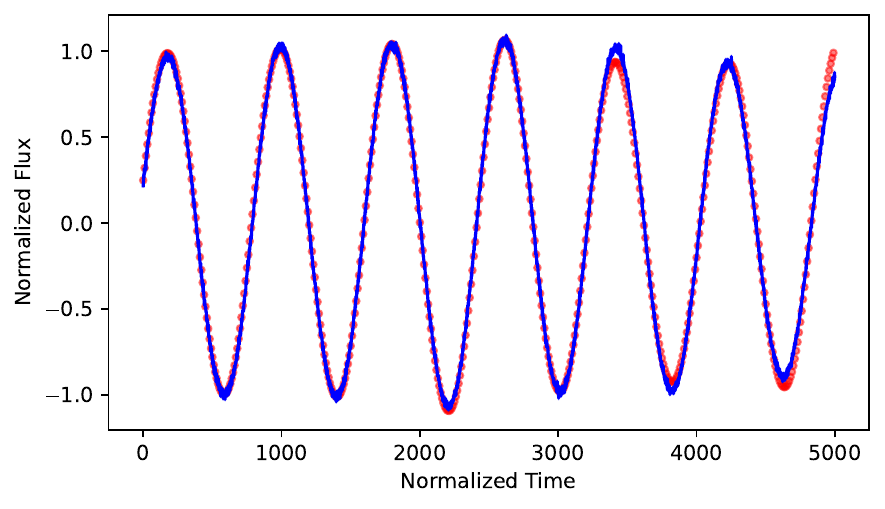} & 
        \includegraphics[width=0.4\textwidth]{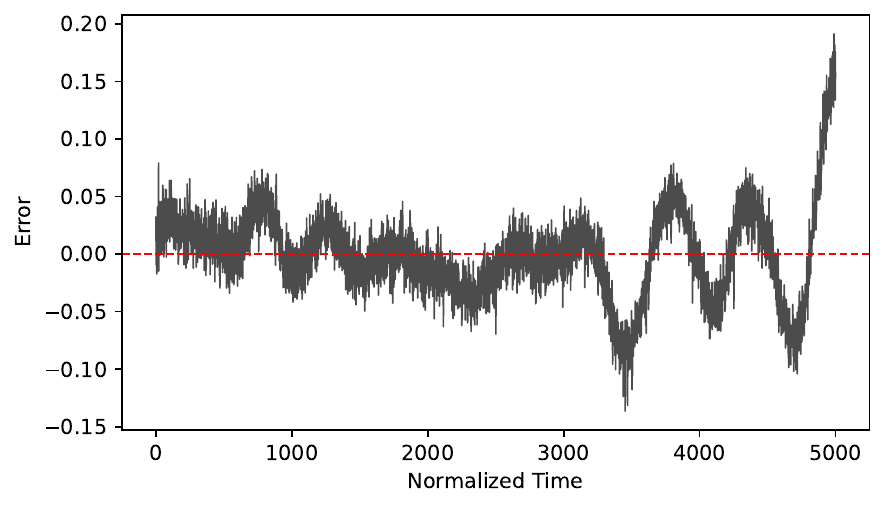} \\
        
        \raisebox{1.5cm}{\textbf{Medium}} & 
        \includegraphics[width=0.4\textwidth]{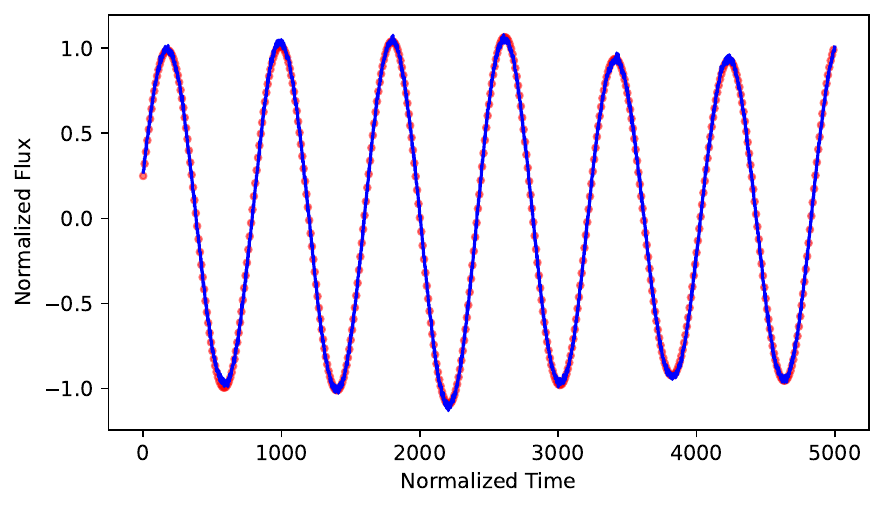} & 
        \includegraphics[width=0.4\textwidth]{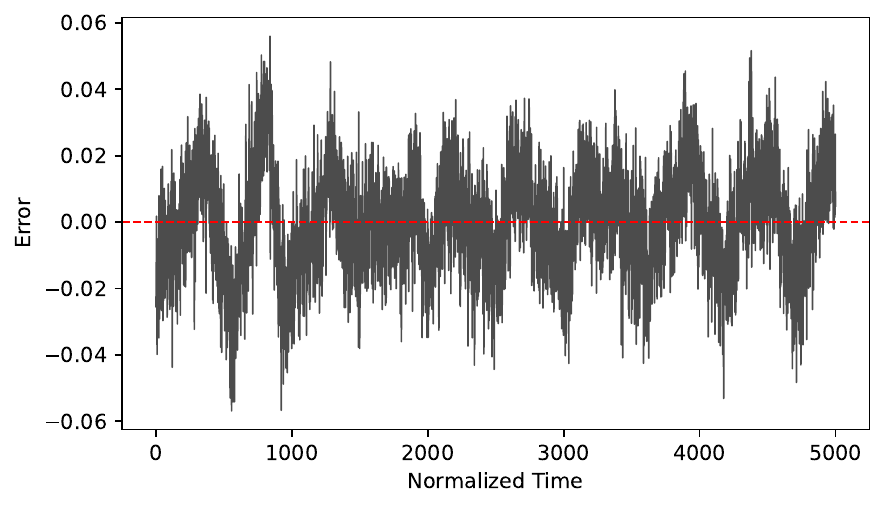} \\
        
        \raisebox{1.5cm}{\textbf{Large}} & 
        \includegraphics[width=0.4\textwidth]{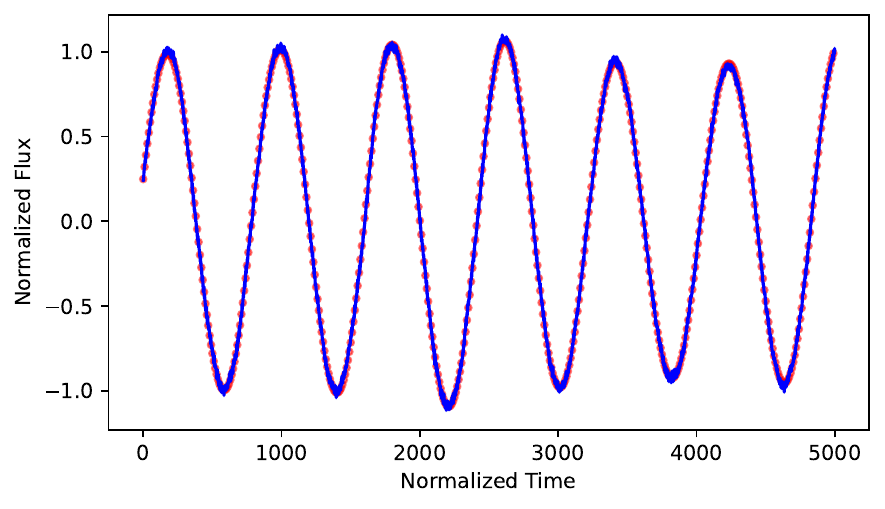} & 
        \includegraphics[width=0.4\textwidth]{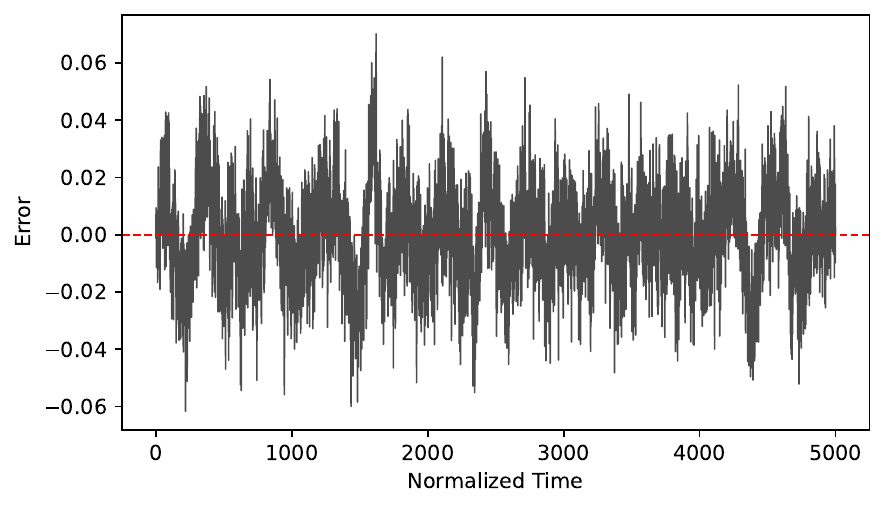} \\
        
        \raisebox{1.5cm}{\textbf{Ultra}} & 
        \includegraphics[width=0.4\textwidth]{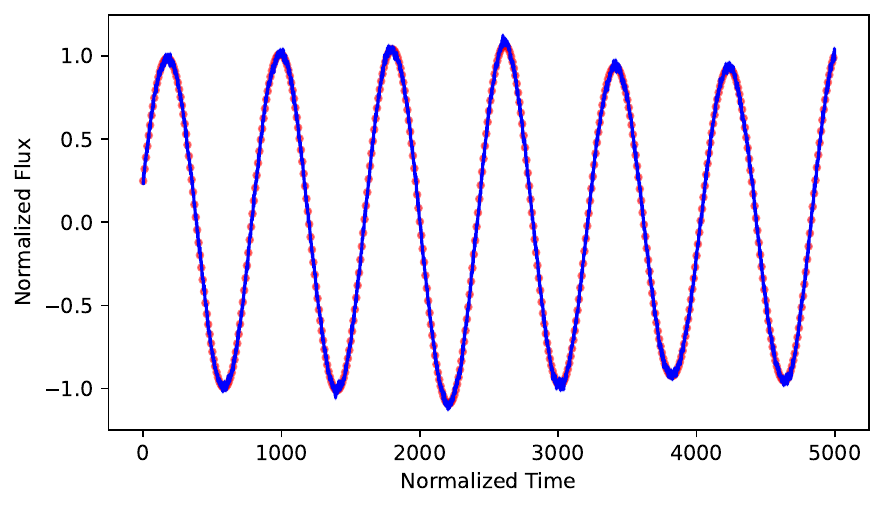} & 
        \includegraphics[width=0.4\textwidth]{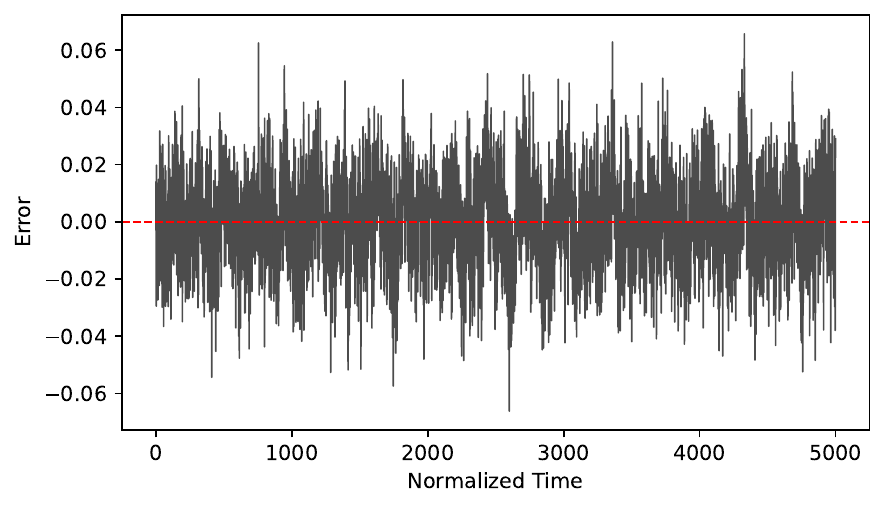} \\
    \end{tabular}

    \vspace{0.5cm} 

    \includegraphics[width=0.6\textwidth]{VAE_fitting_shared_legend.pdf} 

    \vspace{0.5cm} 
    
    \captionof{figure}{Diagnostic plots for \textbf{Scenario 1 ($N=5000$)}. The left column displays the true GP trend versus the surrogate VAE approximation. The right column displays the corresponding residual errors.}
    \label{fig:S1_N5000}
\end{center}

\begin{center}
    \begin{tabular}{ccc}
        & \textbf{VAE Approximation Fit} & \textbf{Residuals} \\
        
        \raisebox{1.5cm}{\textbf{Tiny}} & 
        \includegraphics[width=0.4\textwidth]{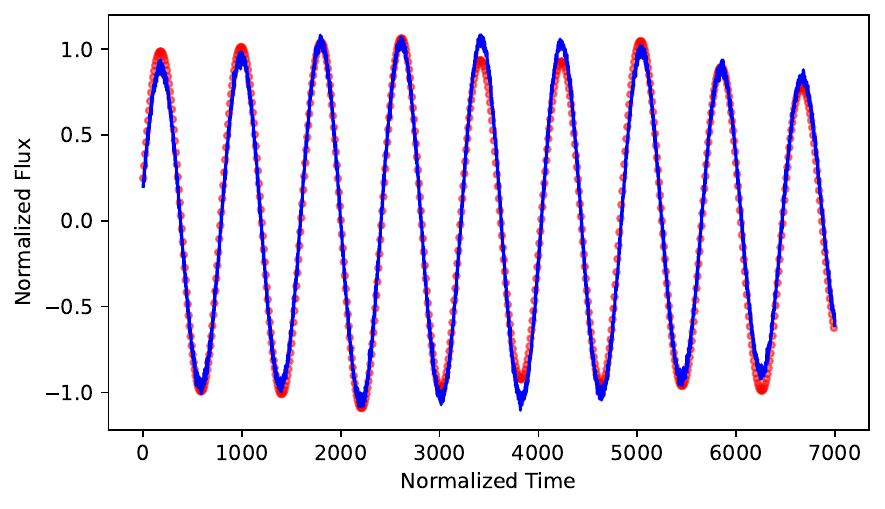} & 
        \includegraphics[width=0.4\textwidth]{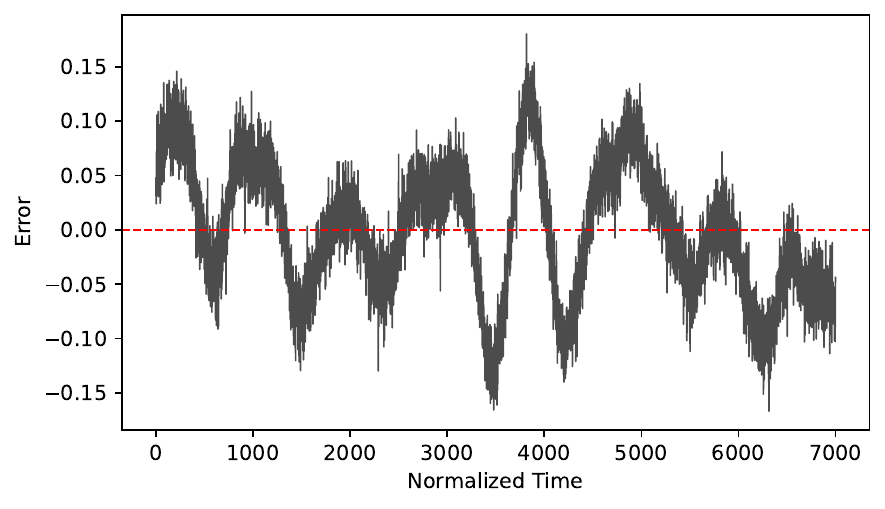} \\
        
        \raisebox{1.5cm}{\textbf{Small}} & 
        \includegraphics[width=0.4\textwidth]{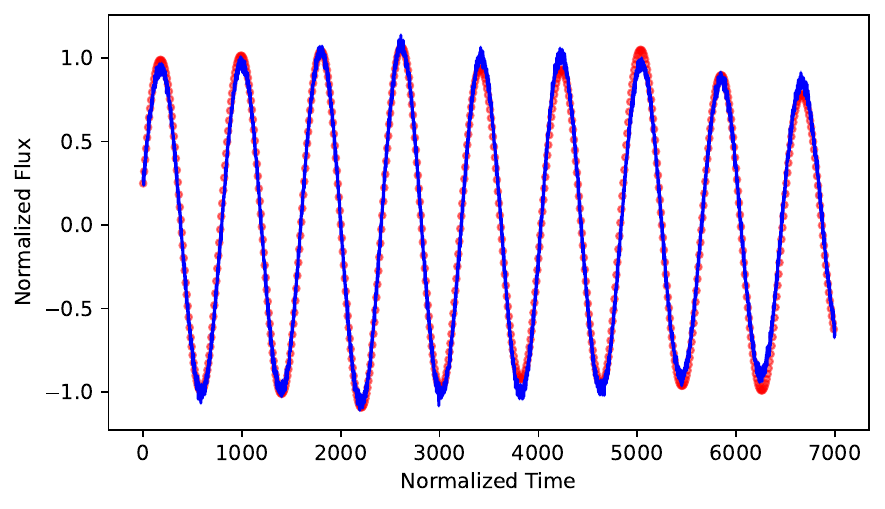} & 
        \includegraphics[width=0.4\textwidth]{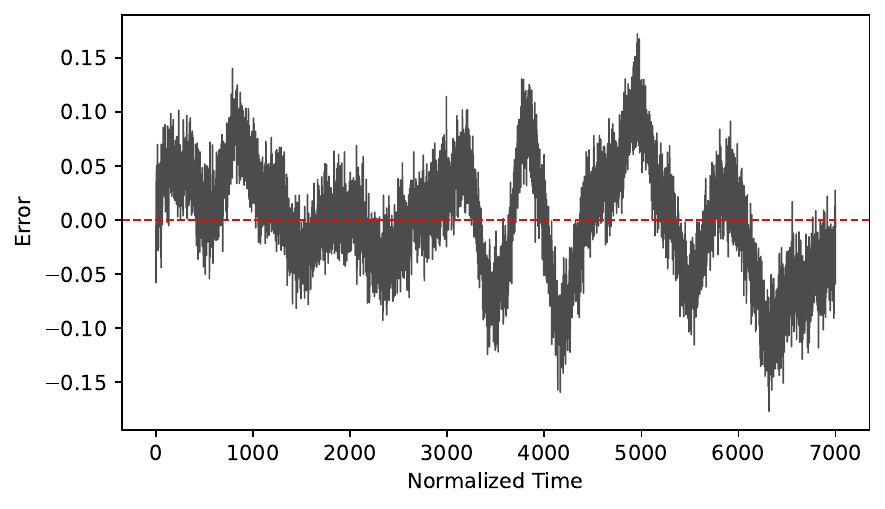} \\
        
        \raisebox{1.5cm}{\textbf{Medium}} & 
        \includegraphics[width=0.4\textwidth]{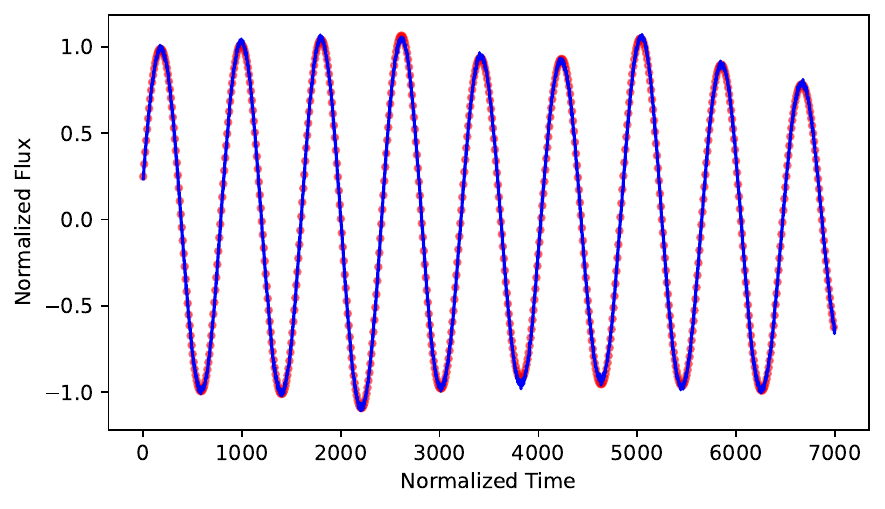} & 
        \includegraphics[width=0.4\textwidth]{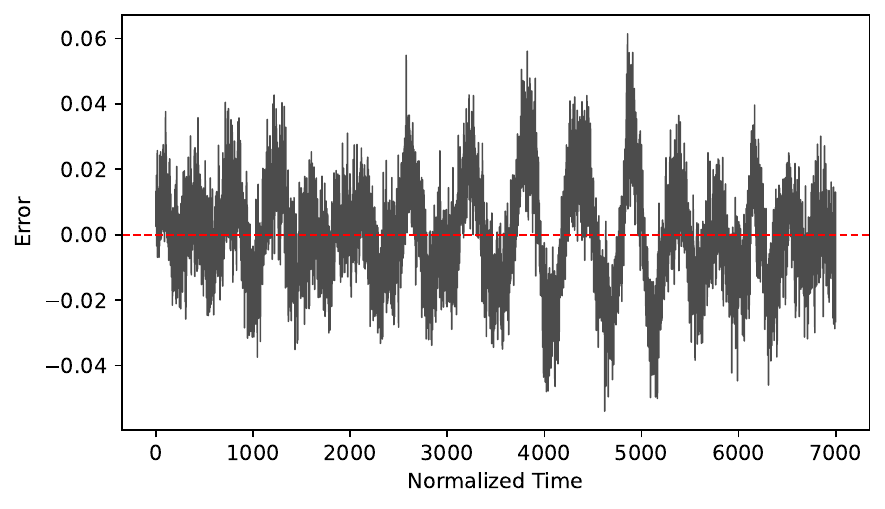} \\
        
        \raisebox{1.5cm}{\textbf{Large}} & 
        \includegraphics[width=0.4\textwidth]{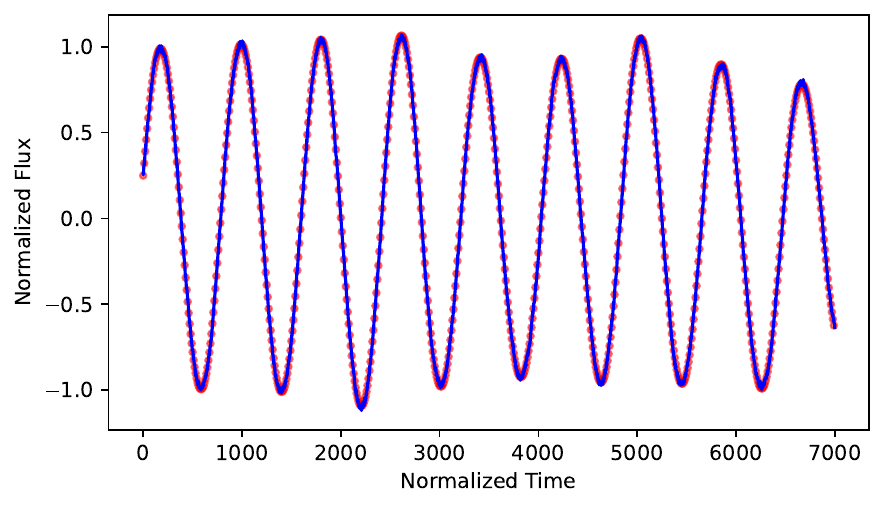} & 
        \includegraphics[width=0.4\textwidth]{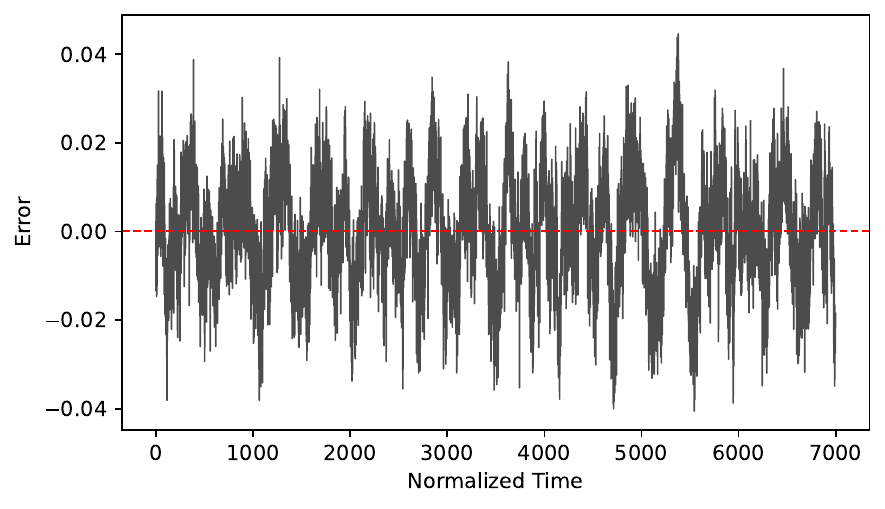} \\
        
        \raisebox{1.5cm}{\textbf{Ultra}} & 
        \includegraphics[width=0.4\textwidth]{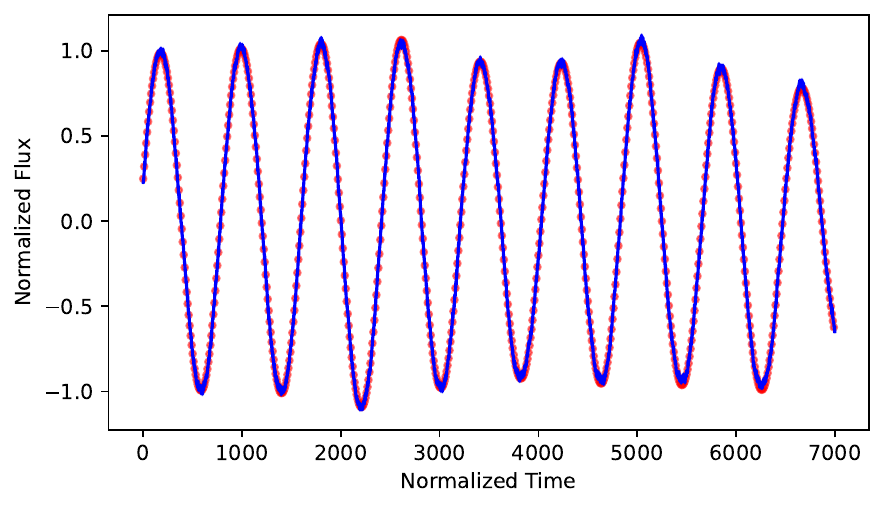} & 
        \includegraphics[width=0.4\textwidth]{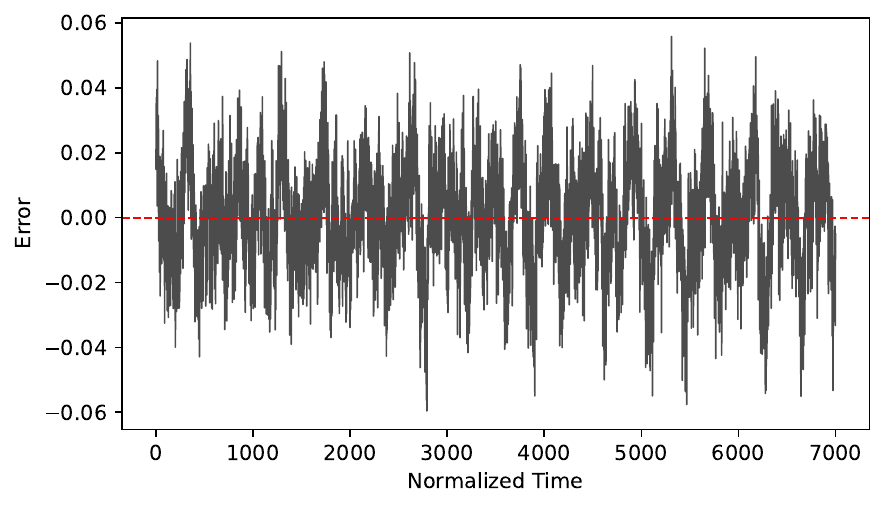} \\
    \end{tabular}

    \vspace{0.5cm} 

    \includegraphics[width=0.6\textwidth]{VAE_fitting_shared_legend.pdf} 

    \vspace{0.5cm} 
    
    \captionof{figure}{Diagnostic plots for \textbf{Scenario 1 ($N=7000$)}. The left column displays the true GP trend versus the surrogate VAE approximation. The right column displays the corresponding residual errors.}
    \label{fig:S1_N7000}
\end{center}

\clearpage

\subsection{Scenario 2: Fast \& Noisy}


\begin{center}
    \begin{tabular}{ccc}
        & \textbf{VAE Approximation Fit} & \textbf{Residuals} \\
        
        \raisebox{1.5cm}{\textbf{Tiny}} & 
        \includegraphics[width=0.4\textwidth]{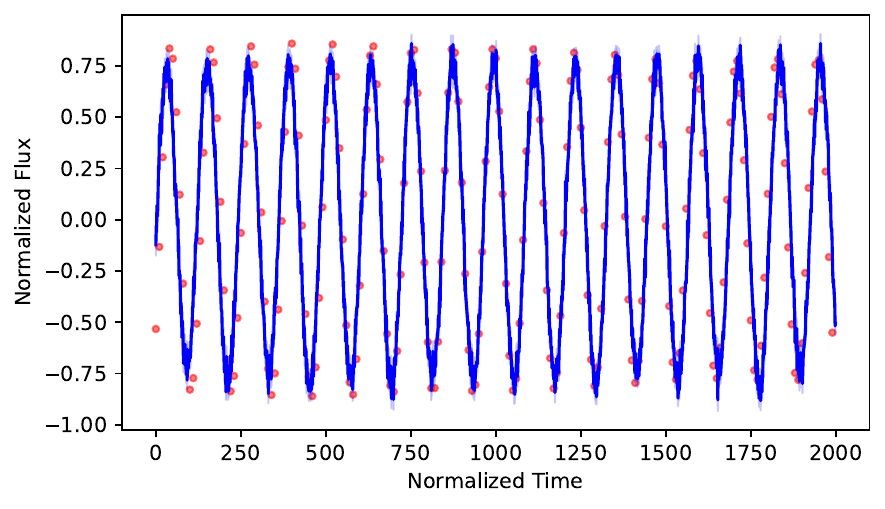} & 
        \includegraphics[width=0.4\textwidth]{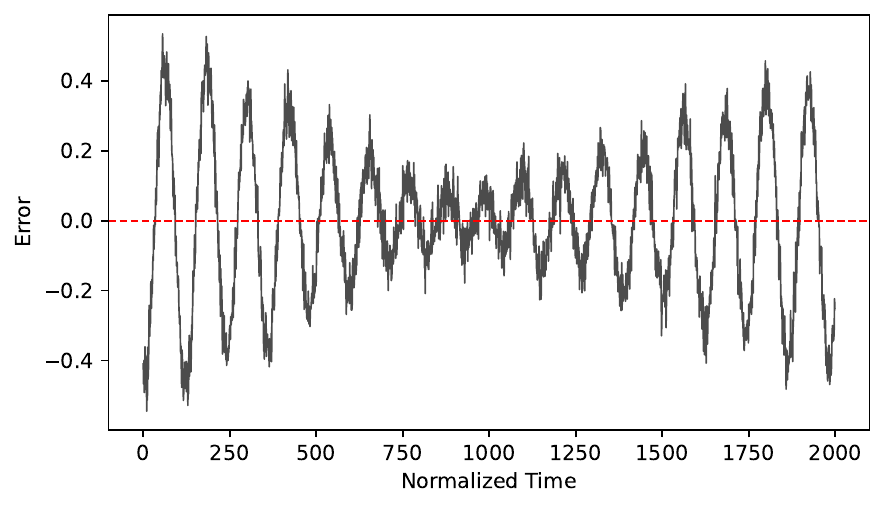} \\
        
        \raisebox{1.5cm}{\textbf{Small}} & 
        \includegraphics[width=0.4\textwidth]{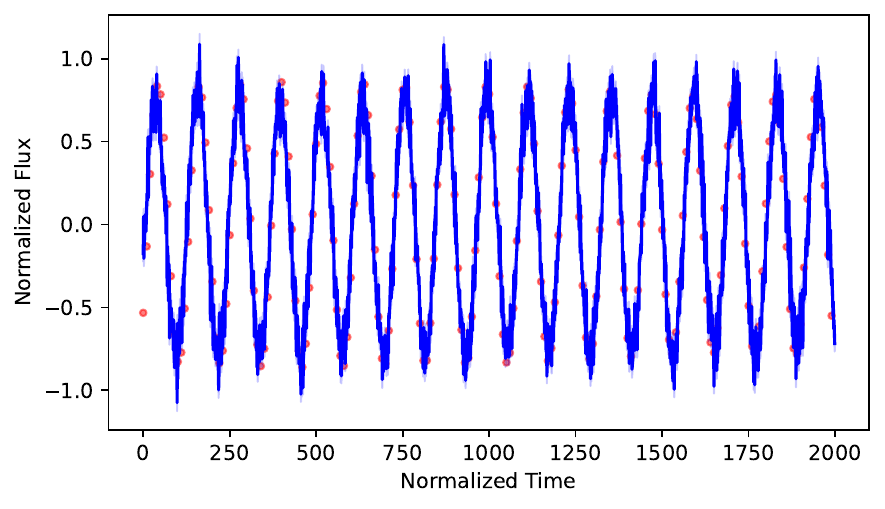} & 
        \includegraphics[width=0.4\textwidth]{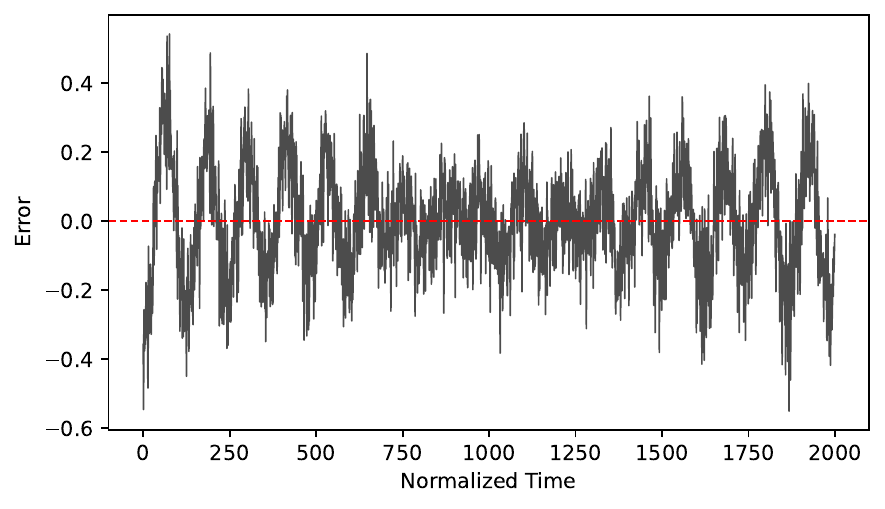} \\
        
        \raisebox{1.5cm}{\textbf{Medium}} & 
        \includegraphics[width=0.4\textwidth]{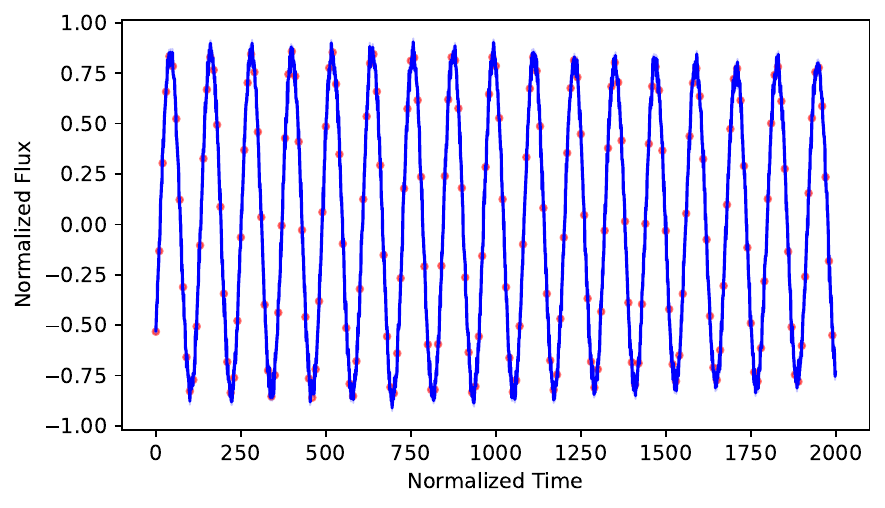} & 
        \includegraphics[width=0.4\textwidth]{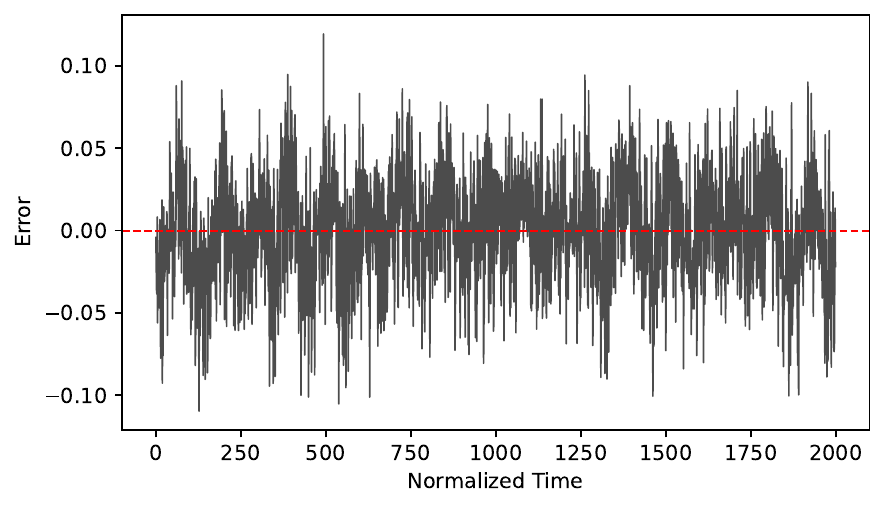} \\
        
        \raisebox{1.5cm}{\textbf{Large}} & 
        \includegraphics[width=0.4\textwidth]{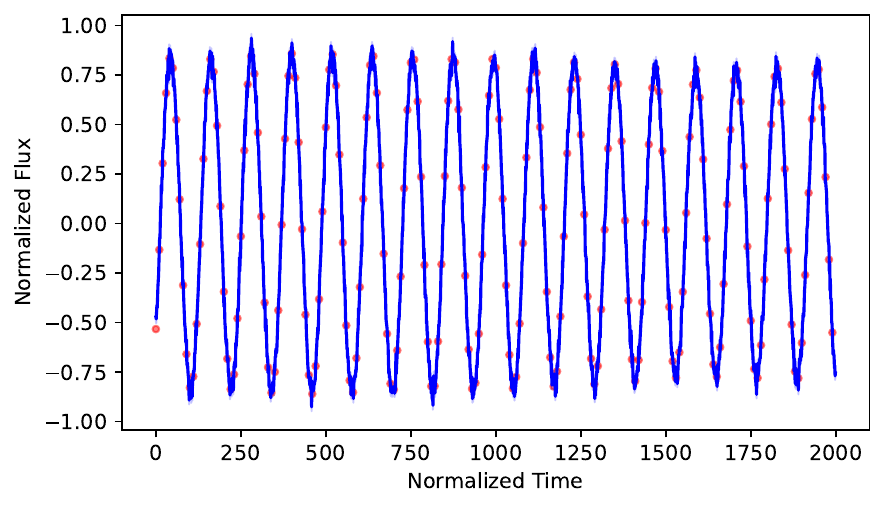} & 
        \includegraphics[width=0.4\textwidth]{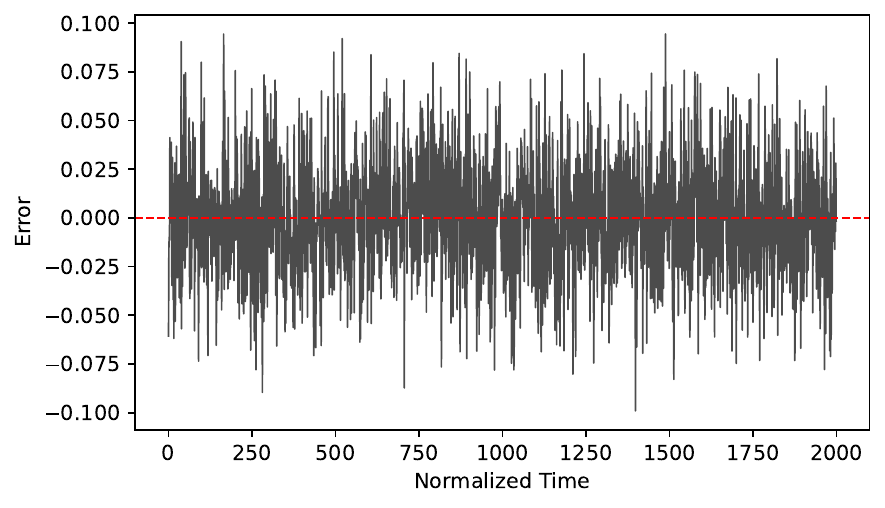} \\
        
        \raisebox{1.5cm}{\textbf{Ultra}} & 
        \includegraphics[width=0.4\textwidth]{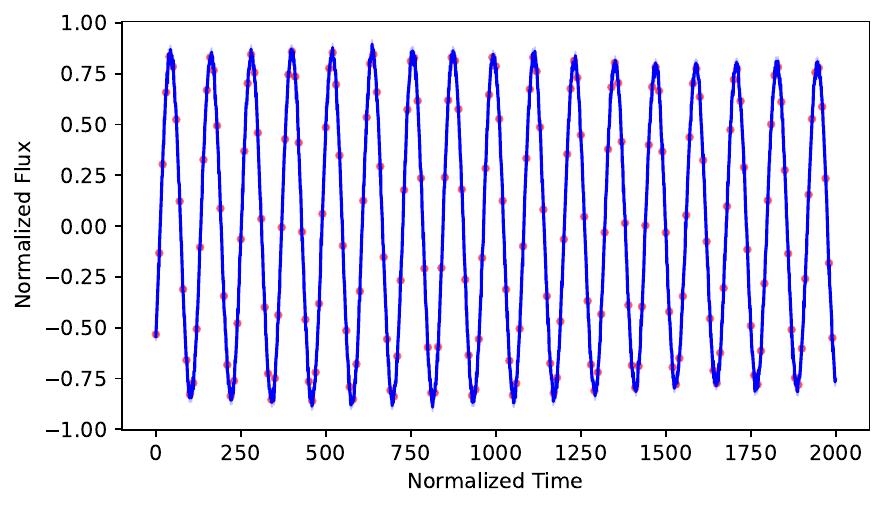} & 
        \includegraphics[width=0.4\textwidth]{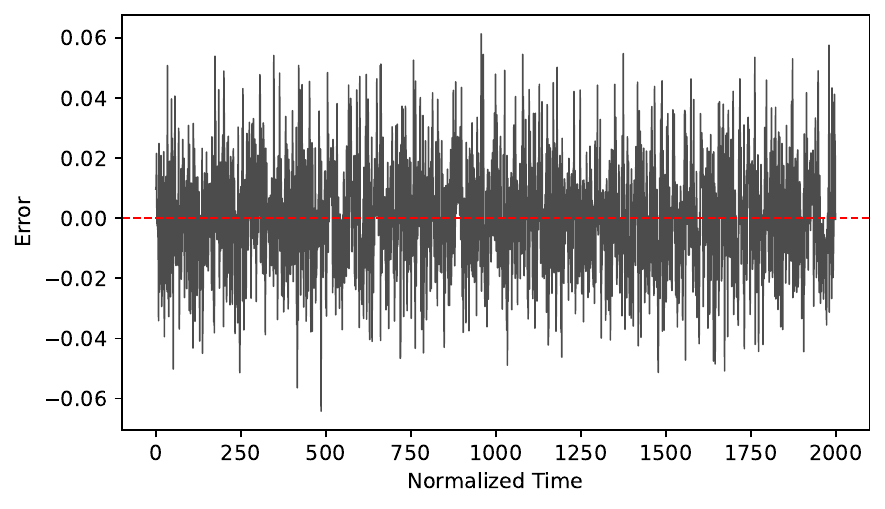} \\
    \end{tabular}

    \vspace{0.5cm} 

    \includegraphics[width=0.6\textwidth]{VAE_fitting_shared_legend.pdf} 

    \vspace{0.5cm} 
    
    \captionof{figure}{Diagnostic plots for \textbf{Scenario 2 ($N=2000$)}. The left column displays the true GP trend versus the surrogate VAE approximation. The right column displays the corresponding residual errors.}
    \label{fig:S2_N2000}
\end{center}


\begin{center}
    \begin{tabular}{ccc}
        & \textbf{VAE Approximation Fit} & \textbf{Residuals} \\
        
        \raisebox{1.5cm}{\textbf{Tiny}} & 
        \includegraphics[width=0.4\textwidth]{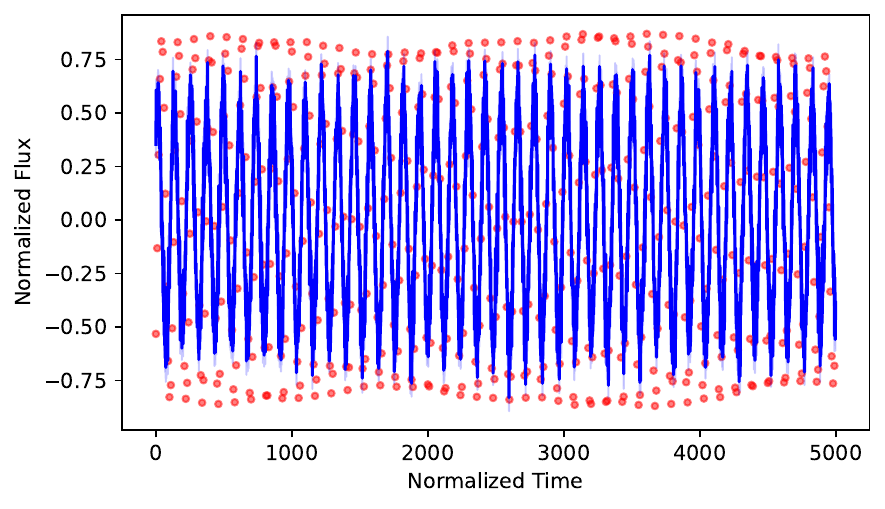} & 
        \includegraphics[width=0.4\textwidth]{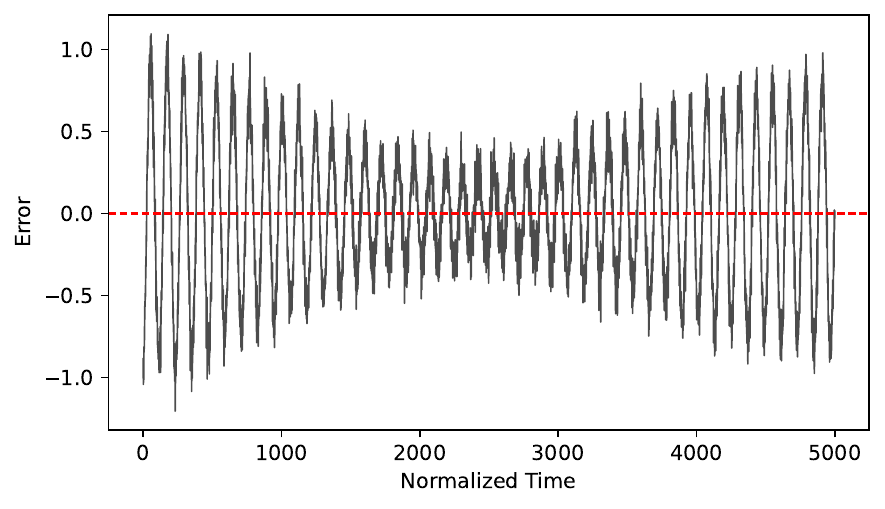} \\
        
        \raisebox{1.5cm}{\textbf{Small}} & 
        \includegraphics[width=0.4\textwidth]{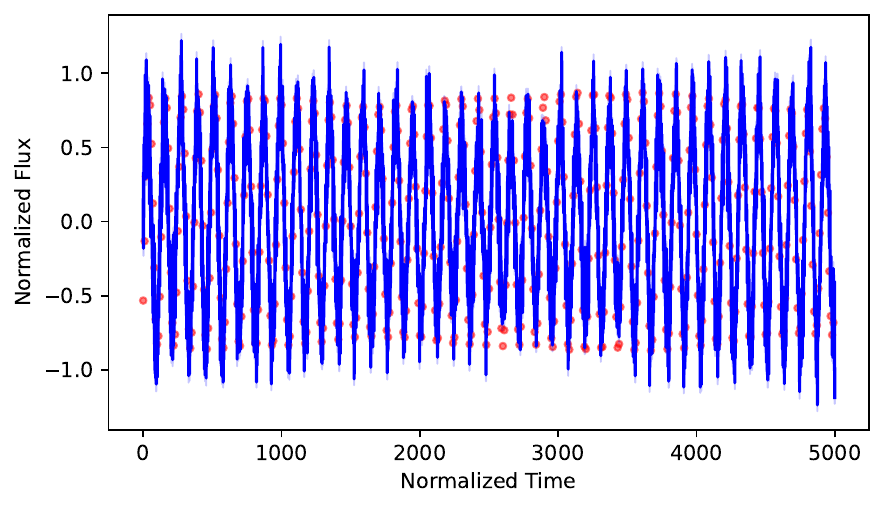} & 
        \includegraphics[width=0.4\textwidth]{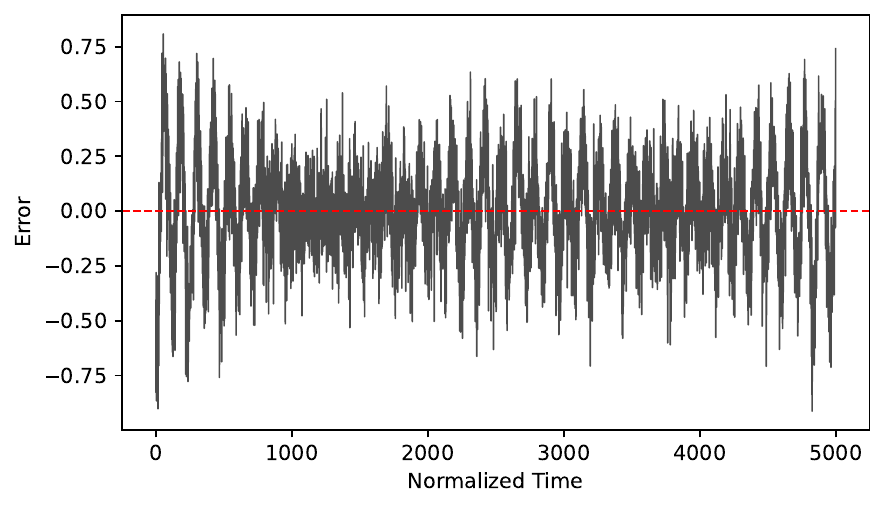} \\
        
        \raisebox{1.5cm}{\textbf{Medium}} & 
        \includegraphics[width=0.4\textwidth]{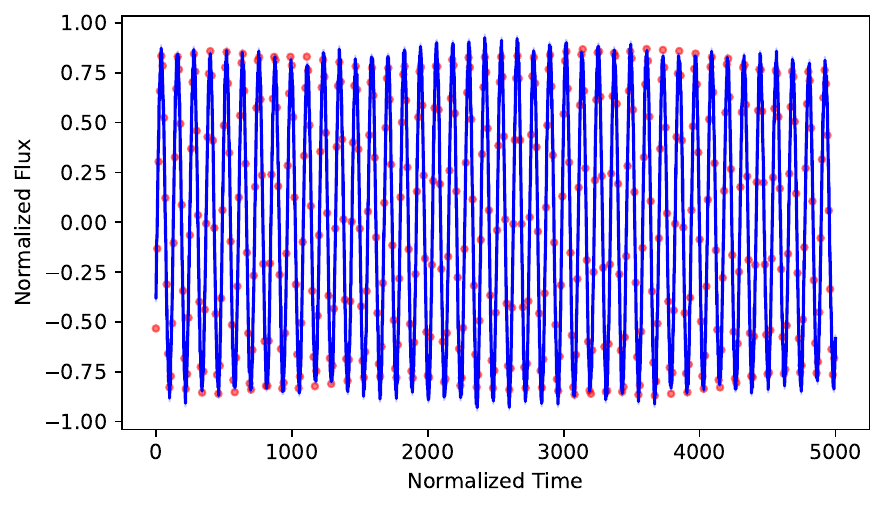} & 
        \includegraphics[width=0.4\textwidth]{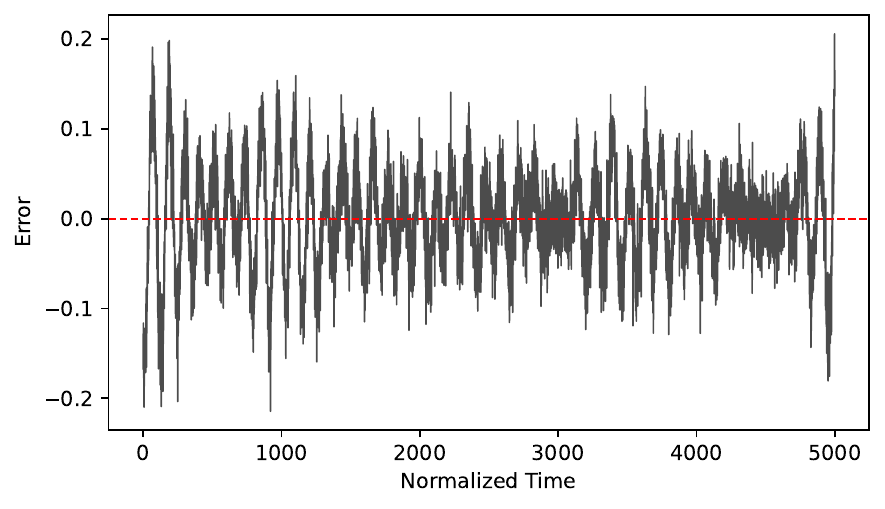} \\
        
        \raisebox{1.5cm}{\textbf{Large}} & 
        \includegraphics[width=0.4\textwidth]{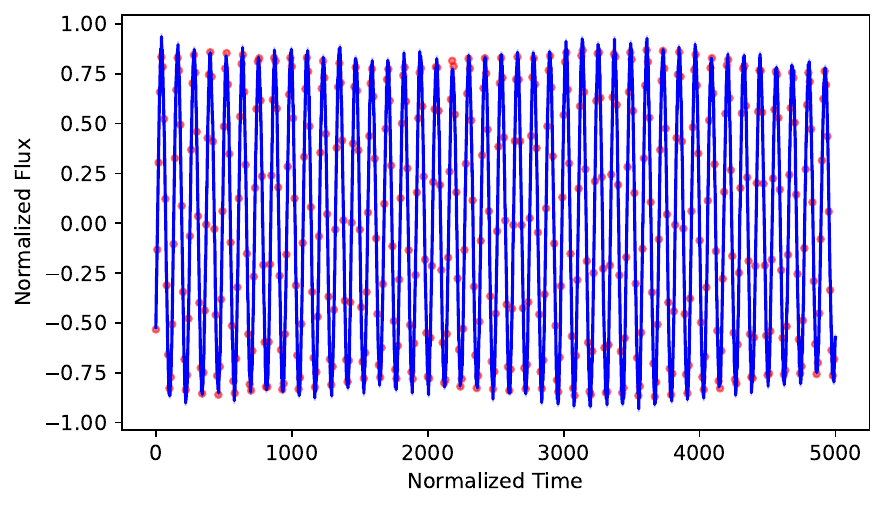} & 
        \includegraphics[width=0.4\textwidth]{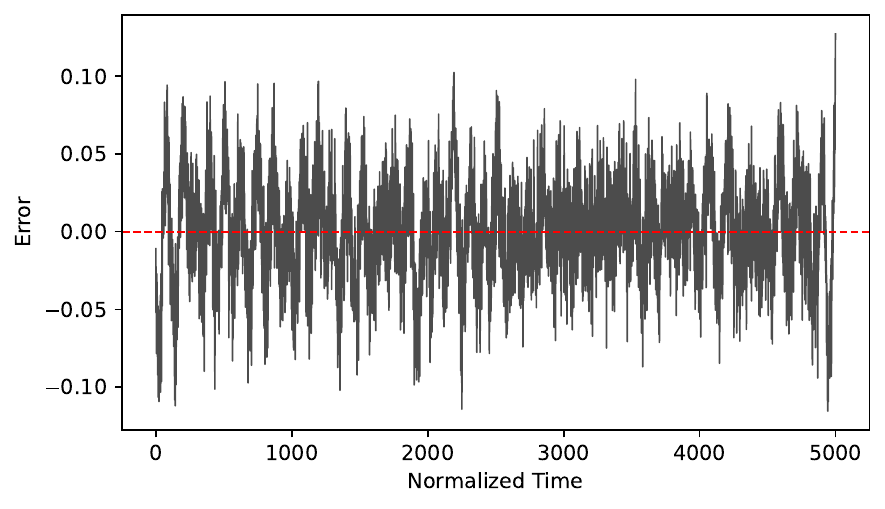} \\
        
        \raisebox{1.5cm}{\textbf{Ultra}} & 
        \includegraphics[width=0.4\textwidth]{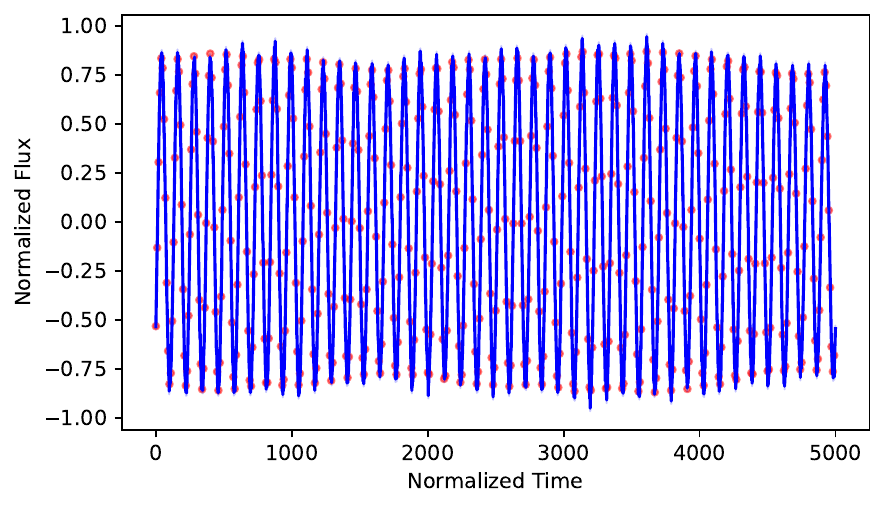} & 
        \includegraphics[width=0.4\textwidth]{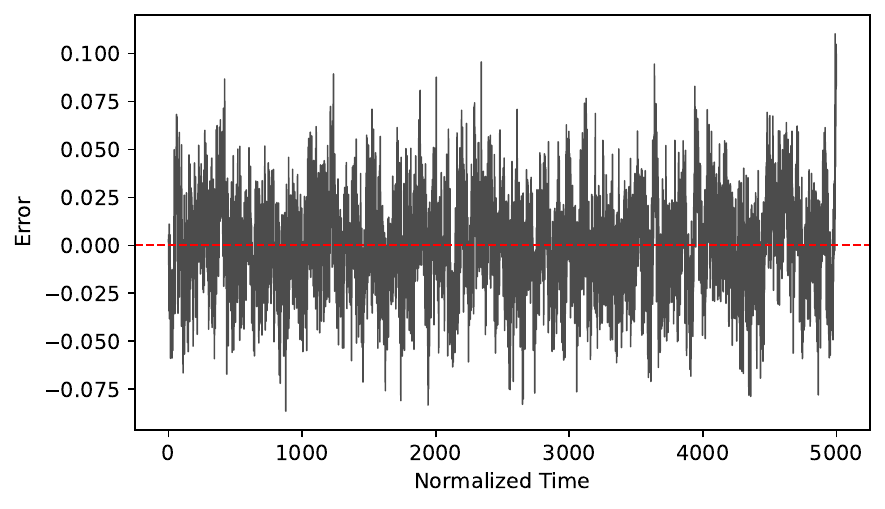} \\
    \end{tabular}

    \vspace{0.5cm} 

    \includegraphics[width=0.6\textwidth]{VAE_fitting_shared_legend.pdf} 

    \vspace{0.5cm} 
    
    \captionof{figure}{Diagnostic plots for \textbf{Scenario 2 ($N=5000$)}. The left column displays the true GP trend versus the surrogate VAE approximation. The right column displays the corresponding residual errors.}
    \label{fig:S2_N5000}
\end{center}


\begin{center}
    \begin{tabular}{ccc}
        & \textbf{VAE Approximation Fit} & \textbf{Residuals} \\
        
        \raisebox{1.5cm}{\textbf{Tiny}} & 
        \includegraphics[width=0.4\textwidth]{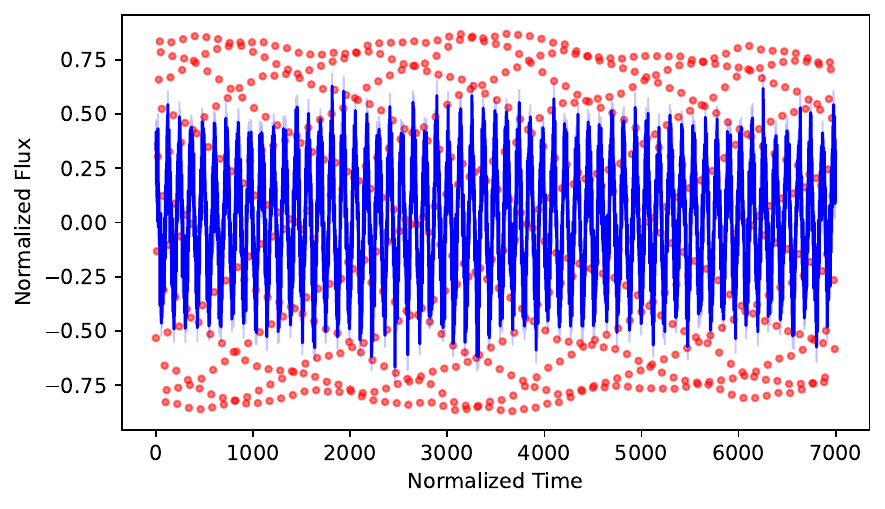} & 
        \includegraphics[width=0.4\textwidth]{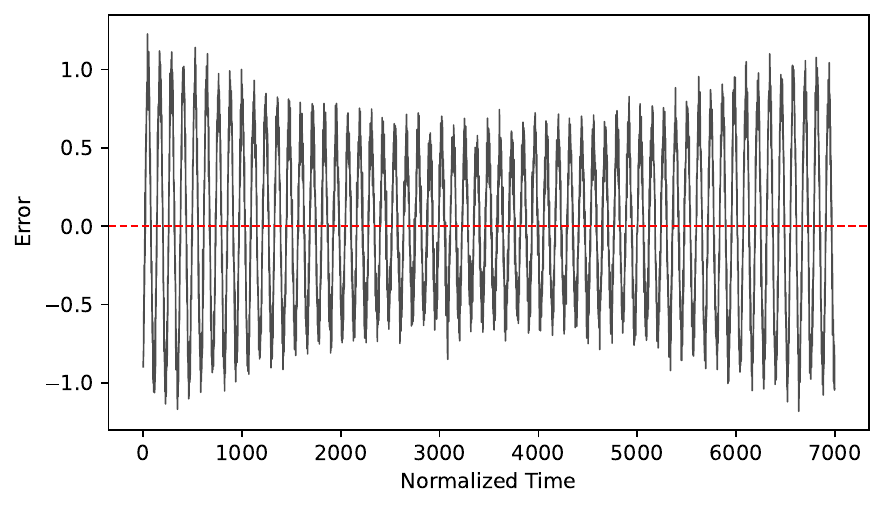} \\
        
        \raisebox{1.5cm}{\textbf{Small}} & 
        \includegraphics[width=0.4\textwidth]{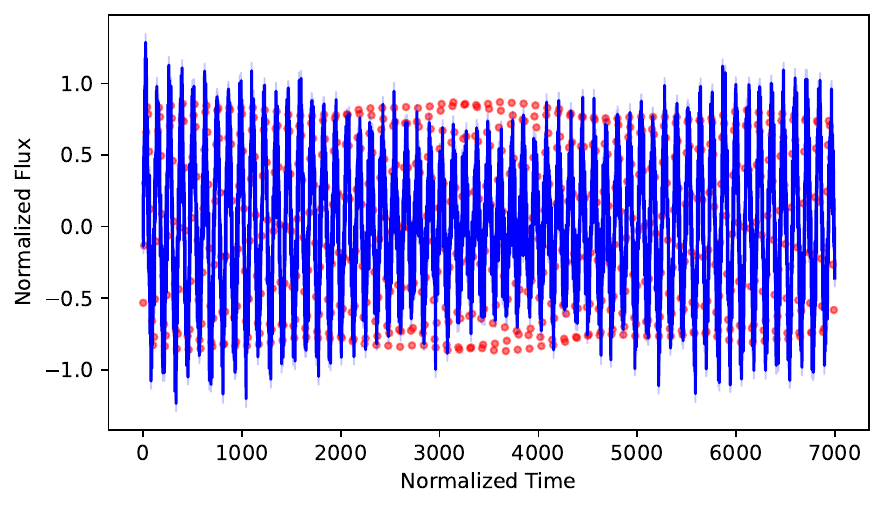} & 
        \includegraphics[width=0.4\textwidth]{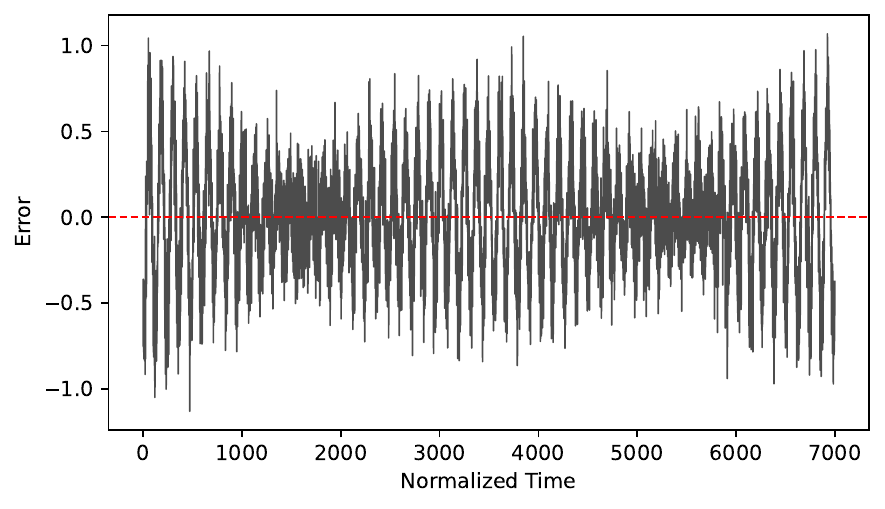} \\
        
        \raisebox{1.5cm}{\textbf{Medium}} & 
        \includegraphics[width=0.4\textwidth]{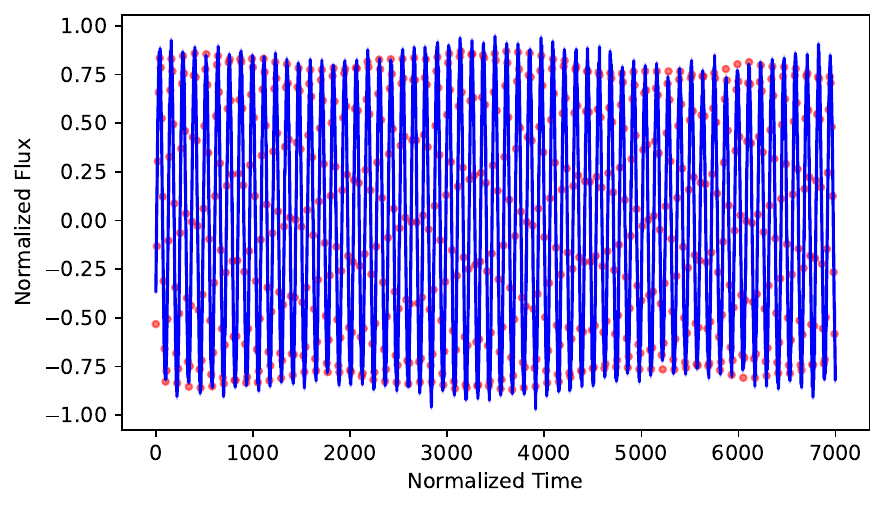} & 
        \includegraphics[width=0.4\textwidth]{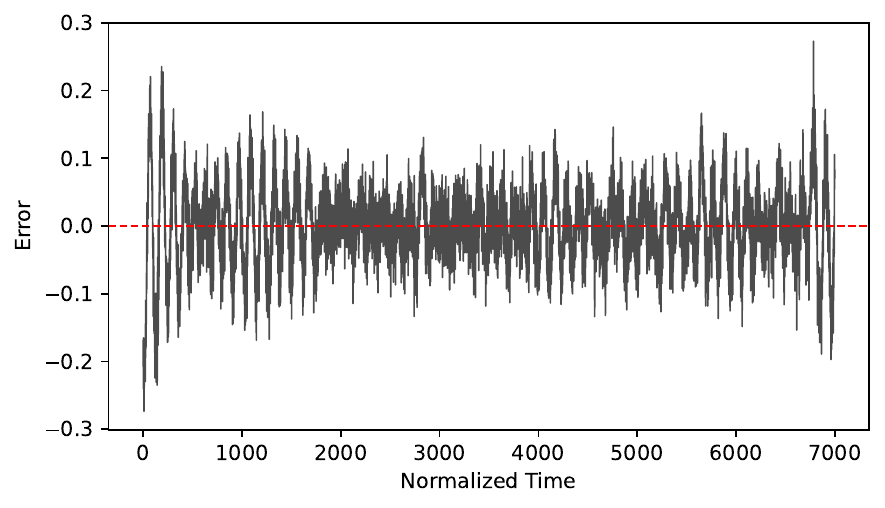} \\
        
        \raisebox{1.5cm}{\textbf{Large}} & 
        \includegraphics[width=0.4\textwidth]{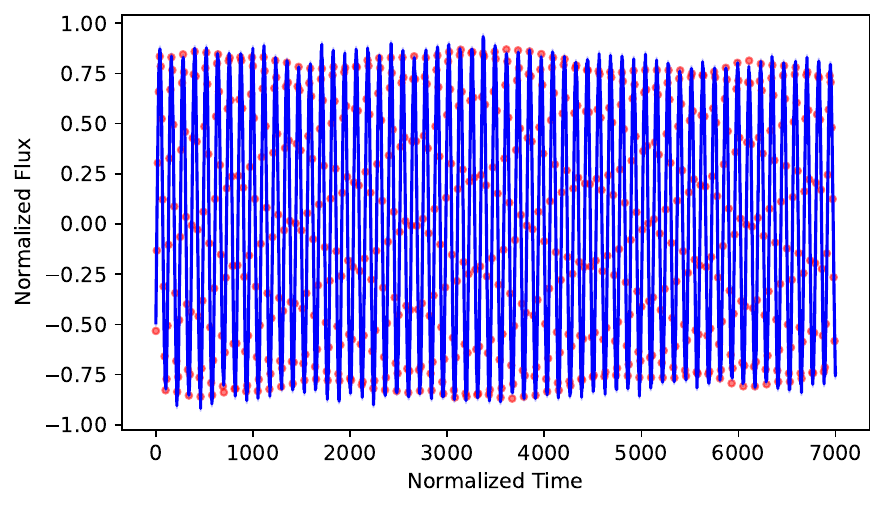} & 
        \includegraphics[width=0.4\textwidth]{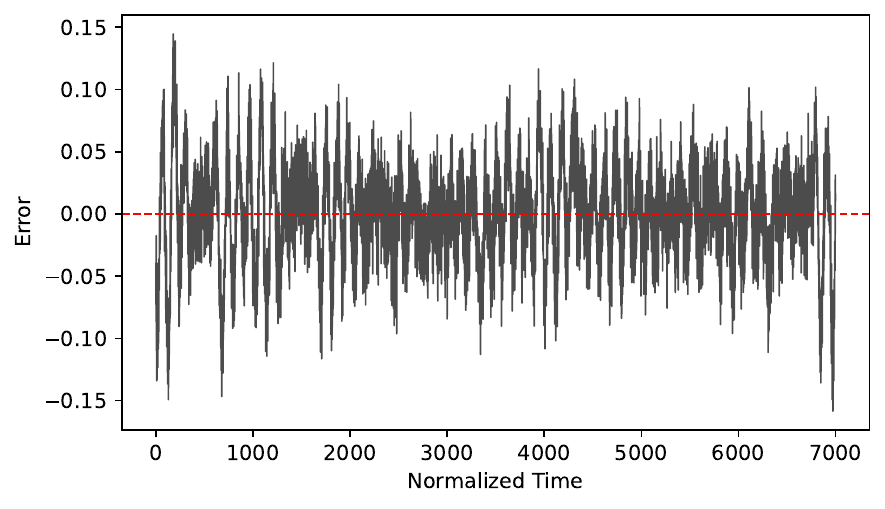} \\
        
        \raisebox{1.5cm}{\textbf{Ultra}} & 
        \includegraphics[width=0.4\textwidth]{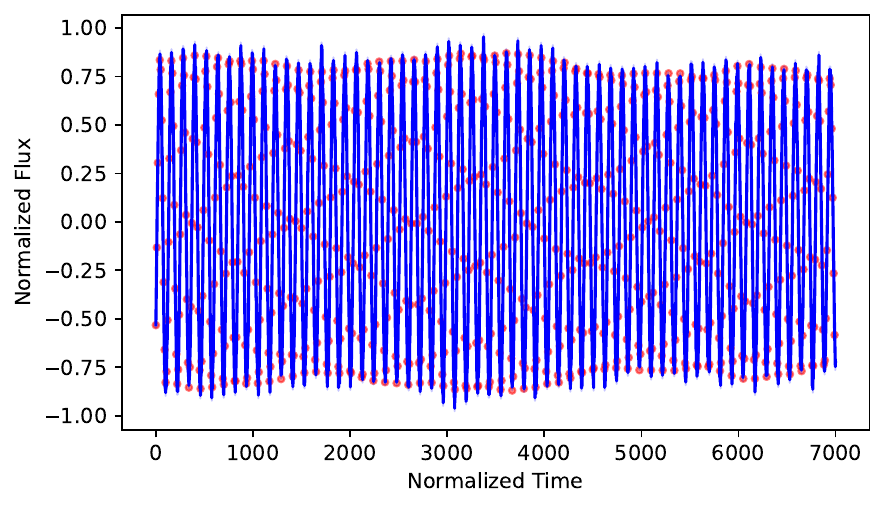} & 
        \includegraphics[width=0.4\textwidth]{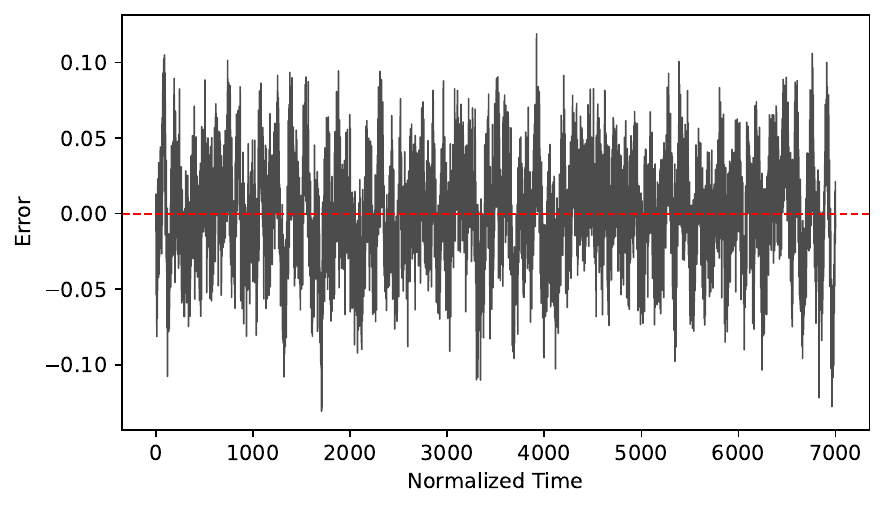} \\
    \end{tabular}

    \vspace{0.5cm} 

    \includegraphics[width=0.6\textwidth]{VAE_fitting_shared_legend.pdf} 

    \vspace{0.5cm} 
    
    \captionof{figure}{Diagnostic plots for \textbf{Scenario 2 ($N=7000$)}. The left column displays the true GP trend versus the surrogate VAE approximation. The right column displays the corresponding residual errors.}
    \label{fig:S2_N7000}
\end{center}

\clearpage

\subsection{Scenario 3: Complex \& Amplitude Modulated}


\begin{center}
    \begin{tabular}{ccc}
        & \textbf{VAE Approximation Fit} & \textbf{Residuals} \\
        
        \raisebox{1.5cm}{\textbf{Tiny}} & 
        \includegraphics[width=0.4\textwidth]{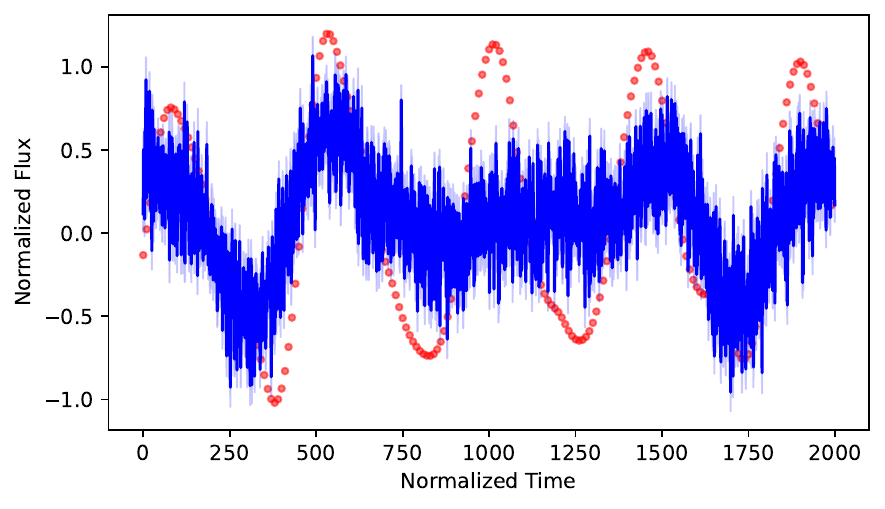} & 
        \includegraphics[width=0.4\textwidth]{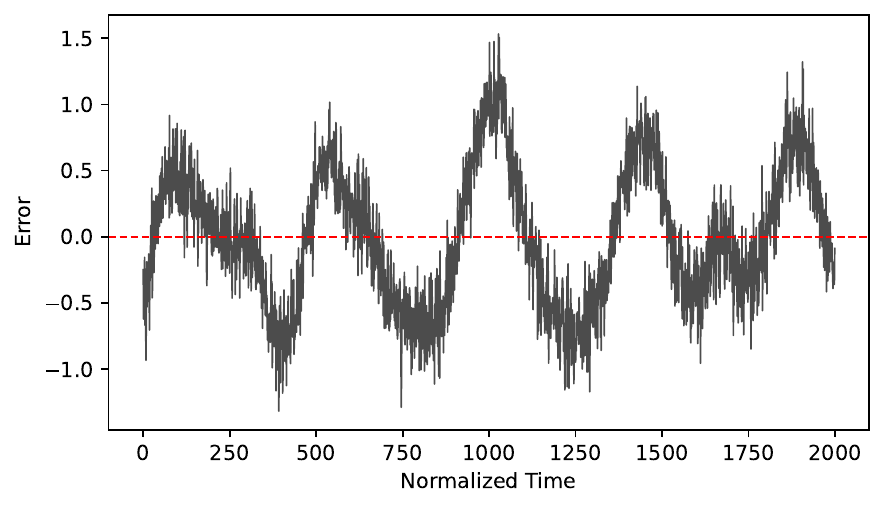} \\
        
        \raisebox{1.5cm}{\textbf{Small}} & 
        \includegraphics[width=0.4\textwidth]{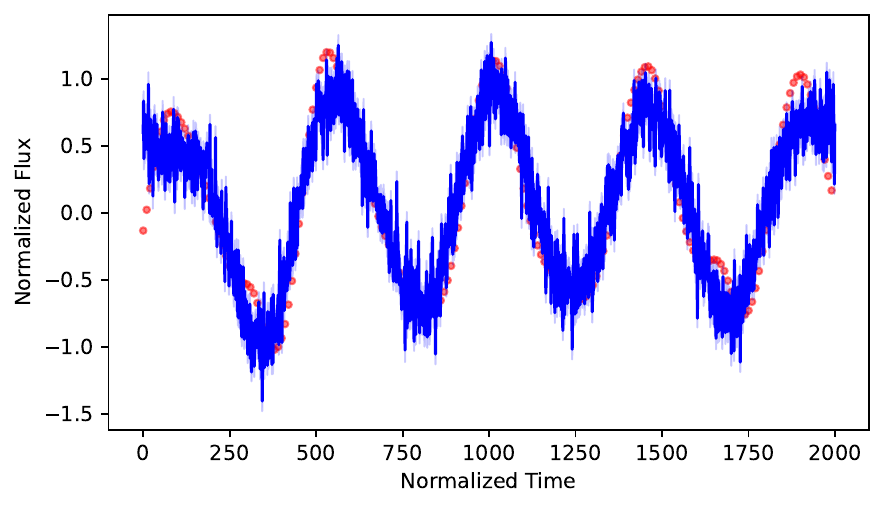} & 
        \includegraphics[width=0.4\textwidth]{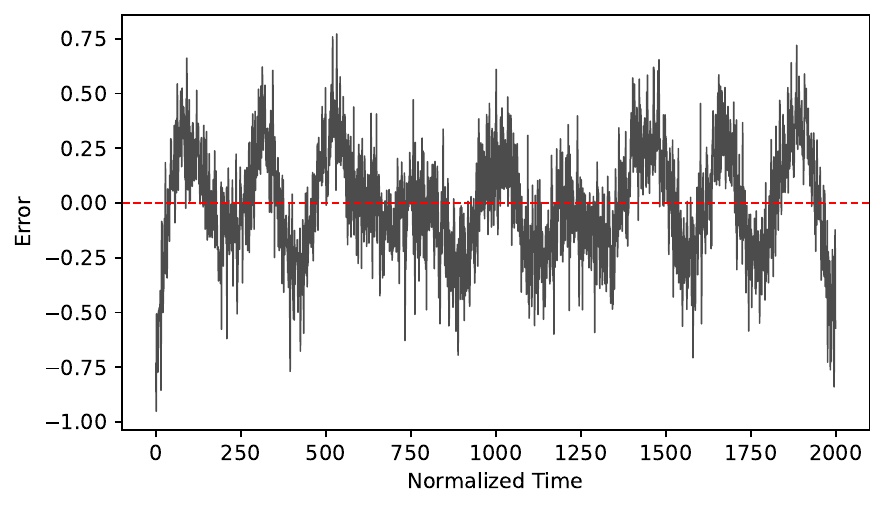} \\
        
        \raisebox{1.5cm}{\textbf{Medium}} & 
        \includegraphics[width=0.4\textwidth]{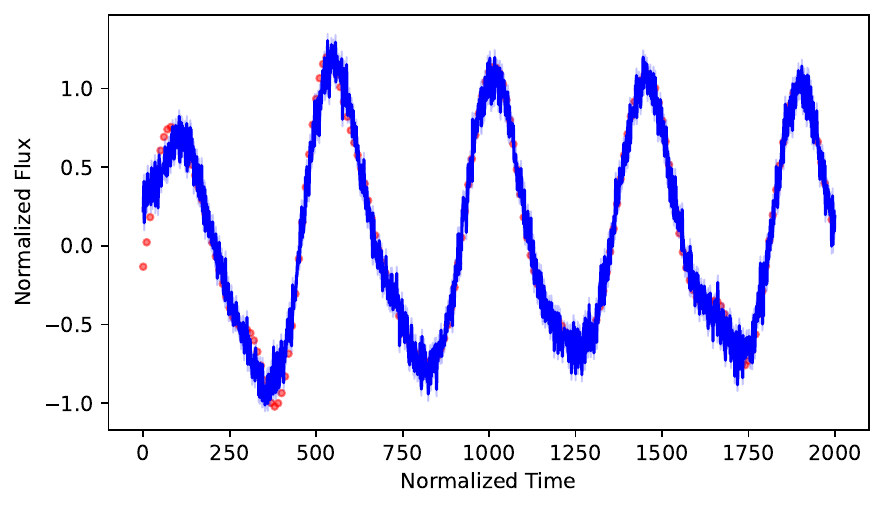} & 
        \includegraphics[width=0.4\textwidth]{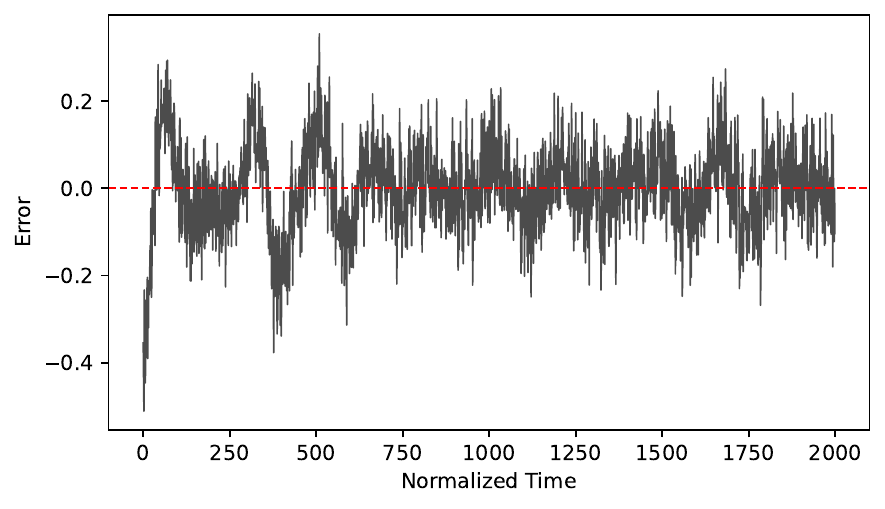} \\
        
        \raisebox{1.5cm}{\textbf{Large}} & 
        \includegraphics[width=0.4\textwidth]{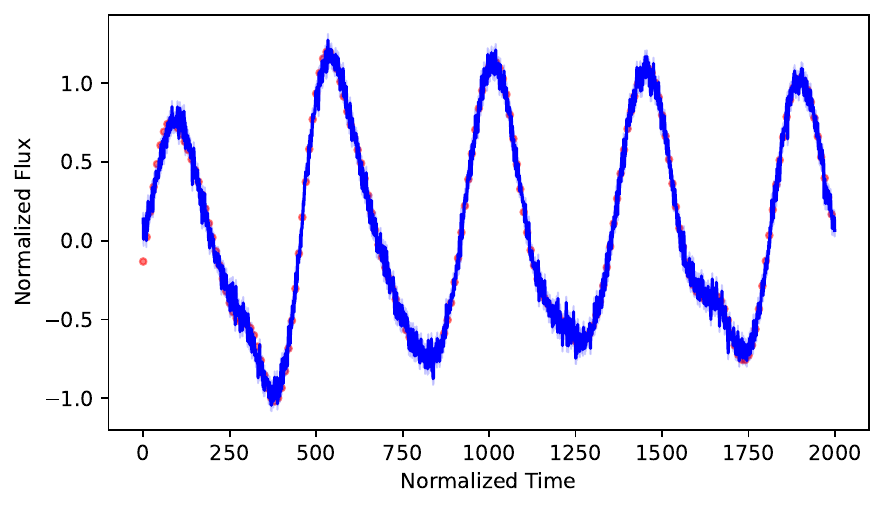} & 
        \includegraphics[width=0.4\textwidth]{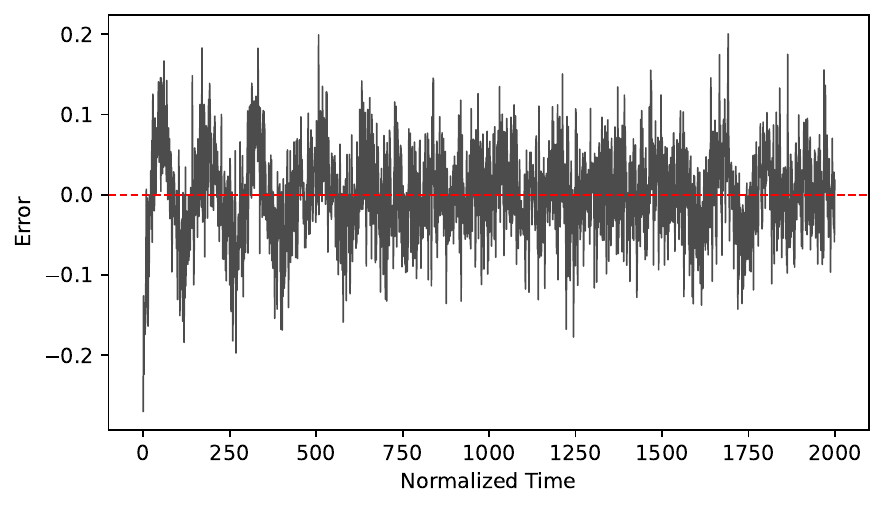} \\
        
        \raisebox{1.5cm}{\textbf{Ultra}} & 
        \includegraphics[width=0.4\textwidth]{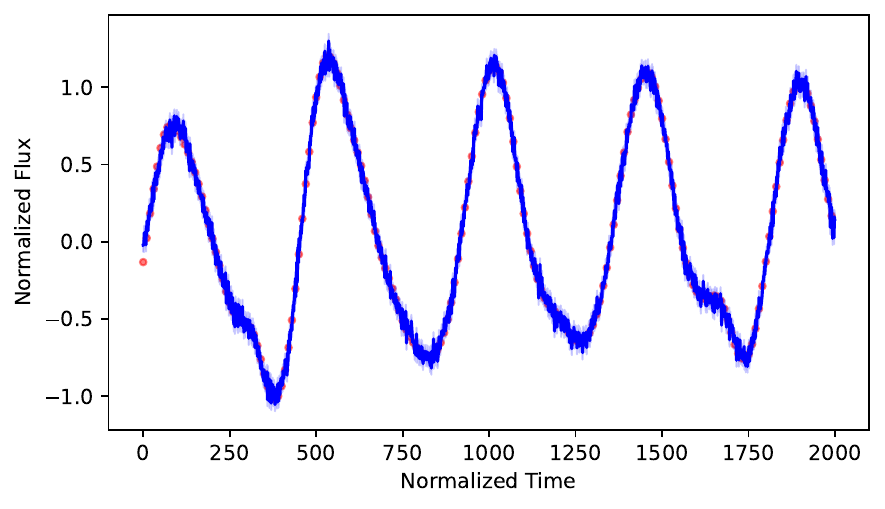} & 
        \includegraphics[width=0.4\textwidth]{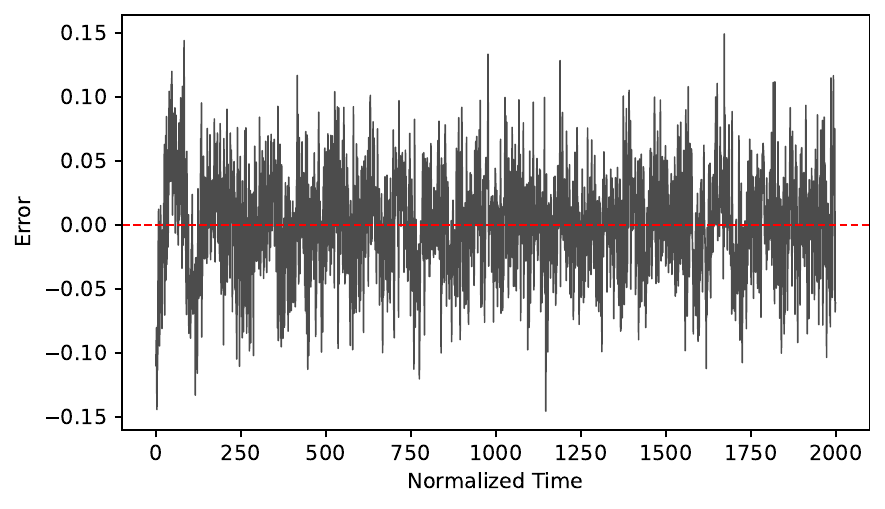} \\
    \end{tabular}

    \vspace{0.5cm} 

    \includegraphics[width=0.6\textwidth]{VAE_fitting_shared_legend.pdf} 

    \vspace{0.5cm} 
    
    \captionof{figure}{Diagnostic plots for \textbf{Scenario 3 ($N=2000$)}. The left column displays the true GP trend versus the surrogate VAE approximation. The right column displays the corresponding residual errors.}
    \label{fig:S3_N2000}
\end{center}


\begin{center}
    \begin{tabular}{ccc}
        & \textbf{VAE Approximation Fit} & \textbf{Residuals} \\
        
        \raisebox{1.5cm}{\textbf{Tiny}} & 
        \includegraphics[width=0.4\textwidth]{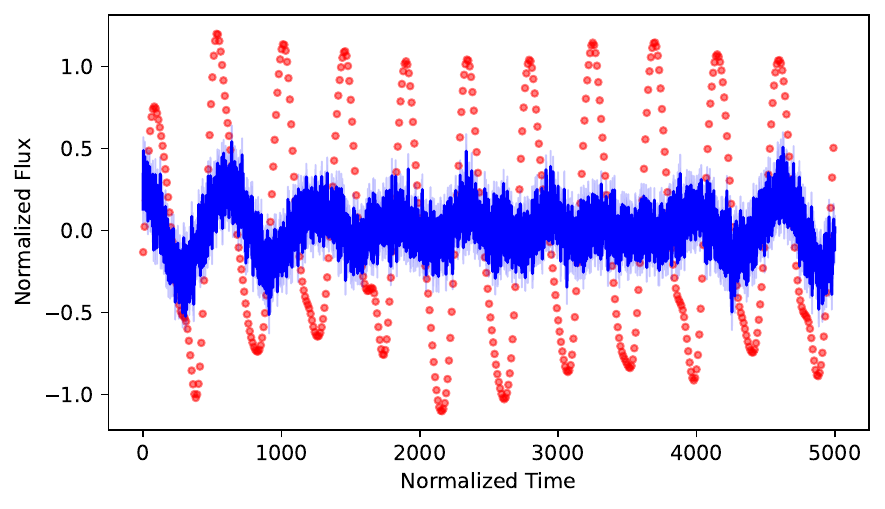} & 
        \includegraphics[width=0.4\textwidth]{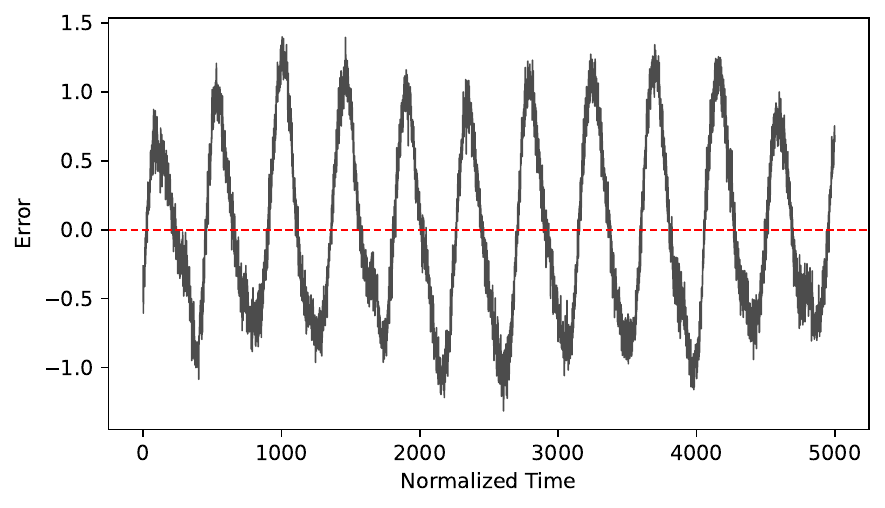} \\
        
        \raisebox{1.5cm}{\textbf{Small}} & 
        \includegraphics[width=0.4\textwidth]{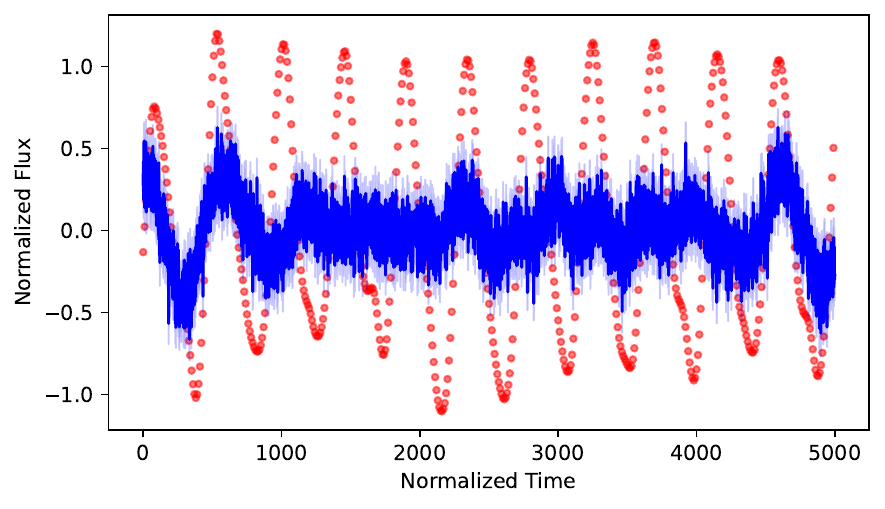} & 
        \includegraphics[width=0.4\textwidth]{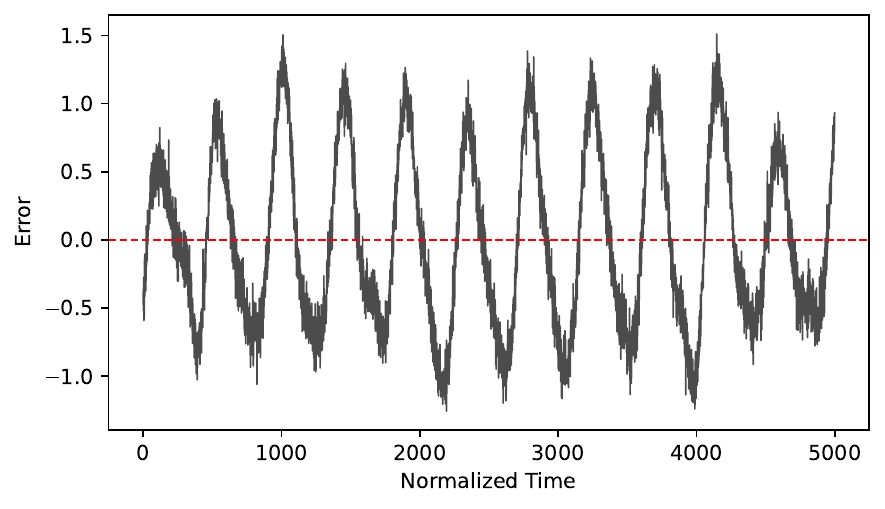} \\
        
        \raisebox{1.5cm}{\textbf{Medium}} & 
        \includegraphics[width=0.4\textwidth]{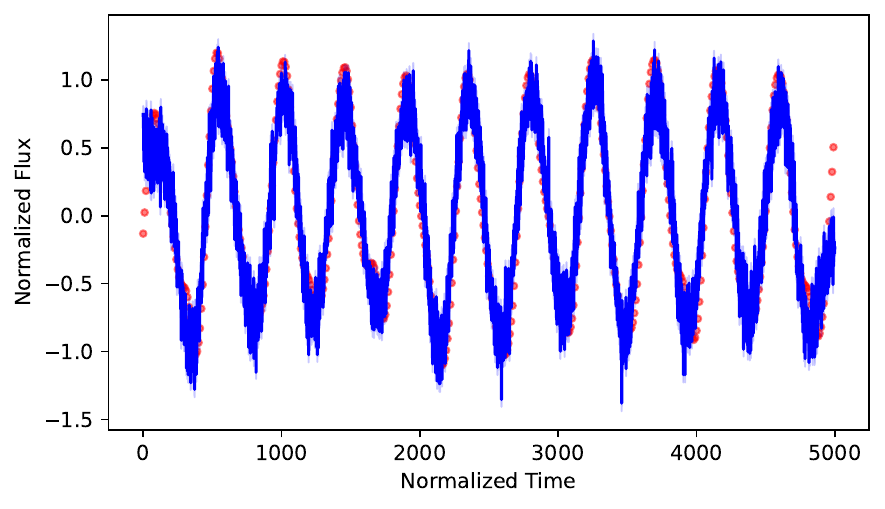} & 
        \includegraphics[width=0.4\textwidth]{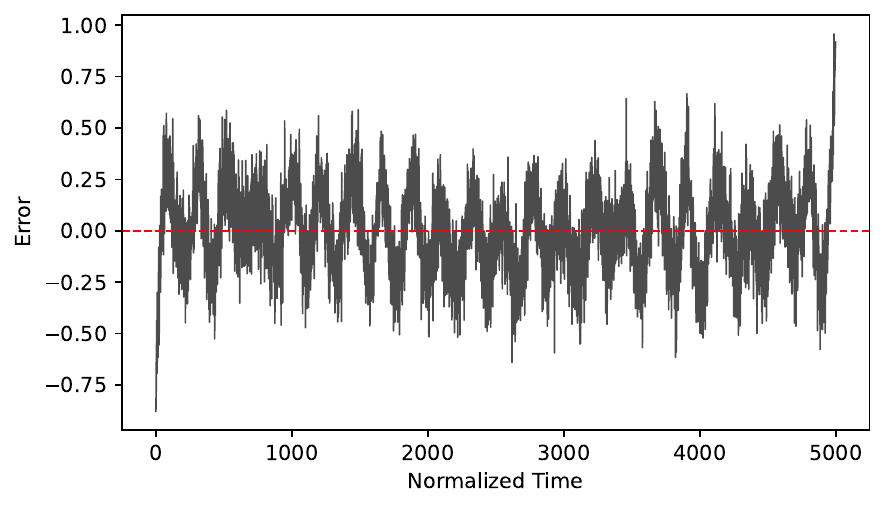} \\
        
        \raisebox{1.5cm}{\textbf{Large}} & 
        \includegraphics[width=0.4\textwidth]{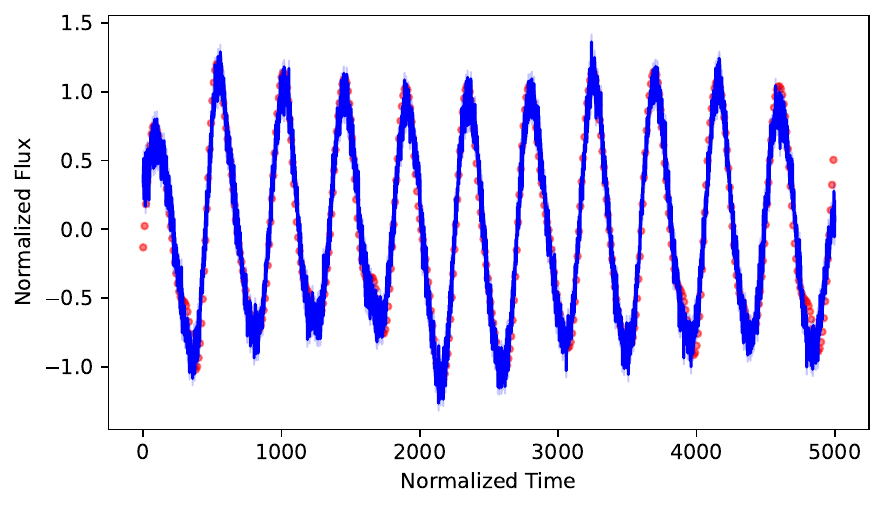} & 
        \includegraphics[width=0.4\textwidth]{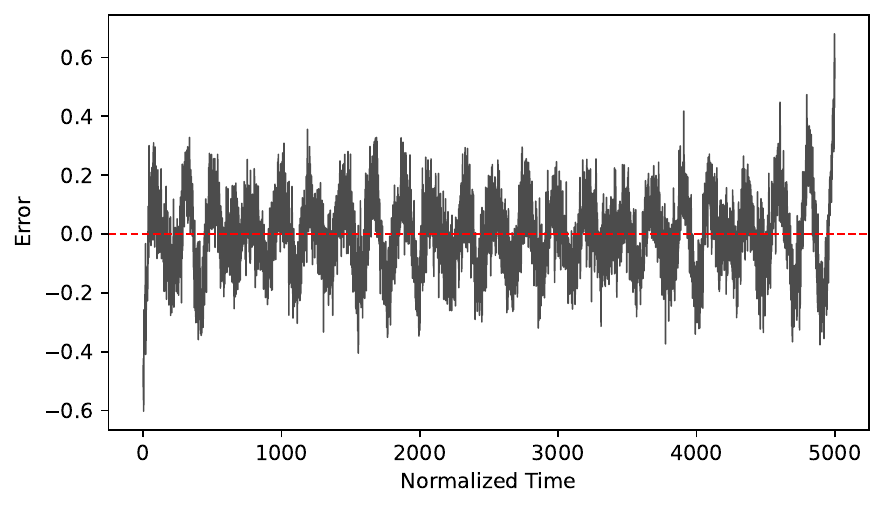} \\
        
        \raisebox{1.5cm}{\textbf{Ultra}} & 
        \includegraphics[width=0.4\textwidth]{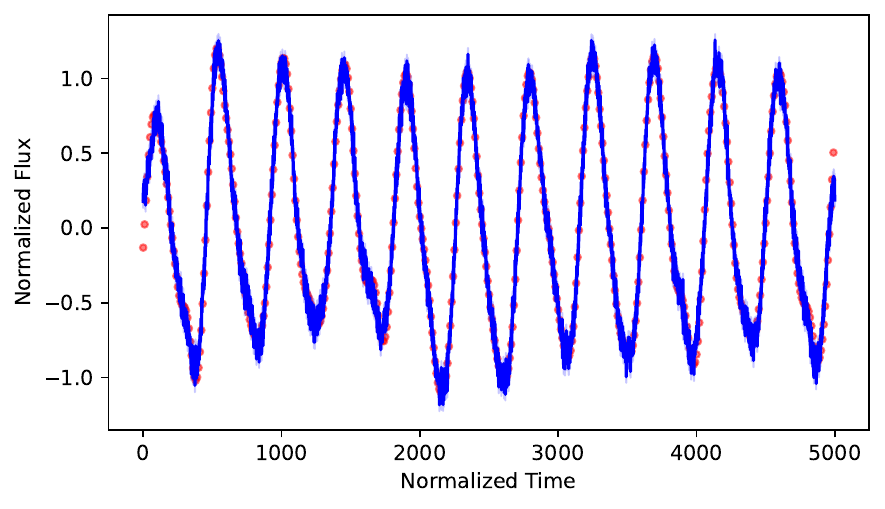} & 
        \includegraphics[width=0.4\textwidth]{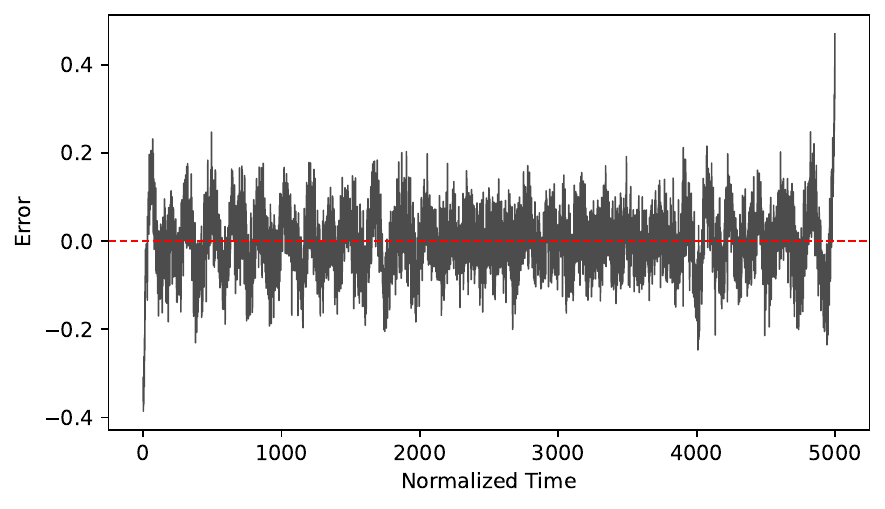} \\
    \end{tabular}

    \vspace{0.5cm} 

    \includegraphics[width=0.6\textwidth]{VAE_fitting_shared_legend.pdf} 

    \vspace{0.5cm} 
    
    \captionof{figure}{Diagnostic plots for \textbf{Scenario 3 ($N=5000$)}. The left column displays the true GP trend versus the surrogate VAE approximation. The right column displays the corresponding residual errors.}
    \label{fig:S3_N5000}
\end{center}


\begin{center}
    \begin{tabular}{ccc}
        & \textbf{VAE Approximation Fit} & \textbf{Residuals} \\
        
        \raisebox{1.5cm}{\textbf{Tiny}} & 
        \includegraphics[width=0.4\textwidth]{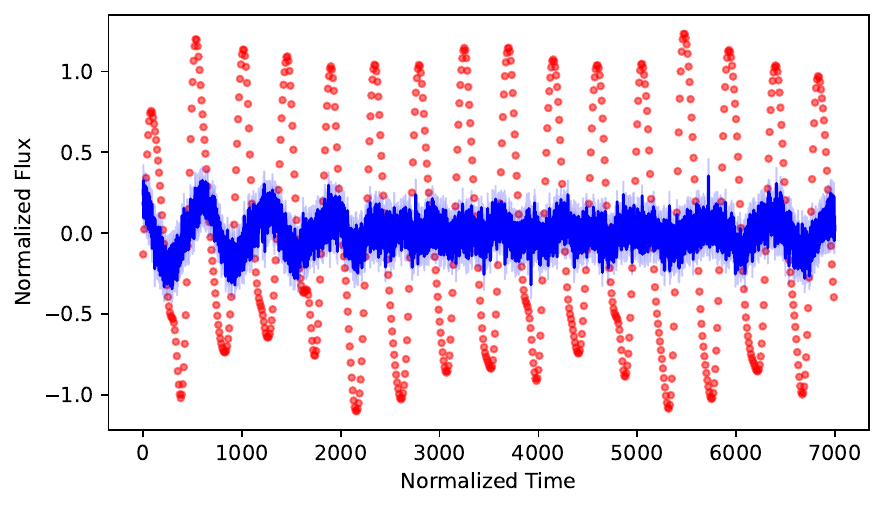} & 
        \includegraphics[width=0.4\textwidth]{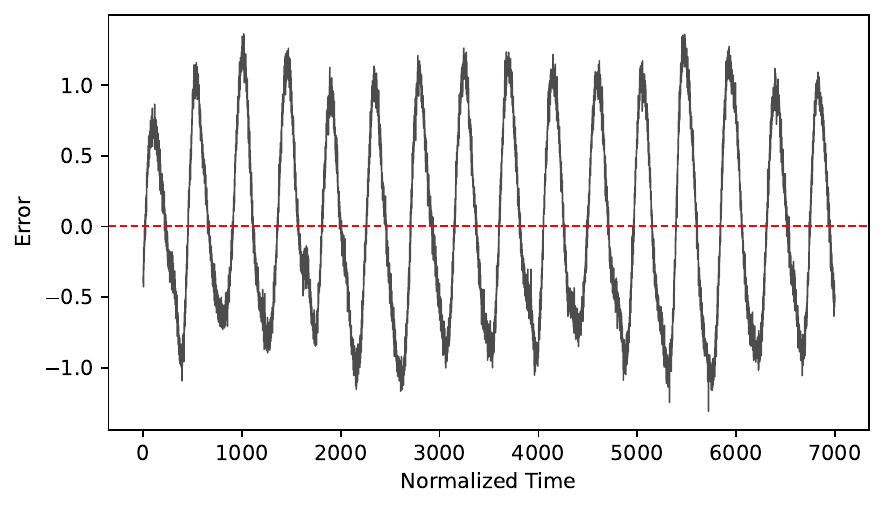} \\
        
        \raisebox{1.5cm}{\textbf{Small}} & 
        \includegraphics[width=0.4\textwidth]{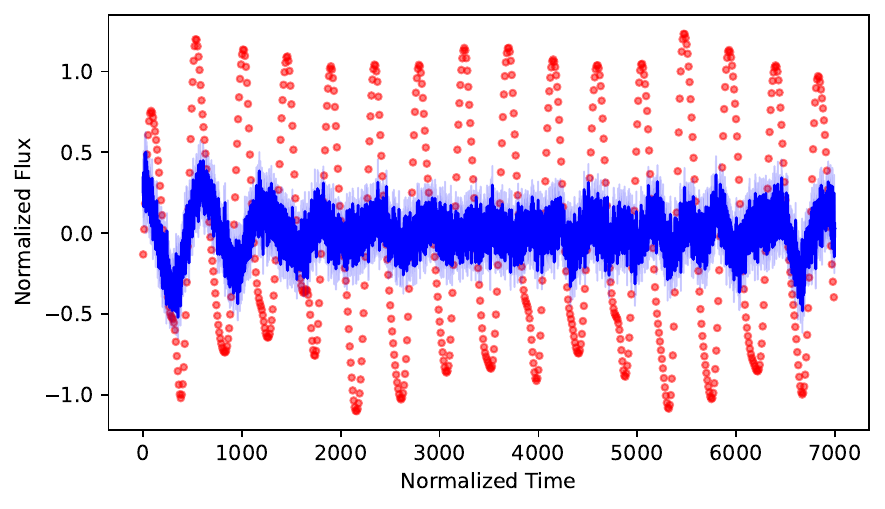} & 
        \includegraphics[width=0.4\textwidth]{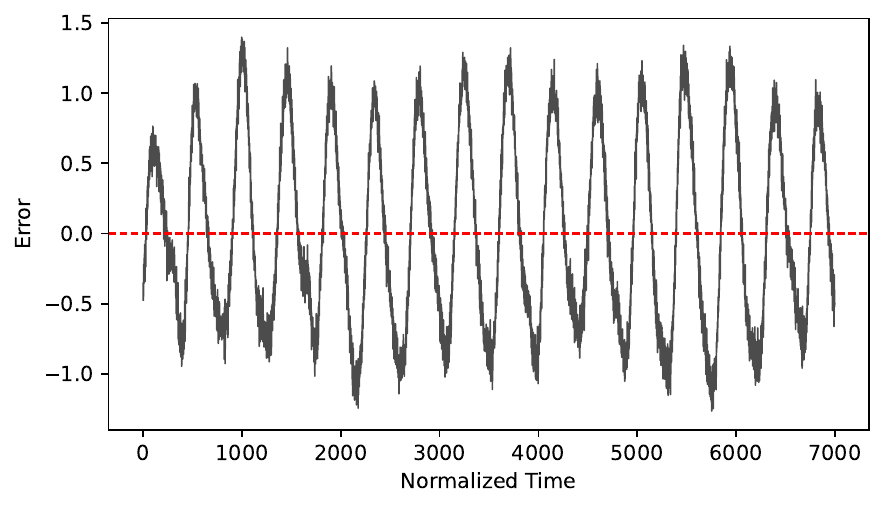} \\
        
        \raisebox{1.5cm}{\textbf{Medium}} & 
        \includegraphics[width=0.4\textwidth]{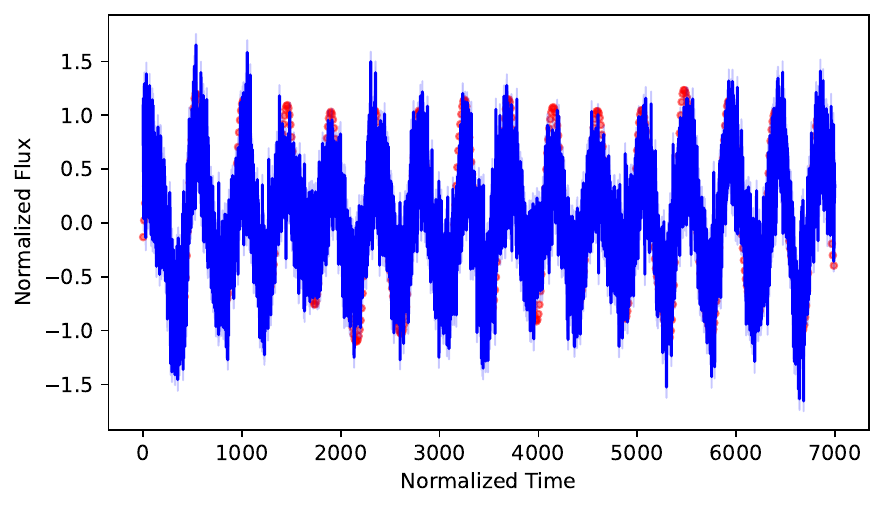} & 
        \includegraphics[width=0.4\textwidth]{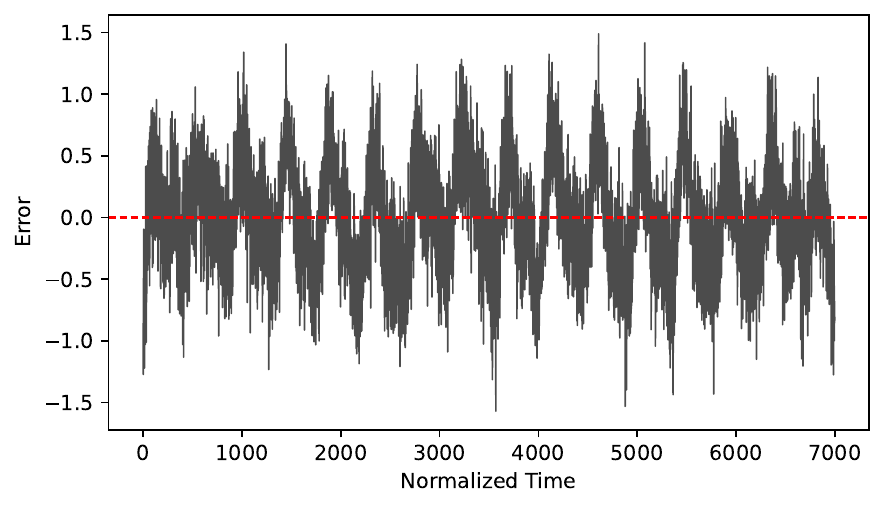} \\
        
        \raisebox{1.5cm}{\textbf{Large}} & 
        \includegraphics[width=0.4\textwidth]{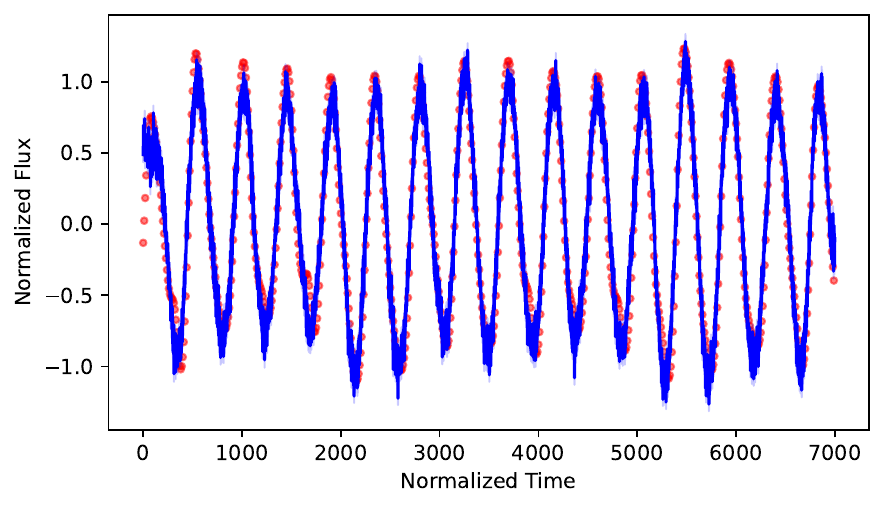} & 
        \includegraphics[width=0.4\textwidth]{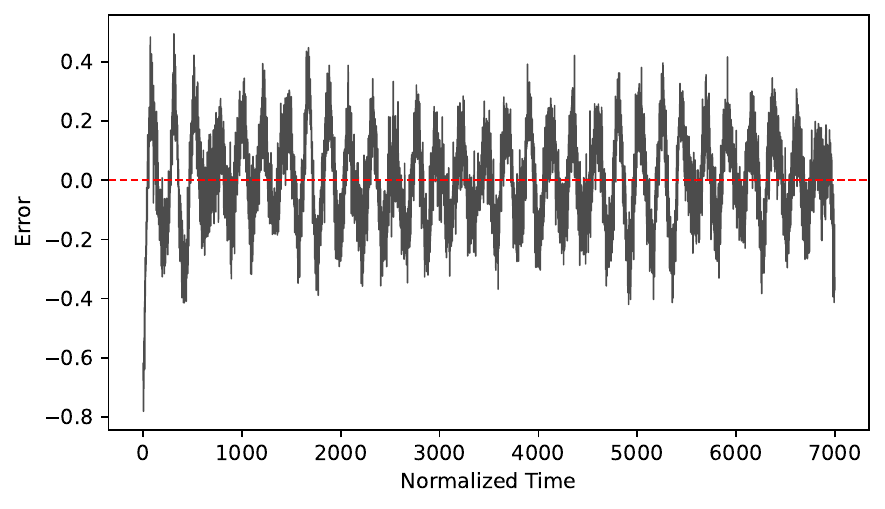} \\
        
        \raisebox{1.5cm}{\textbf{Ultra}} & 
        \includegraphics[width=0.4\textwidth]{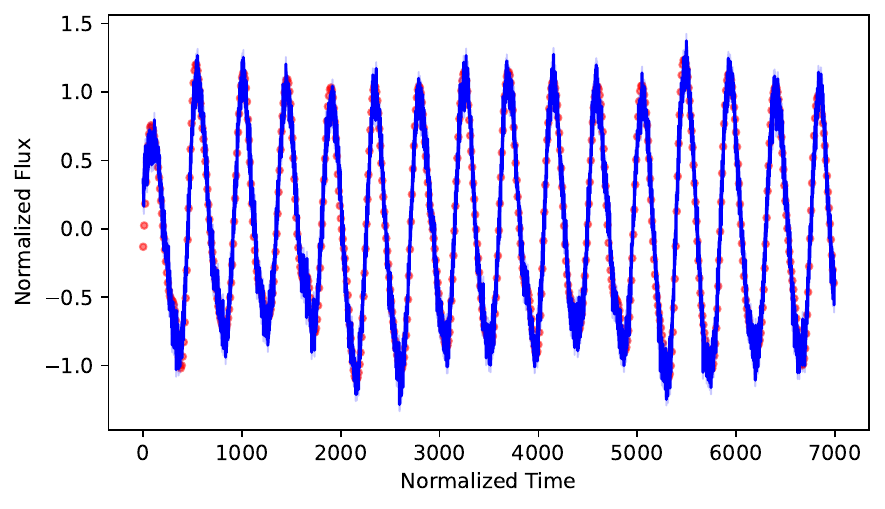} & 
        \includegraphics[width=0.4\textwidth]{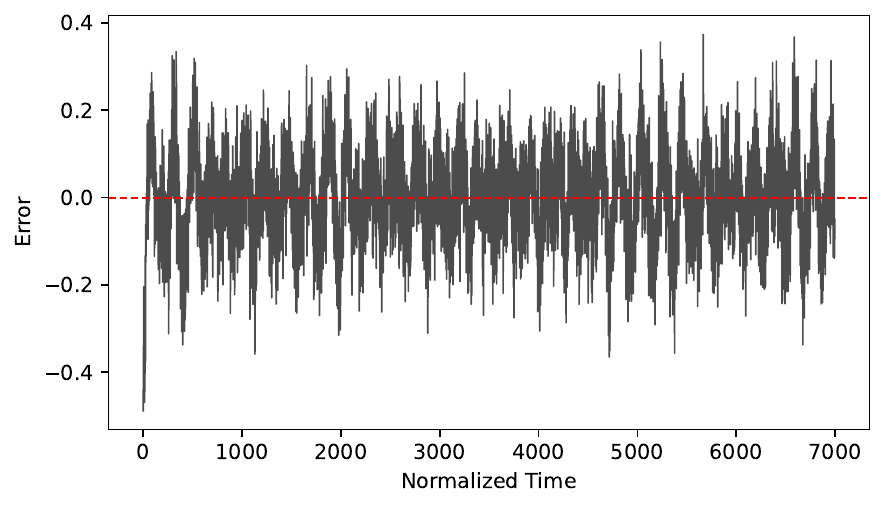} \\
    \end{tabular}

    \vspace{0.5cm} 

    \includegraphics[width=0.6\textwidth]{VAE_fitting_shared_legend.pdf} 

    \vspace{0.5cm} 
    
    \captionof{figure}{Diagnostic plots for \textbf{Scenario 3 ($N=7000$)}. The left column displays the true GP trend versus the surrogate VAE approximation. The right column displays the corresponding residual errors.}
    \label{fig:S3_N7000}
\end{center}

\section{Full Model Results and Convergence Resilience for TIC 031381302}

This supplementary section presents the comprehensive modeling results for the M dwarf target TIC 031381302. Beyond demonstrating the full-scale application of the proposed framework, the analysis of this specific stellar light curve highlights a critical computational advantage of the VAE surrogate over the exact mathematical formulation.

During the evaluation of this target, the standard Celerite+HMM framework exhibited severe pathological behavior. Specifically, the exact Gaussian Process approach struggled to achieve stable Markov chain Monte Carlo (MCMC) convergence within the localized temporal window from $t = 1354.108511$ to $t = 1364.518159$. In this interval, the exact kernel failed to adequately disentangle the complex background morphology from the localized stochastic variation, leading to non-convergent chains and a breakdown of the inference pipeline. 

In stark contrast, the proposed VAE+HMM framework processed the entire observational sequence seamlessly. By leveraging a neural architecture configured with a hidden dimension of $H = 1024$ and a latent dimension of $L = 128$, the surrogate model maintained robust structural fidelity across the entire light curve, successfully resolving the background trend even through the highly complex epoch where the exact \textit{Celerite} model failed. 

Furthermore, the generative surrogate achieved this stable joint inference with high computational efficiency. The complete VAE+HMM pipeline reached full Bayesian convergence for TIC 031381302 in precisely 3,600 seconds. This performance not only demonstrates the capability of the surrogate to accurately detect flares across the full time series, but also underscores its computational resilience in highly complex regimes where the exact Gaussian Process becomes computationally intractable or analytically unstable.

\begin{figure}[htbp]
    \centering
    \includegraphics[width=0.9\textwidth]{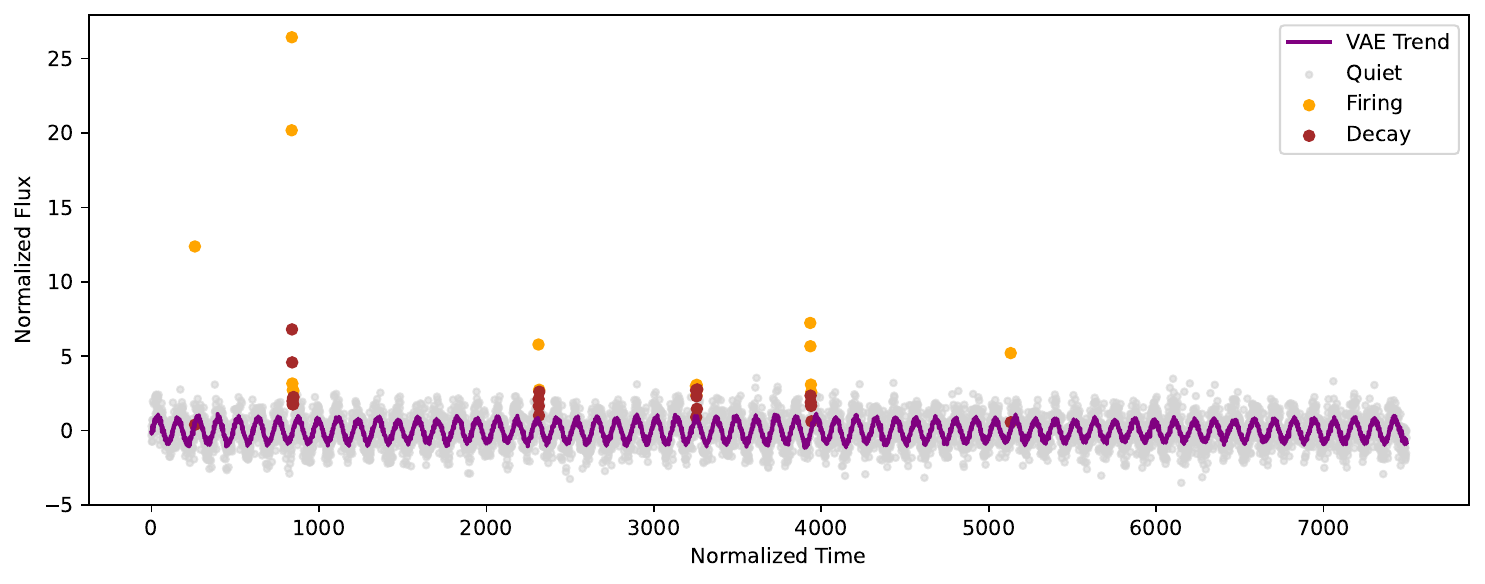} 
    
    \caption{TIC 031381302 mean-centered light curve along with the fit of the proposed VAE+HMM framework, which simultaneously models the continuous background trend using the VAE surrogate and hidden Markov model, as well as assigns discrete states (Quiet, Firing, Decay) to each point in the time series via state decoding post-inference.}
    \label{fig:full_model_starmedium}
\end{figure}

\section{Full Model Results and Extended Time Series Analysis for TIC 234526939}

Following the detailed analysis of our primary target, TIC 089257479, this supplementary section presents the comprehensive modeling results for the M dwarf target TIC 234526939. Mirroring the structure of the main text, this appendix provides both the full-scale application of the proposed VAE+HMM framework and a zoomed-in examination of a localized, complex flaring event. Crucially, the analysis of this specific stellar light curve highlights the structural fidelity and robustness of the surrogate model when applied to a highly challenging dataset characterized by high-frequency background oscillations and dense flare clusters.

\begin{figure}[htbp]
    \centering
    \includegraphics[width=0.9\textwidth]{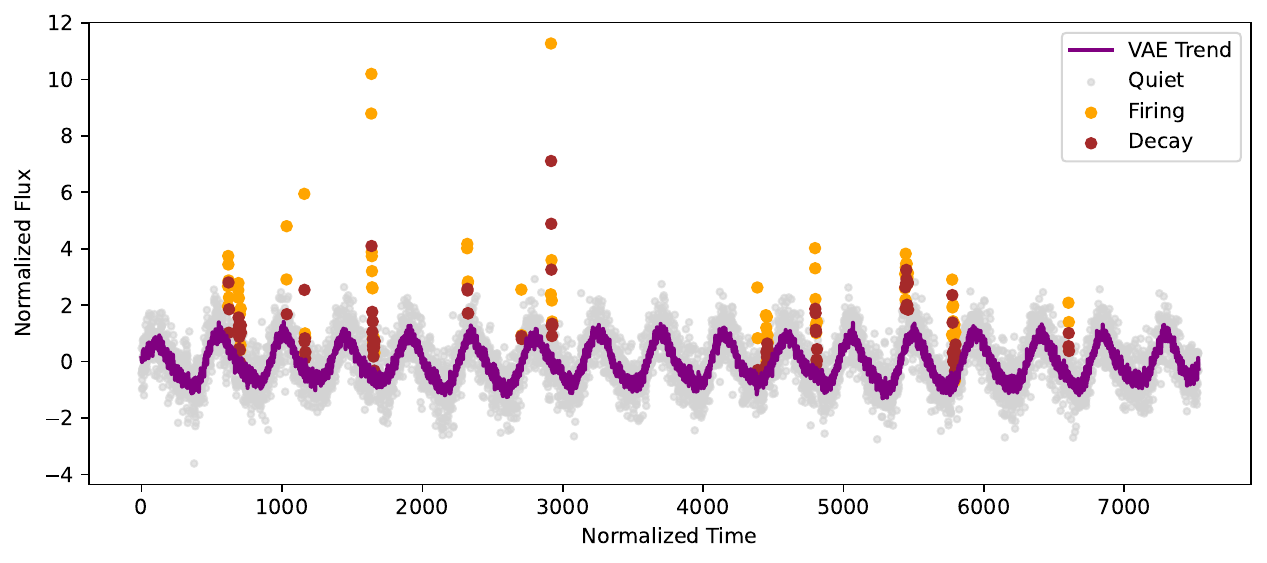} 
    
    \caption{TIC 234526939 mean-centered light curve along with the fit of the proposed VAE+HMM framework, which simultaneously models the continuous background trend using the VAE surrogate and hidden Markov model, as well as assigns discrete Markov states (Quiet, Firing, Decay) to each point in the time series via state decoding post-inference.}
    \label{fig:full_model_starhard}
\end{figure}

Figure \ref{fig:full_model_starhard} presents the full light curve for TIC 234526939, representing a significantly more challenging dataset. The underlying VAE trend must adapt to a highly complex, rapidly oscillating background structure. Despite this high-frequency, noisy variation, the VAE surrogate avoids overfitting to the localized extremes, maintaining a continuous baseline that allows the HMM to successfully detect a dense cluster of flare events scattered heavily throughout the entire observation window.

\begin{figure}[htbp]
    \centering
    
    \begin{minipage}{\textwidth}
        \centering
        \includegraphics[width=0.85\textwidth]{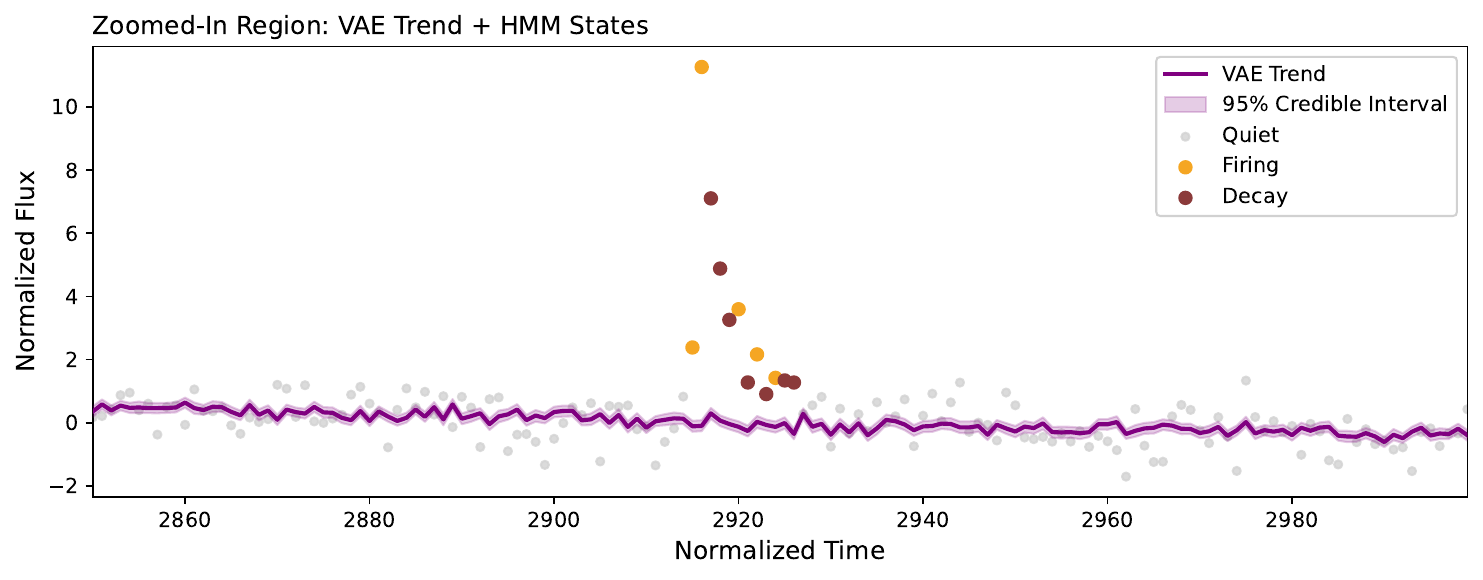} 
    \end{minipage}
    
    \vspace{0.3cm} 
    
    \begin{minipage}{\textwidth}
        \centering
        \includegraphics[width=0.85\textwidth]{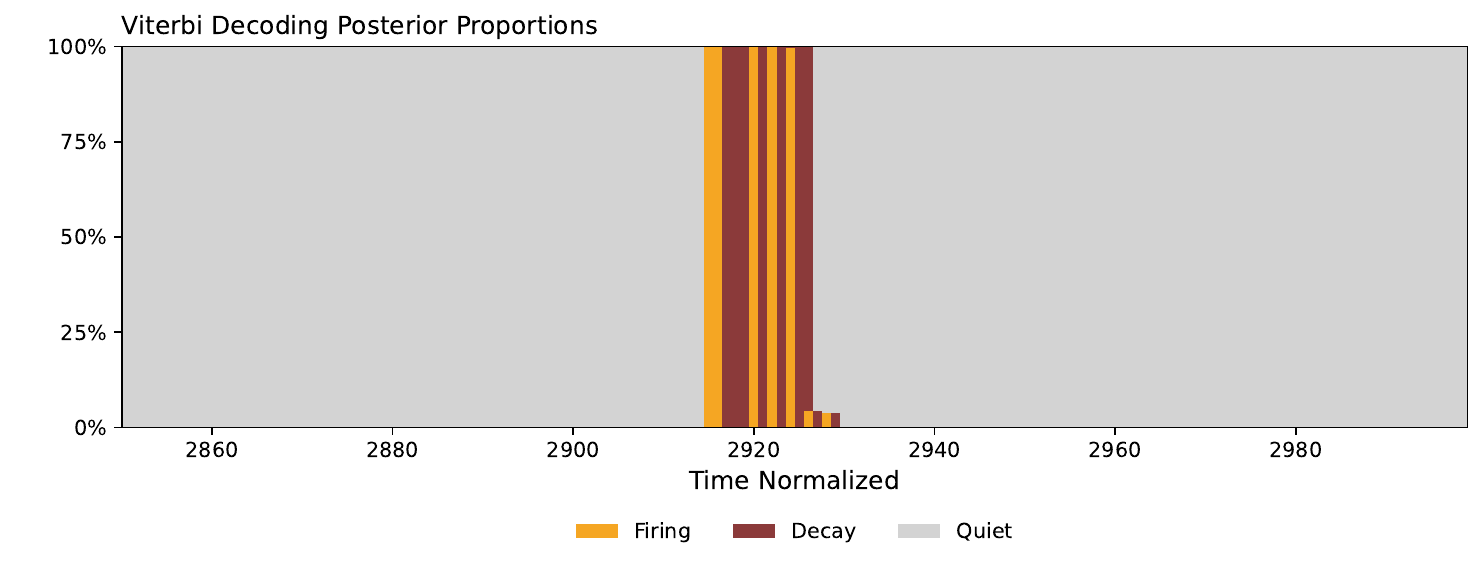} 
    \end{minipage}
    
    \caption{Detected flare example from implementing the VAE+HMM framework on the real time series of TIC 234526939, focusing on the specific normalized time interval from $t = 2850$ to $t = 3000$. \textbf{Top panel:} Shows the fit of the VAE+HMM model to this localized flare event, including the estimated continuous background trend and the assigned state for each point. \textbf{Bottom panel:} State ``decoding distributions'' across all Viterbi sequences per observation, representing the posterior probabilities of the assigned states over this exact same time interval.}
    \label{fig:zoom_flare_starhard}
\end{figure}

Figure \ref{fig:zoom_flare_starhard} focuses on a localized time interval ($t = 2850$ to $t = 3000$) for this complex star. Even with the undulating VAE trend shifting directly beneath the flare, the framework identifies the discrete state transitions. The top panel shows a clear sequence from the Quiet state, to an intense Firing phase, through an extended Decay phase, and smoothly back to Quiet. The corresponding Viterbi proportions in the bottom panel reaffirm this precision; despite the inherent volatility of the star's background emission, the state transitions remain remarkably sharp, maintaining extremely high certainty for the duration of the event.

\end{document}